\documentclass[preprints,article,accept,pdftex,moreauthors]{Definitions/mdpi} 

\firstpage{1} 
\pubvolume{1}
\issuenum{1}
\articlenumber{0}
\pubyear{2022}
\copyrightyear{2022}
\datereceived{} 
\dateaccepted{} 
\datepublished{} 
\hreflink{https://doi.org/} 

\usepackage{bm}
\usepackage{amssymb}
\usepackage{cleveref}
\usepackage{arydshln}
\usepackage{relsize}
\usepackage[toc, nonumberlist]{glossaries}
\newacronym{SRCNN}{SRCNN}{Super-Resolution Convolutional Neural Network}
\newacronym{SR}{SR}{Super-Resolution}
\newacronym{SOTA}{SOTA}{state-of-the-art}
\newacronym{SISR}{SISR}{Single Image \gls{SR}}
\newacronym{HR}{HR}{High-Resolution}
\newacronym{LR}{LR}{Low-Resolution}
\newacronym{EDSR}{EDSR}{Enhanced Deep Super-Resolution}
\newacronym{MDSR}{MDSR}{Multi-scale Deep Super-Resolution}
\newacronym{LR2}{LR}{learning-rate}
\newacronym{ResNet}{ResNet}{Residual Network}
\newacronym{SRResNet}{SRResNet}{Super-Resolution Residual Network}
\newacronym{GAN}{GAN}{Generative Adversarial Network}
\newacronym{SRGAN}{SRGAN}{\acrlong{SR} \acrshort{GAN}}
\newacronym{StackGAN}{StackGAN}{Stacked \acrshort{GAN}}
\newacronym{CBN}{CBN}{Cascaded Bi-Network}
\newacronym{UR-DGN}{UR-DGN}{Ultra-Resolution by Discriminative Generative Networks}
\newacronym{AWGN}{AWGN}{Additive White Gaussian Noise}
\newacronym{SRMD}{SRMD}{Super-Resolution network for Multiple Degradations}
\newacronym{SRMDNF}{SRMDNF}{\acrshort{SRMD} Noise-Free}
\newacronym{TDAE}{TDAE}{Transformative Discriminative Autoencoder}
\newacronym{ESPCN}{ESPCN}{Efficient Sub-pixel Convolutional Neural Network}
\newacronym{VESPCN}{VESPCN}{Video \acrshort{ESPCN}}
\newacronym{AE}{AE}{Autoencoder}
\newacronym{VAE}{VAE}{Variational \acrshort{AE}}
\newacronym{RBM}{RBM}{Restricted Boltzmann Machine}
\newacronym{DBN}{DBN}{Deep Belief Network}
\newacronym{LSTM}{LSTM}{Long-Short Term Memory}
\newacronym{LM-CSS}{LM-CSS}{Linear Models of Coupled Sparse Support}
\newacronym{LINE}{LINE}{Locality-constrained Iterative Neighbour Embedding}
\newacronym{SPADE}{SPADE}{SPatially-Adaptive (DE)normalization}
\newacronym{VSRnet}{VSRnet}{Video \acrlong{SR} network}
\newacronym{FRVSR}{FRVSR}{Frame-recurrent video \acrlong{SR}}
\newacronym{3DSRnet}{3DSRnet}{3D \acrlong{CNN} for video \acrlong{SR}}
\newacronym{LPIPS}{LPIPS}{Learned Perceptual Image Patch Similarity}
\newacronym{FID}{FID}{Fr\'echet Inception Distance}
\newacronym{PWF}{PWF}{Pixel-Wise network Fusion}
\newacronym{PNF}{PNF}{Progressive Network Fusion}
\newacronym{CNF}{CNF}{Content-wise Network Fusion}
\newacronym{DRCN}{DRCN}{Deeply-Recursive Convolutional Network}
\newacronym{CARN}{CARN}{Cascading Residual Network}
\newacronym{CinCGAN}{CinCGAN}{Cycle-in-Cycle \acrshort{GAN}}
\newacronym{ESRGAN}{ESRGAN}{Enhanced \acrshort{SRGAN}}
\newacronym{Real-ESRGAN}{Real-ESRGAN}{Real \acrshort{ESRGAN}}
\newacronym{RRDB}{RRDB}{Residual-in-Residual Dense Block}
\newacronym{RCAN}{RCAN}{Residual Channel Attention Network}
\newacronym{LapSRN}{LapSRN}{Laplacian Pyramid Super-Resolution Network}
\newacronym{VDSR}{VDSR}{Very Deep \acrshort{SR}}
\newacronym{DRRN}{DRRN}{Deep Recursive Residual Network}
\newacronym{BSRGAN}{BSRGAN}{Blind \acrshort{SR} \acrshort{GAN}}
\newacronym{BSRNet}{BSRNet}{Blind \acrshort{SR} Network}
\newacronym{DPSR}{DPSR}{Deep Plug-and-Play \acrshort{SR}}
\newacronym{UDVD}{UDVD}{Unified Dynamic Convolutional Network for Variational Degradations}
\newacronym{IKC}{IKC}{Iterative Kernel Correction}
\newacronym{DAN}{DAN}{Deep Alternating Network}
\newacronym{KernelGAN}{KernelGAN}{Kernel \acrshort{GAN}}
\newacronym{ZSSR}{ZSSR}{Zero-Shot \acrshort{SR}}
\newacronym{FSSR}{FSSR}{Frequency Separation for real-world \acrshort{SR}}
\newacronym{KOALAnet}{KOALAnet}{Kernel-Oriented Adaptive Local Adjustment network}
\newacronym{moco}{MoCo}{Momentum Contrast}
\newacronym{supmoco}{SupMoCo}{Supervised \acrshort{moco}}
\newacronym{supcon}{SupCon}{Supervised Contrastive}
\newacronym{simclr}{SimCLR}{Simple framework for Contrastive Learning  of visual Representations}
\newacronym{CMDSR}{CMDSR}{Conditional hyper-network framework for \acrshort{SR} with Multiple Degradations}
\newacronym{DASR}{DASR}{Degradation-Aware \acrshort{SR}}
\newacronym{IDMBSR}{IDMBSR}{Implicit Degradation Modelling Blind \acrshort{SR}}
\newacronym{SAN}{SAN}{Second-Order Attention Network}
\newacronym{HAN}{HAN}{Holistic Attention Network}
\newacronym{FAN}{FAN}{Face Alignment Network}
\newacronym{Super-FAN}{Super-FAN}{Super-\protect\glsunset{FAN}\acrlong{FAN}}
\newacronym{SFT}{SFT}{Spatial Feature Transform}
\newacronym{SFTMD}{SFTMD}{\protect\glsunset{SFT}\acrlong{SFT} Multiple Degradations}
\newacronym{moesr}{MoESR}{Mixture of Experts \acrshort{SR}}
\newacronym{DASR2}{DASR}{Domain-distance Aware \acrshort{SR}}
\newacronym{DGFEM}{DGFEM}{Degradation-Guided Feature Extraction Module}
\newacronym{EDT}{EDT}{Encoder-Decoder-based Transformer}
\newacronym{SwinIR}{SwinIR}{Swin Image Restoration}
\newacronym{ESRT}{ESRT}{Efficient \acrshort{SR} Transformer}
\newacronym{HAT}{HAT}{Hybrid Attention Transformer} 
\newacronym{ELAN}{ELAN}{Efficient Long-Range Attention Network}
 
\newacronym{GPU}{GPU}{Graphical Processing Unit}
\newacronym{CPU}{CPU}{Central Processing Unit}
\newacronym{cudnn}{cuDNN}{NVIDIA CUDA Deep Neural Network}

\newacronym{NN2}{NN}{Neural Network}
\newacronym{CNN}{CNN}{Convolutional \protect\glsunset{NN2}\acrlong{NN2}}
\newacronym{DCNN}{DCNN}{Deep \protect\glsunset{CNN}\glsfirst{CNN}\protect\glsreset{CNN}}
\newacronym{DL}{DL}{Deep Learning}
\newacronym{RELU}{ReLU}{Rectified Linear Unit}
\newacronym{PRELU}{PReLU}{Parametric \acrshort{RELU}}
\newacronym{ANN}{ANN}{Artificial Neural Network}
\newacronym{DNN}{DNN}{Deep \protect\glsunset{NN2}\acrlong{NN2}}

\newacronym{SNR}{SNR}{Signal-to-Noise Ratio}
\newacronym{PSNR}{PSNR}{Peak Signal-to-Noise Ratio}
\newacronym{MAD}{MAD}{Mean Absolute Difference}

\newacronym{TSNE}{t-SNE}{t-Distributed Stochastic Neighbour Embedding}

\newacronym{MSE}{MSE}{Mean Square Error}
\newacronym{SSIM}{SSIM}{Structural SIMilarity index}
\newacronym{IQA}{IQA}{Image Quality Assessment}
\newacronym{VQA}{VQA}{Video Quality Assessment}
\newacronym{FR}{FR}{Full-Reference}
\newacronym{RR}{RR}{Reduced-Reference}
\newacronym{NR}{NR}{No-Reference}

\newacronym{CCTV}{CCTV}{Closed-Circuit Television}
\newacronym{I-frame}{I-frame}{Intra-coded frame}
\newacronym{P-frame}{P-frame}{Predicted frame}
\newacronym{B-frame}{B-frame}{Bi-directional predicted frame}
\newacronym{PSF}{PSF}{Point Spread Function}
\newacronym{ISP}{ISP}{Image Signal Processing}
\newacronym{PCA}{PCA}{Principal Component Analysis}
\newacronym{MAP}{MAP}{Maximum A Posteriori}
\newacronym{NLP}{NLP}{Natural Language Processing}

\newacronym{DA}{DA}{Degradation-aware}
\newacronym{DGFMB}{DGFMB}{Degradation-Guided Feature Modulation Block}
\newacronym{MA}{MA}{Meta-Attention}
\newacronym{MLP}{MLP}{Multi-Layer Perceptron}

\glsdisablehyper 
\newcolumntype{P}[1]{>{\centering\arraybackslash}p{#1}}

\newcolumntype{L}[1]{>{\raggedright\let\newline\\\arraybackslash\hspace{0pt}}m{#1}}
\newcolumntype{C}[1]{>{\centering\let\newline\\\arraybackslash\hspace{0pt}}m{#1}}
\newcolumntype{R}[1]{>{\raggedleft\let\newline\\\arraybackslash\hspace{0pt}}m{#1}}

\Title{The Best of Both Worlds: a Framework for Combining Degradation Prediction with High Performance Super-Resolution Networks}

\TitleCitation{Title}

\Author{Matthew Aquilina $^{1,2}$\orcidA{}*, Keith George Ciantar $^{3}$\orcidB{}, Christian Galea $^{1}$\orcidC{}, Kenneth P. Camilleri $^{4}$\orcidD{}, Reuben A. Farrugia $^{1}$\orcidE{} and John Abela $^{5}$\orcidF{}}

\AuthorNames{Firstname Lastname, Firstname Lastname and Firstname Lastname}

\AuthorCitation{Lastname, F.; Lastname, F.; Lastname, F.}

\address{%
$^{1}$ \quad Department of Communications \& Computer Engineering, Faculty of ICT,  University of Malta, Msida, Malta; matthew.aquilina@um.edu.mt; christian.p.galea@um.edu.mt; reuben.farrugia@um.edu.mt\\
$^{2}$ \quad Deanery of Molecular, Genetic \& Population Health Sciences, University of Edinburgh, Edinburgh, Scotland, UK; m.aquilina@ed.ac.uk\\
$^{3}$ \quad Ascent, 90/3, Alpha Centre, Tarxien Road, Luqa, Malta; keith.ciantar@ascent.io\\
$^{4}$ \quad Department of Systems \& Control Engineering, Faculty of Engineering, University of Malta, Msida, Malta; kenneth.camilleri@um.edu.mt\\
$^{5}$ \quad Department of Computer Information Systems, Faculty of ICT,  University of Malta, Msida, Malta; john.abela@um.edu.mt\\
}
\corres{Correspondence: matthew.aquilina@um.edu.mt}

\abstract{To date, the best-performing blind super-resolution (SR) techniques follow one of two paradigms: A) generate and train a standard SR network on synthetic low-resolution – high-resolution (LR – HR) pairs or B) attempt to predict the degradations an LR image has suffered and use these to inform a customised SR network. Despite significant progress, subscribers to the former miss out on useful degradation information that could be used to improve the SR process.  On the other hand, followers of the latter rely on weaker SR networks, which are significantly outperformed by the latest architectural advancements.  In this work, we present a framework for combining any blind SR prediction mechanism with any deep SR network, using a metadata insertion block to insert prediction vectors into SR network feature maps.  Through comprehensive testing, we prove that state-of-the-art contrastive and iterative prediction schemes can be successfully combined with high-performance SR networks such as RCAN and HAN within our framework.  We show that our hybrid models consistently achieve stronger SR performance than both their non-blind and blind counterparts.  Furthermore, we demonstrate our framework’s robustness by predicting degradations and super-resolving images from a complex pipeline of blurring, noise and compression.  Our framework is available at: \url{https://github.com/um-dsrg/RUMpy}.}

\keyword{blind super-resolution; meta-attention; degradation prediction; metadata fusion; iterative prediction; contrastive learning; deep learning} 

\begin{document}



\section{Introduction}\label{sec:intro}

\gls{SR} is the process by which a \gls{LR} image is upscaled, with the aim of enhancing both the image's quality and level of detail.  This operation enables the exposure of previously-hidden information which can then subsequently be used to improve the performance of any tasks depending on the super-resolved image. \gls{SR} is thus highly desirable in a vast number of important applications such as medical imaging \cite{Gupta20,Ahmad22}, remote sensing \cite{Haut18,Wang22}, and in the identification of criminals depicted in \gls{CCTV} cameras during forensic investigations \cite{Chen22, Rasti16}. 

\gls{SISR} is typically formulated as the restoration of \gls{HR} images that have been bicubically downsampled or blurred and downsampled.  On these types of \gls{LR} images, \gls{SOTA} \gls{SR} models can achieve extremely high performance, either by optimizing for high pixel fidelity to the \gls{HR} image \cite{rcan, san, han, Vella21, swinir, elan}, or by improving perceptual quality \cite{Ledig17, Wang19, Wang21}.  However, real-world images are often affected by additional factors such as sensor noise, complex blurring, and compression \cite{Zhang22, Liu21, Zhang21}, which further deteriorate the image content and make the restoration process significantly more difficult. Moreover, many \gls{SR} methods are trained on synthetically generated pairwise \gls{LR}/\gls{HR} images which only model a subset of the potential degradations encountered in real-world imaging systems \cite{Jiang21, Liu21}. As a result, the domain gap between synthetic and realistic data often causes such \gls{SR} methods to perform poorly in the real world, hindering their practical use \cite{Kohler20, Chen22, Liu21}. 

The field of blind \gls{SR} is actively attempting to design techniques for image restoration which can deal with more realistic images containing unknown and complex degradations \cite{Liu21}. These methods often break down the problem by first estimating the degradations within an image, after which this prediction is used to improve the performance of an associated \gls{SR} model.  Prediction systems can range from the explicit, such as estimating the shape/size of a specific blur kernel, to the implicit, such as the abstract representation of a degradation within a \gls{DNN} \cite{Liu21}.  In the explicit domain, significant progress has been made in improving the accuracy and reliability of the degradation parameter estimation process.  Recent mechanisms based on iterative improvement \cite{Jinjin19, DAN} and contrastive learning \cite{DASR21, IDMBSR} have been capable of predicting the shape, size and noise of applied blur kernels with little to no error.  However, such methods then go on to apply their prediction mechanisms with \gls{SR} architectures that are smaller and less sophisticated than those used for \gls{SOTA} non-blind \gls{SR}.  

In this work, we investigate how larger \gls{SR} architectures could be modified to benefit from prominent blind degradation prediction systems. 
We use a \textit{metadata insertion block} to link the prediction and \gls{SR} mechanisms, an operation which interfaces degradation vectors with \gls{SR} network feature maps.  We implement a variety of \gls{SR} architectures and integrate these with the latest techniques for contrastive learning and iterative degradation prediction.  Our results show that by using just a single \gls{MA} layer \cite{Aquilina21}, high-performance \gls{SR} models such as the \gls{RCAN} \cite{rcan} and \gls{HAN} \cite{han} can be infused with degradation information to yield \gls{SR} results which outperform those of the original blind \gls{SR} networks trained under the same conditions.  

We further extend our premise by performing blind degradation prediction and \gls{SR} on images with blurring, noise and compression, constituting a significantly more complex degradation pipeline than that studied to-date by other prediction networks \cite{Jinjin19, DAN, DASR21, IDMBSR, Luo22}.  We show that, even on such a difficult dataset, our framework is still capable of generating improved \gls{SR} performance when combined with a suitable degradation prediction system.

The main contributions of this paper are thus as follows:
\begin{enumerate}
    \item A framework for the integration of degradation prediction systems into \gls{SOTA} non-blind \gls{SR} networks.
    \item A comprehensive evaluation of different methods for the insertion of blur kernel metadata into \gls{CNN} \gls{SR} networks.  Our results show that simple metadata insertion blocks, such as \gls{MA}, can match the performance of more complex metadata insertion systems when used in conjunction with large \gls{SR} networks.
    \item Blind \gls{SR} results using a combination of non-blind \gls{SR} networks and \gls{SOTA} degradation prediction systems. These hybrid models show improved performance over both the original \gls{SR} network and the original blind prediction system.
    \item A thorough comparison of unsupervised, semi-supervised and supervised degradation prediction methods for both simple and complex degradation pipelines.
    \item The successful application of combined blind degradation prediction and \gls{SR} of images degraded with a complex pipeline involving multiple types of noise, blurring and compression.
\end{enumerate}

The rest of this paper is organised as follows: \Cref{sec:literature} provides an overview of related work on general and blind \gls{SR}, including the methods selected for our framework. \Cref{sec:method} follows up with a detailed description of our proposed methodology for combining degradation prediction methods with \gls{SOTA} \gls{SISR} architectures. Our framework implementation details, evaluation protocol, degradation prediction and \gls{SR} results are presented and discussed in \Cref{sec:experiments}. Finally, \Cref{sec:conc} provides concluding remarks and potential areas for further exploration.

\section{Related Work}\label{sec:literature}
Numerous methods have been proposed for \gls{SR}, from the seminal \gls{SRCNN} \cite{srcnn} to more advanced networks such as \gls{RCAN} \cite{rcan}, \gls{HAN} \cite{han}, \gls{SAN} \cite{san}, \gls{SRGAN} \cite{Ledig17}, and \gls{ESRGAN} \cite{Wang19} among others. Domain-specific methods have also been implemented, such as those geared for the super-resolution of face images (including \gls{Super-FAN} \cite{Bulat18} and the methods proposed in \cite{Rasti16, Huang19, Xin18, Yu18, Huang17, Ze18, Cao17, Xin18b}), and satellite imagery \cite{Haut18,Wang22,Nguyen22} among others.

Given that the proposed framework combines techniques designed for both general-purpose \gls{SR} and blind \gls{SR}, an overview of popular and \gls{SOTA} networks for both methodologies will be provided in this section. 
Most approaches derive a \gls{LR} image from a \gls{HR} image using a \textit{degradation model}, which formulates how this process is performed and the relationship between the \gls{LR} and \gls{HR} images. Hence, an overview of common degradation models, including the one used as the basis for the proposed \gls{SR} framework, is first provided and discussed. 

\subsection{Degradation Models}
\label{sec:deg}
Numerous works in the literature have focused on the degradation models considered in the \gls{SR} process, which define how \gls{HR} images are degraded to yield \gls{LR} images. However, the formation of a \gls{LR} image $I^{LR}$ can be generally expressed by the application of a function $f$ on the \gls{HR} image ${I^{HR}}$, as follows: 

\begin{equation}
    I^{LR} = f(I^{HR}, \theta_D)
\end{equation}

where $\theta_D$ is the set of degradation parameters, which are unknown in practice. The function $f$ can be expanded to consider the general set of degradations applied to ${I^{HR}}$, yielding the `classical' degradation model as follows \cite{Jinjin19, Chen22, Jiang21, Luo21, DAN, Liu21, srmd, Zhang21, Xiao20, Yue22, MoESR22, Kang22}:

\begin{equation}
\label{eq:deg1}
    I^{LR} = (I^{HR} \otimes {k})\downarrow_s + {n}
\end{equation}

where $\otimes$ represents the convolution operation, 
${k}$ is a kernel (typically a Gaussian blurring kernel, but it can also represent other functions such as the \gls{PSF}), 
${n}$ represents additive noise, 
and $\downarrow_s$ is a downscaling operation (typically assumed to be bicubic downsampling \cite{Liu21}) with scale factor $s$.  However, this model has been criticised for being too simplistic and unable to generalise well to more complex degradations that are found in real-world images, thereby causing substantial performance losses when \gls{SR} methods based on this degradation model are applied to non-synthetic images \cite{Zhang21, Zhang22, srmd}. 
More complex and realistic degradation types have thus been considered, such as compression (which is typically signal-dependent and non-uniform, in contrast to the other degradations considered \cite{Zhang21}), to yield a more general degradation model \cite{Liu21, Liu20, Zhang22}: 

\begin{equation}
\label{eq:deg2}
    I^{LR} = ((I^{HR} \otimes {k})\downarrow_s + {n})_{C}
\end{equation}

where $C$ is a compression scheme such as JPEG. The aim of \gls{SR} is then to solve the inverse function of $f$, denoted by $f^{-1}$, which can be applied on the \gls{LR} image ($I^{LR}$) to reverse the degradation process and yield an image $\hat{I}^{HR}$ which approximates the original image ($I^{HR}$): 

\begin{equation}
    \hat{I}^{HR} = f^{-1}(I^{LR}, \theta_R) \approx I^{HR}
\end{equation}

where $\theta_R$ represents the parameter set defining the reconstruction process. This degradation model forms the basis of the proposed \gls{SR} framework. 

Other works have extended the general model in Equation \ref{eq:deg2} to more complex cases.  In \cite{Wang21}, training pairs are synthesised using a `high-order' degradation process where the degradation model is applied more than once.   
The authors of \cite{Zhang21} proposed a practical degradation model to train the \gls{ESRGAN}-based \acrshort{BSRNet} and \acrshort{BSRGAN} models \cite{Zhang21}, which consider multiple Gaussian blur kernels, downscaling operators, noise levels modelled by \gls{AWGN}, processed camera sensor noise types, and quality factors of JPEG compression, with random shuffling of the order in which the degradations are applied.

Counter-arguments to these complex models have also been made. For instance, the authors of \cite{Zhang22} argued that `practical' degradation models as proposed in \cite{Zhang21, Wang21} (so called because a wide variety of degradations are considered, similar to practical real-world applications) may achieve promising results on complex degradations but then ignore easier edge cases, namely combinations of degradation subsets. A gated degradation model is thus proposed, which randomly selects the base degradations to be applied. Given that the magnitudes of some degradations in the proposed framework may be reduced to the point where they are practically negligible, the degradation model forming the basis of this work can be said to approximate this gated mechanism. 

\subsection{Non-blind \gls{SR} Methods}
Most methods proposed for \gls{SR} have tended to focus on the case where degradations are assumed to be known, either by designing models for specific degradations or by designing approaches that are able to use supplementary information about the degradations afflicting the image.
However, this information is not estimated or derived from the corrupted image in any way. This limits the use of such methods in the real world where degradations are highly variable, in terms of both their type and magnitude. Despite these limitations, non-blind \gls{SR} methods have served an important role in enabling more rapid development of new techniques on what is arguably a simpler case of \gls{SR}.

Notable non-blind \gls{SR} methods include the \gls{SRCNN} network, considered to be pioneering work in using deep learning and \glspl{CNN} for the task of \gls{SR}. However, it only consists of three layers and requires the \gls{LR} image to first be upsampled using bicubic interpolation, leading to this method being outperformed by most modern approaches.

To facilitate the training of a large number of \gls{CNN} layers, the \gls{ResNet} architecture proposed in \cite{He16} introduced \textit{skip connections} to directly feed feature maps at any level of the network to deeper layers, a process corresponding to the identity function which deep networks find hard to learn.  This counteracted the problem of vanishing gradients apparent in classical deep \gls{CNN} networks, allowing the authors to expand their network size without impacting training performance. \gls{ResNet} was extended to \gls{SR} in \cite{Ledig17}, to create the \gls{SRResNet} approach that was also used as the basis for a \gls{GAN}-based approach termed \gls{SRGAN}. 

\gls{SRGAN} was extended in \cite{Wang19} to yield \gls{ESRGAN}, which included the introduction of a modified adversarial loss to determine the relative `realness' of an image, rather than simply whether the generated image is real or fake. \gls{ESRGAN} also introduced a modified VGG-based perceptual loss.  This uses feature maps extracted from the VGG residual blocks right before the activation layers to reduce sparsity and better supervise brightness consistency and texture recovery. \gls{ESRGAN} was further extended in \cite{Wang21} to yield \gls{Real-ESRGAN}, where the focus was on the implementation of a `high-order' degradation process that allowed the application of the degradation model more than once (as is typically done in other works). 

\gls{EDSR} \cite{edsr} was also based on \gls{ResNet} and incorporated observations noted in previous works such as \gls{SRResNet}, along with other novel contributions that had a large impact on subsequent \gls{CNN}-based \gls{SR} models.  These included the removal of batch normalisation layers to disable restriction of feature values and reduce the memory usage during training that in turn allowed for a greater number of layers and filters to be used. 

The \gls{RCAN} approach proposed in \cite{rcan} is composed of `residual groups' that each contain a number of `channel attention blocks', along with `long' and `short' skip connections to enable the training of very deep \glspl{CNN}. The channel attention blocks allow for the assignment of different levels of importance of low-frequency information across feature map channels. \gls{RCAN} remains one of the top-performing \gls{SR} methods, and was also shown in \cite{rcan} to be beneficial for object recognition after higher accuracies were attained in comparison to images upsampled by other methods. 
The concept of attention introduced by \gls{RCAN} was developed further by methods including \gls{SAN} \cite{san} and \gls{HAN} \cite{han}, where techniques such as channel-wise feature re-scaling and modelling of any inter-dependencies among channels and layers were proposed.

Recently, vision transformers applied to \gls{SR} have also been proposed, such as the \gls{EDT} \cite{Li21}, \gls{ESRT} \cite{Lu22}, and the \gls{SwinIR} \cite{swinir} approach that is based on the Swin Transformer \cite{Ze21}. 
Approaches such as \gls{ELAN} \cite{elan} and \gls{HAT} \cite{hat}, which attempt to combine \gls{CNN} and transformer architectures, have also been proposed with further improvements in \gls{SR} performance. 

A more in-depth review of generic non-blind \gls{SR} methods may be found in \cite{Ha19, Wang21_review}.

\subsection{Blind \gls{SR} Methods}

Although numerous \gls{SR} methods have been proposed, a substantial number of approaches tend to employ the classical degradation model in \Cref{eq:deg1}. Besides not being quite reflective of real-world degradations (as discussed in \Cref{sec:deg}), a substantial number of approaches also assume that the degradations afflicting an image are known, which is largely not the case. Consequently, such approaches tend to exhibit noticeable performance degradation on ``in-the-wild'' images. 

Blind \gls{SR} methods have thus been designed for better robustness when faced with such difficult and unknown degradations, making them more suitable for real-world applications. There exist several types of blind-\gls{SR} methods, based on the type of data used and how they are modelled \cite{Liu21}. 
An overview of the various types of approaches and representative methods will now be provided.

\subsubsection{Approaches Utilising Supplementary Attributes for \gls{SR}} 
Early work focused on the development of methods where ground-truth information on degradations is supplied directly, with the focus then on how this degradation information can be best utilised (as opposed to non-blind \gls{SR} methods which do not use any supplementary information). 
Approaches of this kind generally consider the classical degradation model \cite{Liu21}. 

Notable methods incorporating metadata information in networks include \gls{SRMD} \cite{Zhang18b}, \gls{UDVD} \cite{Xu20}, the \gls{DPSR} framework, and the approach in \cite{Cornillere19}.  Each of these approaches showed that \gls{SR} networks could make use of this degradation information, improving their \gls{SR} performance as a result. Frameworks to enable the extension of existing non-blind \gls{SR} methods to use degradation information have also been proposed, such as the `meta-attention' approach in \cite{Aquilina21} and \gls{CMDSR} proposed in \cite{CMDSR}.

These methods clearly show the plausibility of improving \gls{SR} performance with degradation metadata, although some means of generating or predicting relevant degradation information needs to be present for these methods to function correctly.  However, such methods are highly reliant on the quality of the degradation information input to the networks, which is not a trivial task. Moreover, any deviations in the estimated inputs lead to kernel mismatches and can thus be detrimental to \gls{SR} performance \cite{Liu21,DAN,Luo21}. 

\subsubsection{Iterative Kernel Estimation Methods}

Blur kernel estimation during the \gls{SR} process is one of the most common blind \gls{SR} prediction tasks, and alleviates the problem of kernel mismatches present in methods such as \gls{SRMD} as described above. Often, iterative mechanisms are applied for direct kernel estimation. One such method is \gls{IKC} \cite{Jinjin19}, which leverages the observation that kernel mismatch tends to produce regular patterns by estimating the degradation kernel and correcting it in an iterative fashion using a corrector network. In this way, an acceptable result is progressively approached. The authors of \cite{Jinjin19} also proposed a non-blind \gls{SR} network, \gls{SFTMD}, which was shown to outperform existing methods such as \gls{SRMD} for inserting blur kernel metadata into the \gls{SR} process.

The \gls{DAN} \cite{DAN} method (also known as \gls{DAN}v1) and its updated version \gls{DAN}v2 \cite{Luo21} build upon the \gls{IKC} approach, by combining the \gls{SR} and kernel corrector networks within a single end-to-end trainable network. The corrector was also modified to use the \gls{LR} input conditioned on intermediate super-resolved images, instead of conditioning the super-resolved images on the estimated kernel as done in \gls{IKC}. 
The \gls{KOALAnet} \cite{KOALAnet} is able to adapt to spatially-variant characteristics within an image, which allows a distinction to be made between blur caused by undesirable effects, and between blur introduced intentionally for aesthetic purposes (e.g. Bokeh effect). 
However, such methods also exhibit poor performance when evaluated on images having different degradations than those used to train the model, given that they still rely on kernel estimation. 

\subsubsection{Training \gls{SR} Models on a Single Image}
\label{single_image}

Another group of methods such as \acrshort{KernelGAN} \cite{Kligler19} and \gls{ZSSR} \cite{Assaf18} use intra- and inter-scale recurrence of patches, based on the internal statistics of natural images, to construct an individual model for each input \gls{LR} image. Hence, the data used for training is that which is present internally within the image being super-resolved, circumventing the need to use an external dataset of images. 

Such methods tend to assume that a downscaled version of a patch within a \gls{LR} image should have a similar distribution to the patch in the original \gls{LR} image. However, the assumption of recurring patches within and across scales may not hold true for all images (such as those containing a wide variety of content) \cite{Liu21}. 

\subsubsection{Implicit Degradation Modelling}

Modelling an explicit combination of multiple degradation types can be a very complex task on in-the-wild images. 
Hence, approaches have also attempted to implicitly model the degradation process by comparing the data distribution of real-world \gls{LR} image sets with synthetically created `clean' (containing limited or no degradations) datasets \cite{Liu21}. Methods are typically based on \glspl{GAN}, such as \gls{CinCGAN} \cite{Yuan18} and the approaches in \cite{Zhou20, Maeda20}, and do not require a \gls{HR} reference for training. 

One of the drawbacks of this type of method is that they tend to require vast amounts of data, which may not always be available. Some approaches, such as Degradation \gls{GAN} \cite{Bulat18b} and \gls{FSSR} \cite{Fritsche19}, attempt to counteract this issue by learning the \gls{HR} to \gls{LR} degradation process, to enable the generation of realistic \gls{LR} samples that can be used during the training of the \gls{SR} model. 
However, most models designed for implicit degradation modelling use \glspl{GAN} which are known to be hard to train and can introduce fake textures or artefacts that can be detrimental for some real-world applications \cite{Liu21}.  Developing an implicit modelling approach which can model degradations from just a single image (similar to the approaches in \Cref{single_image}) could significantly help to reduce these methods' drawbacks \cite{Liu21}.

\subsubsection{Contrastive Learning}
\label{sec:cl}

In the image classification domain, \glspl{DNN} are known to be highly capable of learning invariant representations, enabling the construction of good classifiers \cite{supmoco21}. However, it has been argued that \glspl{DNN} are actually too eager to learn invariances \cite{supmoco21}. This is because they often learn only the features necessary to discriminate between classes but then fail to generalise well to new unseen classes in a supervised setting, in what is known as ``supervision collapse'' \cite{supmoco21, Doersch20}. 
Indeed, the ubiquitous cross-entropy loss used to train supervised deep classifier models has received criticism over several shortcomings \cite{Khosla20}, such as its sensitivity to noisy labels \cite{Zhang18c, Sukhbaatar14} and the possibility of poor margins \cite{Gamaleldin18,Cao19,Weiyang16}.

\textit{Contrastive learning} techniques, mostly developed in the \gls{NLP} domain, have recently seen a resurgence and have driven significant advances in self-supervised representation learning in an attempt to mitigate these issues \cite{Khosla20}. 
Contrastive learning is a self-supervised approach where models are trained by comparing and contrasting `positive' image pairs with `negative' pairs \cite{Ting20}. Positive images can be easily created by applying augmentations to a source image (e.g. flipping, rotations, colour jitter etc.). In the \gls{SR} domain, positive samples are typically patches extracted from within the same image while crops taken from other images are labelled as negative examples \cite{moco20,Ting20,DASR21}. 

One such contrastive learning-based approach is \gls{moco}, proposed in \cite{moco20} for the tasks of object detection, classification, and segmentation. \gls{moco} employs a large queue of data samples to enable the use of a dictionary (containing samples observed in preceding mini-batches) which is much larger than the mini-batch size.
However, since a large dictionary also makes it intractable to update the network parameters using back-propagation, a momentum update which tightly controls the parameters' rate of change is also proposed.
\gls{moco} was extended in \cite{chen2020improved} to yield \gls{moco}v2, based on design improvements proposed for \acrshort{simclr} \cite{Ting20}. The two main modifications constitute replacing the fully-connected layer at the head of the network with an \gls{MLP} head, and the inclusion of blur augmentation. 

\gls{supmoco} \cite{supmoco21} was also proposed as an extension of \gls{moco}, whereby class labels are additionally utilised to enable intra-class variations to be learnt whilst retaining knowledge on distinctive features acquired by the self-supervised components. 
\gls{supmoco} was shown to outperform the \gls{supcon} approach \cite{Khosla20}, which also applied supervision to \acrshort{simclr} \cite{Ting20}.

Such self-supervised methods have seen limited use in the \gls{SR} domain thus far. However, the promising performance demonstrated in other domains could encourage further research and development in the \gls{SR} arena. 

The \gls{DASR} network \cite{DASR21} was one of the first \gls{SR} networks to use contrastive learning for blind \gls{SR}.  In their approach, the authors derive an unsupervised content-invariant degradation representation in the latent feature space where the mutual information among all samples is maximised. 

 In \cite{IDMBSR}, \gls{IDMBSR} considers the degrees of difference in degradation between a query image and negative exemplars, in order to determine the amount of `push' to exert. Specifically, the greater the difference between a query and a negative example, the greater the push. In this way, degradation information is used as weak supervision of the representation learning to improve the network's ability to characterise image degradations. 

Contrastive learning frameworks have also been proposed for the super-resolution of remote sensing images \cite{Kang22}. Specifically, degradation representations are acquired in an unsupervised manner, which are used to improve the high-frequency details in a \gls{DGFEM} that is inspired by the attention mechanism \cite{Jie18}. 
Feature fusion is also performed to better exploit low-level features that have sharper textures than those obtained using features extracted in deeper layers. This is because the latter tend to learn semantic information at the expense of yielding images having blurry edges. 

\subsubsection{Other Methodologies}

The mechanisms discussed in this section are some of the most prominent in the blind \gls{SR} literature.  However, many other modalities exist which do not neatly fall into any single category. One such method is \gls{moesr} \cite{MoESR22}, where a panel of expert predictors is used to help optimise the prediction system for blur kernels of different shapes and sizes.  A comprehensive overview of the state of blind \gls{SR} research can be found in \cite{Liu21}.

Given the prominence of iterative and contrastive methods in blind \gls{SR}, a number of these mechanisms were selected and implemented within the proposed framework.  The principles of \gls{supmoco} were also applied to construct a more controllable self-supervised contrastive learning function (ref. \Cref{supmoco_description}).  However, in principle, any degradation estimation system could be coupled to any \gls{SR} model, using the framework described in the rest of this paper.

\section{Methodology}\label{sec:method}
\subsection{Degradation Model}
For all our analyses, we follow the degradation model described in Equation \ref{eq:deg2}, which stipulates that most realistic degradations will involve some amount of blurring, downsampling, noise addition and compression.  For the \gls{LR}/\gls{HR} pairs used for training and testing, a variety of different operations across each type of degradation are applied (the full details of these operations are provided in Section \ref{implementation_details}).  Given that the order and variety of degradations are known in advance, the task of our degradation prediction models is significantly easier than the fully blind case with completely unknown degradations.  However, the degradation prediction principles of each model could be easily extended to more complex and realistic degradation pipelines \cite{Wang21, Zhang21} through adjustments to the degradation modelling process.  Furthermore, we show that even under these conditions, our models are still capable of dealing with real-world degraded images (\Cref{real_data_tests}).

\subsection{Framework for combining SR models with a Blind Predictor}\label{framework}
\begin{figure}[h]
    \begin{adjustwidth}{-\extralength}{0cm}
        \centering
        \includegraphics[width=\linewidth]{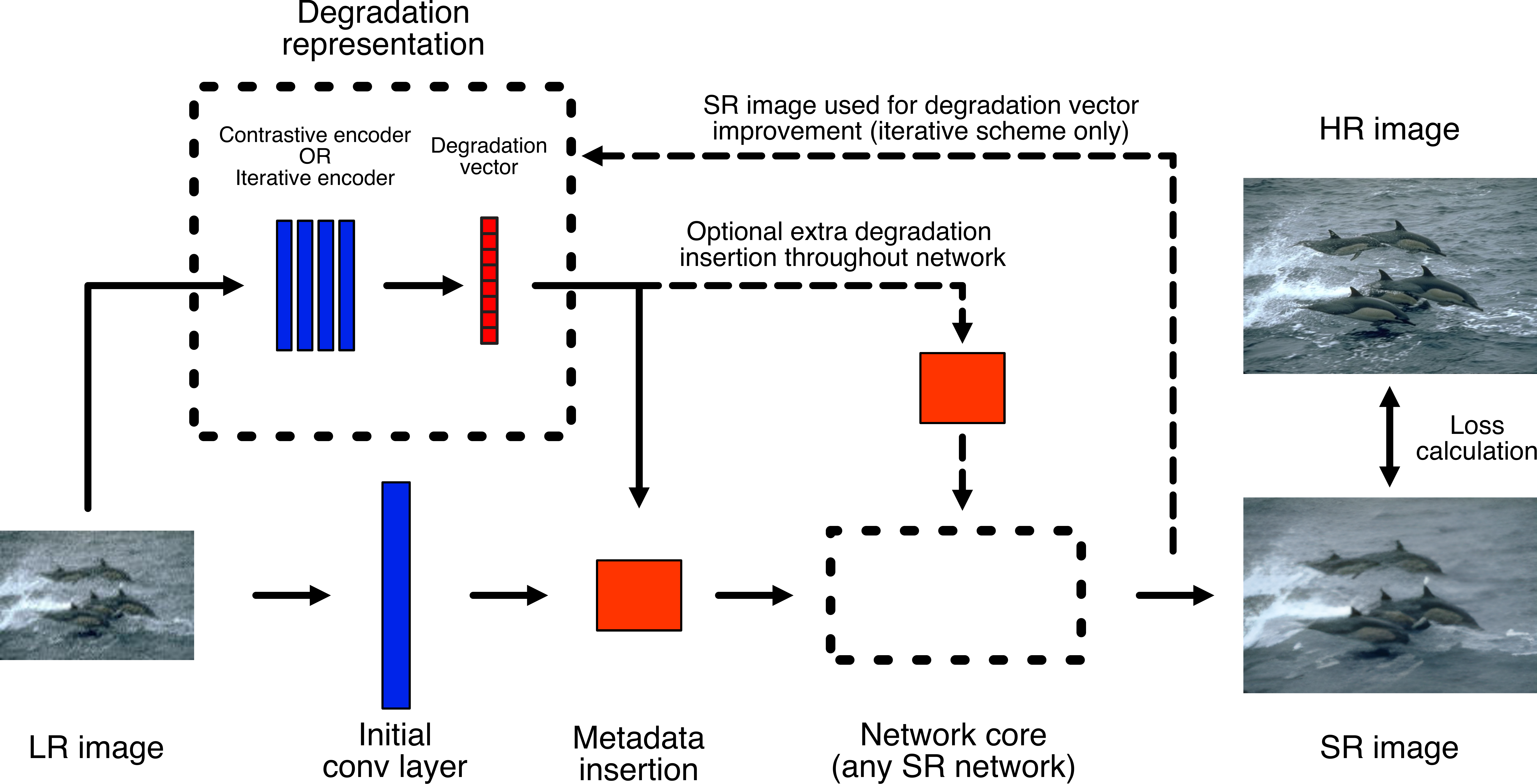}
    \end{adjustwidth}
    \caption{Proposed framework for combining blind degradation systems and SR models.  The metadata insertion block acts as the bridge between the two systems, allowing the SR model to exploit the degradation predictor to improve its performance.  Depending on the blind predictor mechanism chosen, the SR image can be fed back into the predictor to help improve its accuracy.}
    \label{fig:1}
\end{figure}

 Our proposed general framework for combining blind \gls{SR} prediction mechanisms with non-blind \gls{SR} models aims to amplify the strengths of both techniques with minimal architectural changes on either side.  In most cases, explicit blind prediction systems generate vectors to describe the degradations present in an image.  On the other hand, the vast majority of \gls{SR} networks feature convolutional layers and work in image space rather than in a vector space.  To combine the two mechanisms, the prediction model was kept separate from the \gls{SR} core and the two were bridged using a `metadata insertion' block, as shown in Figure \ref{fig:1}.  This makes it relatively simple for different prediction or \gls{SR} blocks to be swapped in and out, while keeping the overall framework unchanged.  We considered a variety of options for each of the three main components of the framework.  The metadata insertion and degradation prediction mechanisms selected and adjusted for our framework are discussed in the remaining sections of the methodology, while the chosen \gls{SR} core networks are provided in Section \ref{implementation_details}.
 
\subsection{Metadata Insertion Block}\label{metadata_insertion}
\begin{figure}[!t]
    \begin{adjustwidth}{-\extralength}{0cm}
        \centering
        \includegraphics[width=\linewidth]{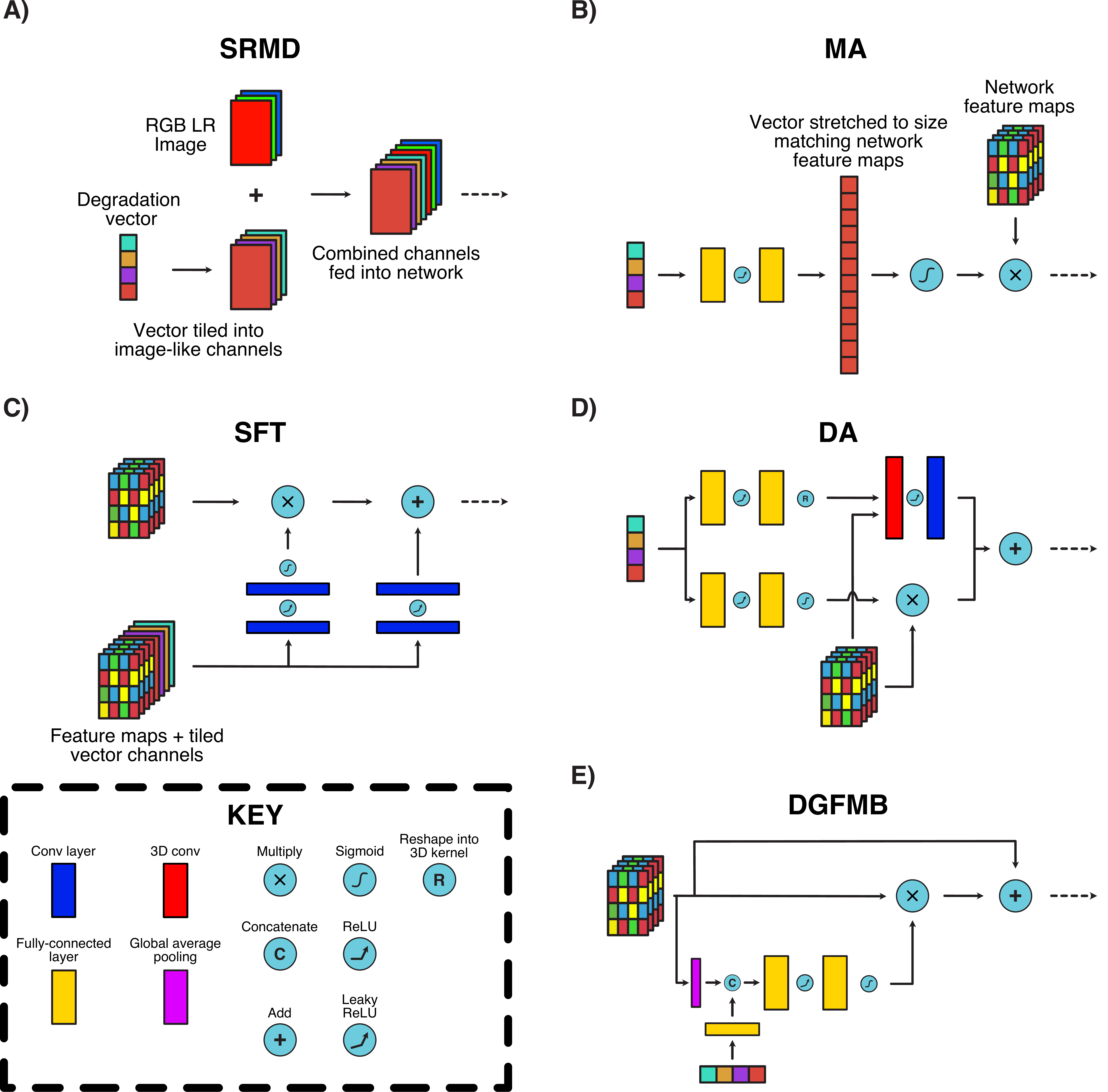}
    \end{adjustwidth}
    \caption{Metadata insertion mechanisms investigated in this paper.  The end result of each mechanism is the trainable modulation of CNN feature maps using the information present in a provided vector.  However, each mechanisms varies significantly in complexity, positioning and the components involved.}
    \label{fig:2}
\end{figure}

The metadata insertion block plays an essential role in our framework, since it converts a degradation vector into a format compatible with \gls{CNN}-based \gls{SR} networks, and ensures that this information is fully utilised throughout the \gls{SR} process.  Despite its importance, the inner workings of the mechanism are poorly understood, as \gls{CNN}-based models (\gls{SR} or otherwise) are notoriously difficult to interpret.  In fact, multiple methods for combining vectors and images within \glspl{CNN} have been proposed, which vary significantly in complexity and size without a clear winner being evident.  We selected and tested some of the most effective mechanisms in the literature within our framework, of which the following is a brief description of each (with a graphical depiction provided in Figure \ref{fig:2}):
\begin{itemize}
    \item \textbf{\gls{SRMD}-style \cite{srmd}:} One of the first methods proposed for metadata insertion in \gls{SR}, the \gls{SRMD} technique involves the transformation of vectors as additional pseudo-image channels.  Each element of the input degradation vector is expanded (by repeated tiling) into a 2D array with the same dimensions as the input \gls{LR} image. These pseudo-channels are then fed into the network along with the real image data, ensuring that all convolutional filters in the first layer have access to the degradation information.  Other variants of this method, which include directly combining the pseudo-channels with \gls{CNN} feature maps, have also been proposed \cite{DAN}.  The original \gls{SRMD} network used this method for \gls{PCA}-reduced blur kernels and noise values.  In our work, we extended this methodology to all degradation vectors considered.
    \item  \textbf{\gls{MA} \cite{Aquilina21}:} \gls{MA} is a trainable channel attention block which was proposed as a way to upgrade any \gls{CNN}-based \gls{SR} network with metadata information.  Its functionality is simple - an input vector is stretched to the same size as the number of feature maps within a target \gls{CNN} network using two fully-connected layers.  Each vector element is normalised to lie in the range $[0,1]$ and then applied to selectively amplify its corresponding \gls{CNN} feature map.  \gls{MA} was previously applied for PCA-reduced blur kernels and compression quality factors only.  We extended this mechanism to all degradation parameters considered by combining them into an input vector which is then fed into the \gls{MA} block.  The fully-connected layer sizes were then expanded as necessary to accommodate the input vector.
    \item  \textbf{\gls{SFT} \cite{Jinjin19}:}  The \gls{SFT} block is based on the \gls{SRMD} concept but with additional layers of complexity.  The input vector is also stretched into pseudo-image channels, but these are added to the feature maps within the network rather than the actual original image channels.  This combination of feature maps and pseudo-channels are then fed into two separate convolutional pathways, one of which is multiplied with the original feature maps and the other added on at the end of the block.  This mechanism is the largest (in terms of parameter count due to the number of convolutional layers) of those considered in this paper.  As with \gls{SRMD}, this method has only been applied for blur kernel and noise parameter values, and we again extended the basic concept to incorporate all degradation vectors considered.
    \item  \textbf{\gls{DA} block \cite{DASR21}:}  The \gls{DA} block was proposed in combination with a contrastive-based blind \gls{SR} mechanism for predicting blur kernel degradations.  It uses two parallel pathways, one of which amplifies feature maps in a manner similar to \gls{MA}, while the other explicitly transforms vector metadata into a 3D kernel, which is applied on the network feature maps.  This architecture is highly specialised to kernel-like degradation vectors, but could still be applicable for general degradation parameters given its dual pathways.  We extended the \gls{DA} block to all degradation vectors as we did with the \gls{MA} system.
    \item  \textbf{\gls{DGFMB} \cite{IDMBSR}:}  This block was conceived as part of another contrastive-based network, again intended for blur and noise degradations.  The main difference here is that the network feature maps are first reduced into vectors and concatenated with the degradation metadata in this form, rather than in image space.  Once concatenated, a similar mechanism to \gls{MA} is implemented to selectively amplify the output network feature maps.  As before, we extended this mechanism to other degradation parameters by combining these into an input vector.  
\end{itemize}

Many of the discussed metadata insertion mechanisms were initially introduced as repeated blocks which should be distributed across the entirety of an \gls{SR} network.  However, this can significantly increase the complexity (in both size and speed) of a network as well as make implementation difficult given the variety of network architectures available. Our testing has shown that, in most cases, simply adding one metadata-insertion block at the front of the network is enough to fully exploit the degradation vector information (results in Section \ref{metadata_insertion_results}).  Further implementation details of each block are provided in \Cref{implementation_details}.

\subsection{Degradation Prediction - Iterative Mechanism}
The simplest degradation prediction system tested for our framework is the \gls{DAN} iterative mechanism proposed in \cite{DAN}.  This network consists of two convolutional sub-modules - a \textit{restorer}, in charge of the super-resolution step and an \textit{estimator}, which predicts the blur kernel applied on an \gls{LR} image (in \gls{PCA} reduced form).  Both modules are co-dependent; the restorer produces an \gls{SR} image based on the degradation prediction while the estimator makes a degradation prediction informed by the \gls{SR} image.  By repeatedly alternating between the two modules, the results of both can be iteratively improved.  Furthermore, both networks can be optimised simultaneously by back-propagating the error of the \gls{SR} and degradation estimates.  

The iterative mechanism is straightforward to introduce into our framework.  We implemented the estimator module from \gls{DAN} directly, and then coupled its output with a core \gls{SR} network through a metadata insertion block (\Cref{fig:1}).  While the authors of \gls{DAN} only considered blur kernels in their work, it should be possible to direct the estimator to predict the parameters of any specified degradation directly.  We tested this hypothesis for both simple and complex degradations, the results of which are provided in \Cref{blur_kernel_prediction,complex_deg_prediction}. 

\subsection{Degradation Prediction - Contrastive Learning}
Contrastive learning is another prominent method for degradation estimation in blind \gls{SR}.  We considered three total methods for contrastive loss calculation, one of which is completely unsupervised and two which are semi-supervised as described in further detail hereunder.

\subsubsection{MoCo - Unsupervised Mechanism}
\begin{figure}[!htbp]
    \begin{adjustwidth}{-\extralength}{0cm}
        \centering
        \includegraphics[width=0.8\linewidth]{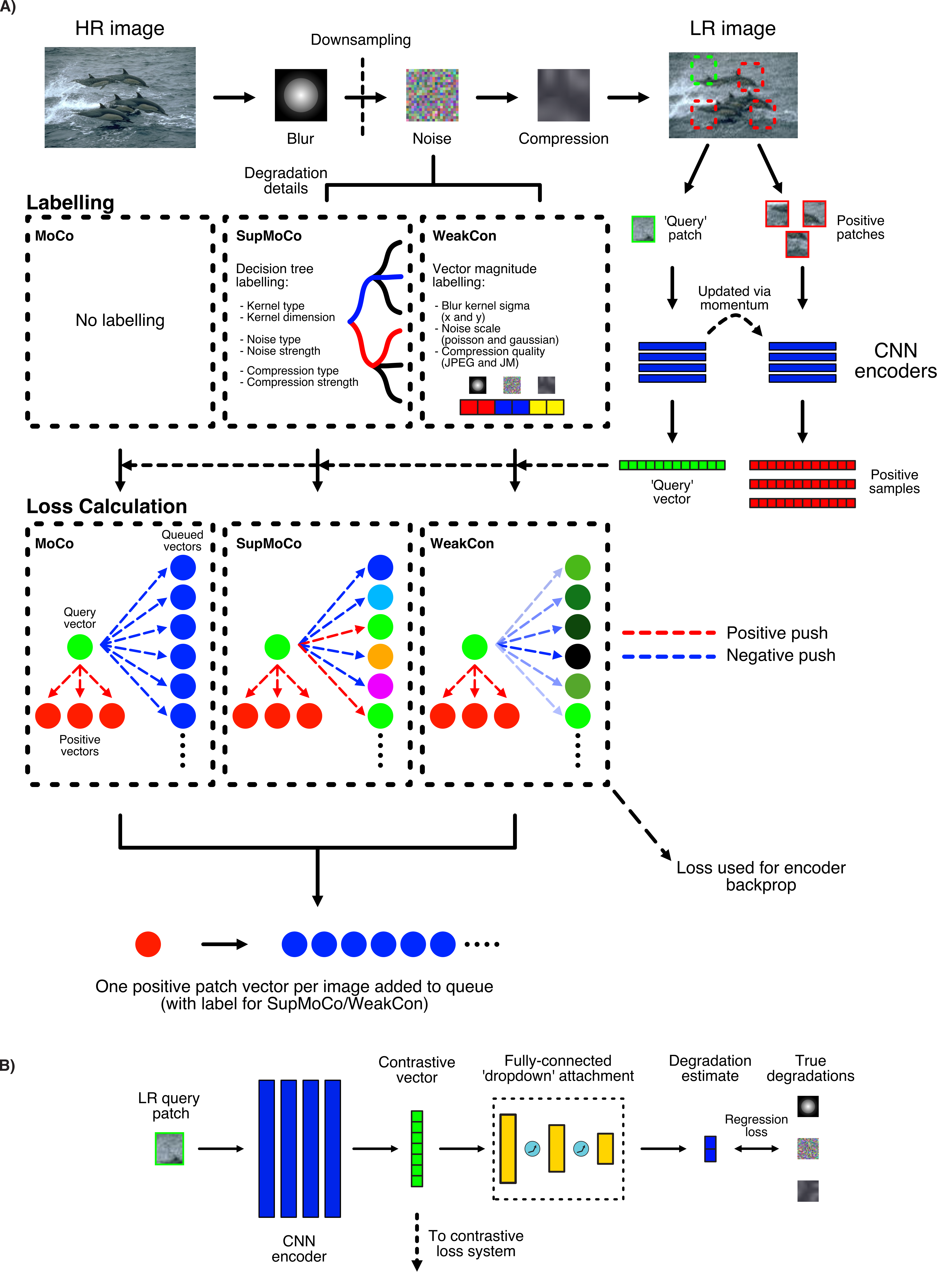}
    \end{adjustwidth}
    \caption{Contrastive learning for \gls{SR}. A) The three contrastive learning mechanisms considered.  While MoCo is the simplest and the only fully unsupervised method, SupMoCo and WeakCon provide more targeted learning at the expense of requiring user-defined labelling systems. B) Direct parameter regression can also be used to add an additional supervised element to the encoder training process.}
    \label{fig:3}
\end{figure}

Contrastive learning for blind \gls{SR} was first proposed in \cite{DASR21}, where the authors used the \gls{moco} \cite{moco20} mechanism to train convolutional encoders that are able to estimate the shape and noise content in blur kernels applied on \gls{LR} images.  The encoder is taught to generate closely-matched vectors for images with identical or similar degradations (e.g. equally sized blur kernels) and likewise generate disparate vectors for vastly different degradations.  These encoded vectors, while not directly interpretable, can be utilised by a downstream \gls{SR} model to inform the \gls{SR} process.  The proposed \gls{moco} encoder training mechanism works as follows:
\begin{itemize}
    \item Two identical encoders are instantiated.  One acts as the `query' encoder and the other as the `key' encoder.  The query encoder is updated directly via backpropagation from computed loss/error, while the key encoder is updated through a momentum mechanism from the query encoder.
    \item The encoders are directed to generate a degradation vector from one separate square patch per \gls{LR} image each (Figure \ref{fig:3}A, right).  The query encoder vector is considered as the reference for loss calculation, while the key encoder vector generated from the second patch acts as a positive sample.  The training objective is to drive the query vector to become more similar to the positive sample vector, while simultaneously repelling the query vector away from encodings generated from all other \gls{LR} images (negative samples).
    \item Negative samples are generated by placing previous key encoder vectors from distinct \gls{LR} images into a queue.  With both positive and negative encoded vectors available, an infoNCE-based \cite{infonce} loss function can be applied:
    
\begin{equation}\label{moco}
L_{MOCO} = \frac{1}{B} \left[ \mathlarger{\sum_{i=1}^{B}} -\log \frac{\exp \left(f_q(x_i^1) \cdot f_k(x_i^2) / \tau\right)}{\exp \left(f_q(x_i^1) \cdot f_k(x_i^2) / \tau\right) + \sum_{j=1}^{N_{\text {queue }}} \exp \left(f_q(x_i^1) \cdot q^j / \tau\right)} \right]
\end{equation}

where $f_q$ and $f_k$ are the query and key encoders, respectively, $x_i^1$ is the first patch from the $i^{th}$ image in a batch (batch size $B$), $q^j$ is the $j^{th}$ entry of the queue of size $N_{queue}$ and $\tau$ is a constant temperature scalar.  With this loss function, the query encoder is updated to simultaneously move its prediction closer to the positive encoding, and farther away from the negative set of encodings (refer to \Cref{fig:3}, \gls{moco} dotted boxes).  This loss should enable the encoder to distinguish between the different degradations present in the positive and negative samples.  
\end{itemize}
In \cite{DASR21}, only one positive patch is used per input image.  However, this can be easily extended to multiple positive patches through the following modifications to the loss function (shown in \textcolor{blue}{blue}):

\begin{equation}\label{moco_multi_patches}
L_{MOCO} =\frac{1}{B \times \textcolor{blue}{P^i}} \left[\mathlarger{\sum_{i=1}^{B}} -\log \frac{\textcolor{blue}{\sum_{l=1}^{P^i}}\exp \left(f_q(x_i^1) \cdot f_k(x_i^l) / \tau\right)}{ \textcolor{blue}{\sum_{l=1}^{P^i}} \exp \left(f_q(x_i^1) \cdot f_k(x_i^l) / \tau\right) + \sum_{j=1}^{N_{\text {queue }}} \exp \left(f_q(x_i^1) \cdot q^k / \tau\right)} \right]
\end{equation}
where $P^i$ is the number of positive patches for the $i^{th}$ image in a batch.  We retain just one positive patch (Equation \ref{moco}) to match \cite{DASR21} for most of our tests, unless indicated. 
\subsubsection{SupMoCo - Semi-Supervised Mechanism}\label{supmoco_description}
In the image classification domain, advances into semi-supervised contrastive learning have resulted in further performance improvements over \gls{moco}.  One such mechanism, \gls{supmoco} \cite{supmoco21}, provides more control over the contrastive training process.  In essence, all encoded vectors can be assigned a user-defined label, including the query vector.  With these labels, the contrastive loss function can be directed to push the query vector towards all key vectors sharing the same label while repelling them away from those that have different labels (\Cref{fig:3}A, \gls{supmoco} dotted boxes):

\begin{equation}
L_{SUPMOCO} =\frac{1}{B \times \textcolor{blue}{F^i}} \left[\mathlarger{\sum_{i=1}^{B}} -\log \frac{\sum_{l=1}^{P^i}\exp \left(f_q(x_i^1) \cdot f_k(x_i^l) / \tau\right) + \textcolor{blue}{\sum_{m=1}^{Q^i}\exp \left(f_q(x_i^1) \cdot f_k(x_i^m) / \tau\right)}}{ \sum_{l=1}^{P^i} \exp \left(f_q(x_i^1) \cdot f_k(x_i^l) / \tau\right) + \sum_{j=1}^{N_{\text {queue }}} \exp \left(f_q(x_i^1) \cdot q^k / \tau\right)} \right]
\end{equation}

where $Q^i$ is the number of samples in the queue with the same label as the query vector and $F^i = P^i + Q^i$.  New additions with reference to Equation \ref{moco_multi_patches} are highlighted in \textcolor{blue}{blue}.  

This system allows more control on the trajectory of the contrastive loss, reducing inaccuracies while pushing the encoder to recognise patterns based on the labels provided.  For our degradation pipeline, we implemented a decision tree system which assigns a unique label to each possible combination of degradations.  In brief, a  label is assigned to each degradation based on a number of factors:

\begin{itemize}
    \item Blur kernel type
    \item Blur kernel size; either low/high, which we refer to as \textit{double} precision (2 clusters per parameter), or low/medium/ high, which we refer to as \textit{triple} precision (3 clusters per parameter), classification.
    \item Noise type
    \item Noise magnitude (either a double or triple precision classification)
    \item Compression type
    \item Compression magnitude (either a double or triple precision classification)
\end{itemize}

\begin{figure}[H]
        \centering
\includegraphics[width=\linewidth]{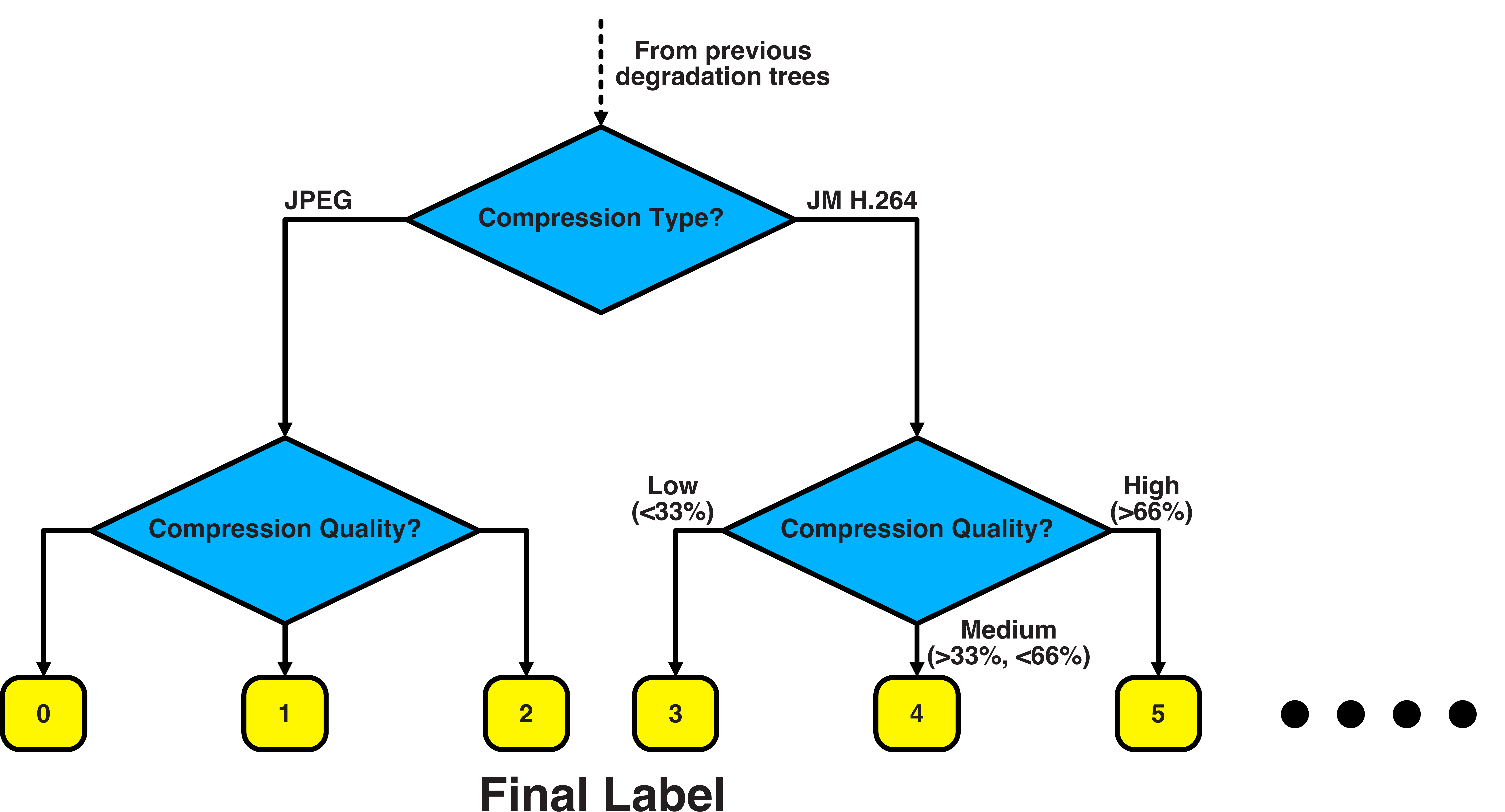}
    \caption{An example of a labelling decision tree for compression degradations.  The final label is used to direct contrastive loss in a \gls{supmoco} system.  Other degradation types can be linked to this tree, further diversifying the labels available.}
    \label{fig:4}
\end{figure}

An example of how this decision tree would work for compression type/magnitude labelling is provided in \Cref{fig:4}.

 The decision tree should push the encoder to more quickly identify the presence of different degradation classes than in the unsupervised case.  Aside from the labelling system, \gls{supmoco} is trained in an identical fashion to \gls{moco}, including the usage of momentum to update the key encoder.  A full description of all degradations applied in our pipeline is provided in \Cref{implementation_details}.

\subsubsection{WeakCon - Semi-Supervised Mechanism}
Another semi-supervised contrastive paradigm (which we refer to as WeakCon) has been proposed in \cite{IDMBSR}.  Instead of assigning discrete labels to each degradation, the authors propose a system for modulating the strength of the contrastive loss.  By calculating the difference between query and negative sample degradation magnitudes, the negative contrastive push can be increased or decreased according to how different the degradations are (\Cref{fig:3}A, WeakCon dotted boxes).  In \cite{IDMBSR}, the authors utilize the Euclidean distance between query/negative sample blur kernel width and noise sigma to calculate a weight $w$ for each negative sample.  With this weighting, the contrastive loss can be controlled as follows:

\begin{equation}
L_{WEAKCON} =\frac{1}{B \times P^i} \left[\mathlarger{\sum_{i=1}^{B}} -\log \frac{\sum_{l=1}^{P^i}\exp \left(f_q(x_i^1) \cdot f_k(x_i^l) / \tau\right)}{ \sum_{l=1}^{P^i} \exp \left(f_q(x_i^1) \cdot f_k(x_i^l) / \tau\right) + \sum_{j=1}^{N_{\text {queue }}} \textcolor{blue}{w_{ij}}\exp \left(f_q(x_i^1) \cdot q^k / \tau\right)} \right]
\end{equation}

where $w_{ij}$ indicates the distance between negative sample $j$ and query sample $i$.  We extend this special case to other degradations by similarly calculating the Euclidean distance between degradation vectors containing blur sigma x/y values, noise scales and compression quality.  New additions with reference to Equation \ref{moco_multi_patches} are highlighted in \textcolor{blue}{blue}. 

\subsubsection{Direct Regression Attachment}
While contrastive representations can be visualised using dimensionality reduction, it is difficult to quantify their prediction accuracy with respect to the true degradation parameters.  To provide further insight into the training process, we attach a further set of fully-connected layers to the contrastive encoder, as shown in \Cref{fig:3}B.  These layers are set to directly transform the contrastive vector into the magnitudes of the various degradations being estimated.  A regression loss (L1 loss between predicted vector and target degradation magnitudes) can also be introduced as an additional supervised element.  This direct parameter prediction can be easily quantified into an estimation error, which can help track training progress.  We train various models with and without these extra layers, with the details provided in \Cref{sec:experiments}.

\subsection{Extensions to Degradation Prediction}
In both the iterative and contrastive cases, our prediction mechanisms are centred around general degradation parameter prediction, and as such could be extended to any degradation which can be parameterised in some form.  Alternatively, degradations could be represented in vector form through the use of dimensionality reduction techniques (as is often done with blur kernels, on which \gls{PCA} is applied).  
Dimensionality reduction can also be used as an imperfect view of the differences between contrastive vectors encoded for different degradations.  We provide our analyses into the contrastive process in \Cref{blur_kernel_prediction,complex_deg_prediction}.  
\section{Experiments \& Results}\label{sec:experiments}
\subsection{Implementation Details}\label{implementation_details}
\subsubsection{Datasets and Degradations}
We created two main \gls{LR} degradation pipelines for our analyses:
\begin{itemize}
    \item \textbf{Simple pipeline (blurring and downsampling):}  For our metadata insertion screening and blind \gls{SR} comparison, we worked with a reduced degradation set of just gaussian blurring and bicubic downsampling corresponding to the `classical' degradation model shown in \Cref{eq:deg1}.  Apart from minimising confounding factors, this allows us to make direct comparisons with pre-trained models provided by the authors of other blind \gls{SR} networks.  For all scenarios, we used only 21$\times$21 isotropic Gaussian kernels with a random width ($\sigma$) in the range $[0.2, 3]$ (as recommended in \cite{Wang21}), and $\times$4 bicubic downsampling.  The $\sigma$ was normalised in the range $[0,1]$ before being fed to models.    
    \item \textbf{Complex pipeline:}  In our extended blind \gls{SR} training schemes, we used a full degradation pipeline matching the model in \Cref{eq:deg2}, i.e. sequential blurring, downsampling, noise addition and compression.  For each operation in the pipeline, a configuration was randomly (with a uniform distribution) selected from the following list:
    \begin{itemize}
        \item \textit{Blurring:}  As proposed in \cite{Wang21}, we sampled blurring from a total of 7 different kernel shapes: iso/anisotropic Gaussian, iso/anisotropic generalised Gaussian, iso/anisotropic plateau, and sinc.  Kernel $\sigma$ values (both vertical and horizontal) were sampled from the range $[0.2, 3]$, kernel rotation ranged from -$\pi$ to $\pi$ (all possible rotations) and the shape parameter $\beta$ ranged from $[0.5, 8]$ for both generalised Gaussian and plateau kernels.  For sinc kernels, we randomly selected the cutoff frequency from the range $[\pi/5, \pi]$.  All kernels were set to a size of 21$\times$21, and in each instance the blur kernel shape was randomly selected from the 7 available options with equal probability.  For a full exposition on the selection of each type of kernel, please refer to \cite{Wang21}.  
        \item \textit{Downsampling:}  As in the initial model screening, we retained just $\times$4 bicubic downsampling for all \gls{LR} images.
        \item \textit{Noise addition:}  Again following \cite{Wang21}, we injected noise using one of two different mechanisms, namely Gaussian (signal independent read noise) and Poisson (signal dependent shot noise).  Additionally, the noise was either independently added to each colour channel (colour noise), or applied to each channel in an identical fashion (grey noise).  The Gaussian and Poisson mechanisms were randomly selected with equal probability, grey noise was selected with a probability of 0.4, and the Gaussian/Poisson sigma/scale values were randomly sampled from the ranges $[1, 30]$ and $[0.05, 3]$ respectively.
        \item \textit{Compression:}  We increased the complexity of compression used in previous works by randomly selecting from either JPEG or JM H.264 (version 19) \cite{JM} compression at runtime.  For JPEG, a quality value was randomly selected from the range $[30, 95]$ (following \cite{Wang21}). For JM H.264, images were compressed as single-frame YUV files where a random I-slice Quantization Parameter (QPI) were selected from the range $[20, 40]$, following \cite{Aquilina21}.
    \end{itemize}
\end{itemize}

All our models were trained on the \gls{LR} images generated from the \gls{HR} images of DIV2K \cite{div2k} (800 images) and Flickr2K \cite{flickr2k} (2,650 images).  Validation and best model selection were performed on the DIV2K validation set (100 images). 

For final results comparison, the standard \gls{SR} test sets Set5 \cite{set5}, Set14 \cite{set14}, BSDS100 \cite{BSDS100}, Manga109 \cite{manga109} and Urban100 \cite{urban100} were utilised.  For these test images, the parameters of each degradation were explicitly selected.  The exact degradation details for each scenario are specified in all tables/figures presented.  The super-resolved images were compared with the corresponding target \gls{HR} images using a number of metrics during testing and validation, namely \gls{PSNR}, \gls{SSIM} \cite{ssim} (direct pixel comparison metrics) and \gls{LPIPS} \cite{lpips} (perceptual quality metric).  In all cases, images were first converted to YCbCr, and the Y channel used to compute metrics. 

The degradation pipelines and further implementation details are fully available in our linked PyTorch \cite{pytorch} \href{https://github.com/um-dsrg/RUMpy}{codebase}.

\subsubsection{Model Implementation, Training and Validation}
Due to the diversity of models investigated in this work, a number of different training and validation schemes have been followed according to the task and network being investigated:
\begin{itemize}
    \item \textbf{Non-blind \gls{SR} model training}:  For non-blind model training, we initialised networks with the hyperparameters recommended by their authors, unless specified.  All models were trained from scratch on \gls{LR}/\gls{HR} pairs generated from the DIV2K/Flickr2K datasets using either the simple or complex pipeline.  For the simple pipeline, one \gls{LR} image was generated from each \gls{HR} image.  For the complex pipeline, five \gls{LR} images were generated per \gls{HR} image to improve the diversity of degradations available.  For both cases, the \gls{LR} image set was generated once and used to train all models.   All simple pipeline networks were trained for 1,000 epochs, whereas the complex pipeline networks were trained for 200 epochs to ensure fair comparisons (since each epoch contains 5 times as many samples as the simple case).  This training duration was selected as a compromise between obtaining meaningful results and keeping the total training time low.  
    
    For both pipelines, training was carried out on 64$\times$64 \gls{LR} patches with the Adam \cite{adam} optimiser.  Variations in batch size and learning rate scheduling were made for specific models as necessary to ensure training stability and limit \gls{GPU} memory requirements.  The configurations for the non-blind \gls{SR} models tested are as follows:
    \begin{itemize}
        \item \gls{RCAN} \cite{rcan} and \gls{HAN} \cite{han}:  For these models, the batch size was set to 8 in most cases, and a cosine annealing scheduler \cite{cosine} was used with a warm restart after every 125,000 iterations and an initial learning rate of 1e-4.  Training was driven solely by the L1 loss function comparing the \gls{SR} image with the target \gls{HR} image.  After training, the epoch checkpoint with the highest validation \gls{PSNR} was selected for final testing. 
        \item \gls{Real-ESRGAN} \cite{Wang21}:  The same scheme described for the original implementation was used to train this model.  This involved two phases: (i) a pre-training stage where the generator was trained with just an L1 loss, and (ii) a multi-loss stage where a discriminator and VGG perceptual loss network were introduced (further details are provided in \cite{Wang21}).  We pre-trained the model for 715 and 150 epochs (which match the pretrain:GAN ratio as originally proposed in \cite{Wang21}) for the simple and complex pipelines, respectively.  For both cases, the pre-training optimiser learning rate was fixed at 2e-4, while the multi-loss stage involved a fixed learning rate of 1e-4.  A batch size of 8 was used in all cases.  After training, the model checkpoint with the lowest validation \gls{LPIPS} score in the last 10\% of epochs was selected for testing.
        \item ELAN \cite{elan}:  For this model a batch size of 8 and a constant learning rate of 2e-4 was used in all cases.  As with \gls{RCAN} and \gls{HAN}, the L1 loss was used to drive training and the epoch checkpoint with the highest validation \gls{PSNR} was selected for final testing.
    \end{itemize}
    \item \textbf{Iterative Blind \gls{SR}}:  Since the \gls{DAN} iterative scheme requires the \gls{SR} image to improve its degradation estimate, the predictor model needs to be trained simultaneously with the \gls{SR} model.  We used the same \gls{CNN}-based predictor network described in DANv1 \cite{DAN} for our models and fixed the iteration count to four in all cases (matching the implementation as described in \cite{DAN}).  We coupled this predictor with our non-blind \gls{SR} models using the framework described in \Cref{framework}.  We trained all \gls{DAN} models by optimising for the \gls{SR} L1 loss (identical to the non-blind models) and an additional L1 loss component comparing the prediction and ground-truth vectors.  Target vectors varied according to the pipeline, the details of each are provided in their respective results sections.  For each specific model architecture, the hyperparameters and validation selection criteria were all set to be identical to that of the base, non-blind model.  The batch size for all models was adjusted to 4 due to the increased \gls{GPU} memory requirements needed for the iterative training scheme.  Accordingly, whenever a warm restart scheduler was used, the restart point was adjusted to 250,000 iterations (to maintain the same total number of iterations as performed by the other models that utilised a batch size of 8 for 125,000 iterations).

    Additionally, we also trained the original \gls{DAN}v1 model from scratch, using the same hyperparameters from \cite{DAN} and the same validation scheme as the other \gls{DAN} models.  The batch size was also fixed to 4 in all cases.
    \item \textbf{Contrastive Learning}:  We used the same encoder from \cite{DASR21} for most of our contrastive learning schemes.  This encoder consists of a convolutional core connected to a set of three fully-connected layers.  During training, we used the output of the fully-connected layers (\textit{Q}) to calculate loss values (i.e. $f_q$ and $f_k$ in Equation \ref{moco}) and update the encoder weights, following \cite{DASR21}.  Before coupling the encoder with an \gls{SR} network, we first pre-trained the encoder directly.  For this pre-training, the batch size was set to 32 and data was generated online i.e. each \gls{LR} image was synthesised on the fly at runtime.  All encoders were trained with a constant learning rate of 1e-3, a patch size of 64$\times$64 and the Adam optimiser.  The encoders were trained until the loss started to plateau and \gls{TSNE} clustering of degradations generated on a validation set composed of 400 images from CelebA \cite{celeba} and BSDS200 \cite{BSDS100} was clearly visible (more details on this process are provided in \Cref{contrastive_analysis}).  In all cases, the temperature hyperparameter, momentum value, queue length, and encoder output vector size were set to 0.07, 0.999, 8192 and 256 respectively (matching the models from \cite{DASR21}). 

    After pre-training, each encoder was coupled to non-blind \gls{SR} networks using the framework discussed in Section \ref{framework}. For standard encoders, the \textit{encoding} (i.e. the output from the convolutional core that bypasses the fully-connected layers) is typically fed into metadata insertion blocks directly, unless specified.  For encoders with a regression component (see \Cref{fig:2}B), the dropdown output is fed to the metadata insertion block instead of the encoding.  The combined encoder + \gls{SR} network was then trained using the same dataset and hyperparameters as the non-blind case.  The encoder weights were frozen and no gradients were generated for the encoding at runtime, unless specified. 
\end{itemize}

In our analysis, we use the simple pipeline as our primary blind \gls{SR} task and the complex pipeline as an extension scenario for the best performing methods.  Section \ref{metadata_insertion_results} will discuss our metadata insertion block testing, while sections \ref{blur_kernel_prediction} and \ref{simple_pipeline_results} will present our degradation prediction and \gls{SR} analysis on the simple pipeline respectively.  Sections \ref{complex_deg_prediction} and \ref{complex_sr} will follow up with our analysis on the complex pipeline and Section \ref{real_data_tests} will present some of our blind \gls{SR} results on real-world degraded images.
\subsection{Metadata Insertion Block Testing} \label{metadata_insertion_results}
To test and compare the various metadata insertion blocks selected, we implemented each block into \gls{RCAN}, and trained a separate model from scratch on our simple pipeline dataset.  Each metadata insertion block was fed with the real blur kernel (normalised in the range $[0,1]$) width or PCA-reduced kernel representation for each \gls{LR} image. The test results for each model are presented in \Cref{non_blind_psnr} (PSNR) and \Cref{fig:non_blind} (bar graph comparison), with additional results shown in the supplementary information (Table S1 containing SSIM results). 

\begin{table}[!htbp]
	\caption{\gls{PSNR} (dB) \gls{SR} results on simple pipeline comparing metadata insertion blocks. 'low' refers to a $\sigma$ of 0.2, `med' refers to a $\sigma$ of 1.6 and `high' refers to a $\sigma$ of 3.0.   The models in the 'Non-Blind' category are all \gls{RCAN} models upgraded with the indicated metadata insertion block (only 1 block per network unless indicated).  \gls{MA} (all) and \gls{DA} (all) refers to \gls{RCAN} with 200 individual \gls{MA} or \gls{DA} blocks inserted throughout the network, respectively.  \gls{MA} (\gls{PCA}) refers to \gls{RCAN} with a single \gls{MA} block provided with a 10-element \gls{PCA}-reduced vector of the blur kernel applied.  \gls{DGFMB} (no FC) refers to \gls{RCAN} with a \gls{DGFMB} layer where the metadata input is not passed through a fully-connected layer prior to concatenation (refer to \Cref{fig:2}E). The best result for each set is shown in \textcolor{red}{red}, while the second-best result is shown in \textcolor{blue}{blue}.}
	\renewcommand{\arraystretch}{1.3}
 \setlength{\tabcolsep}{4.5pt}
 	  \begin{adjustwidth}{-\extralength}{0cm}
	    \begin{center}
     \tablesize{\footnotesize} 
            \begin{tabularx}{\linewidth}{cccccccccccccccc}
\toprule
           \textbf{Model} & \multicolumn{3}{c}{\textbf{Set5}} & \multicolumn{3}{c}{\textbf{Set14}} & \multicolumn{3}{c}{\textbf{BSDS100}} & \multicolumn{3}{c}{\textbf{Manga109}} & \multicolumn{3}{c}{\textbf{Urban100}} \\
                 &                      \textbf{low} &                      \textbf{med} &                     \textbf{high} &                      \textbf{low} &                      \textbf{med} &                     \textbf{high} &                      \textbf{low} &                      \textbf{med} &                     \textbf{high} &                      \textbf{low} &                      \textbf{med} &                     \textbf{high} &                      \textbf{low} &                      \textbf{\textbf{med}} &                     \textbf{high} \\
\midrule
\rowcolor{gray!5}[\dimexpr\tabcolsep + 1pt\relax]           \textbf{Baselines} &&&&&&&&&&&&&&& \\
     \rowcolor{gray!5}[\dimexpr\tabcolsep + 1pt\relax]  Bicubic &                    27.084 &                    25.857 &                    23.867 &                    24.532 &                    23.695 &                    22.286 &                    24.647 &                    23.998 &                    22.910 &                    23.608 &                    22.564 &                    20.932 &                    21.805 &                    21.104 &                    19.944 \\
    \rowcolor{gray!5}[\dimexpr\tabcolsep + 1pt\relax]    Lanczos &                    27.462 &                    26.210 &                    24.039 &                    24.760 &                    23.925 &                    22.409 &                    24.811 &                    24.173 &                    23.007 &                    23.923 &                    22.850 &                    21.071 &                    21.989 &                    21.293 &                    20.046 \\
      \rowcolor{gray!5}[\dimexpr\tabcolsep + 1pt\relax]     RCAN &                    30.675 &                    30.484 &                    29.635 &                    27.007 &                    27.003 &                    26.149 &                    26.337 &                    26.379 &                    25.898 &                    29.206 &                    29.406 &                    27.956 &                    24.962 &                    24.899 &                    23.960 \\
      \rowcolor{gray!15}[\dimexpr\tabcolsep + 1pt\relax]           \textbf{Non-Blind} &&&&&&&&&&&&&&& \\
       \rowcolor{gray!15}[\dimexpr\tabcolsep + 1pt\relax]      MA &                    30.956 &   \textcolor{red}{30.973} &                    29.880 &   \textcolor{red}{27.091} &   \textcolor{red}{27.094} &                    26.383 &                    26.389 &                    26.440 &                    26.002 &                    29.529 &                    29.834 &                    28.799 &                    25.059 &                    25.008 &                    24.218 \\
          \rowcolor{gray!15}[\dimexpr\tabcolsep + 1pt\relax]    MA (all) &                    30.926 &                    30.911 &                    29.911 &                    27.057 &                    27.066 &                    26.373 &                    26.386 &                    26.441 &                    26.005 &   \textcolor{red}{29.542} &   \textcolor{red}{29.874} &   \textcolor{red}{28.865} &                    25.043 &                    24.995 &                    24.224 \\
           \rowcolor{gray!15}[\dimexpr\tabcolsep + 1pt\relax]   MA (PCA) &                    30.947 &                    30.942 &                    29.873 &                    27.052 &                    27.038 &                    26.382 &                    26.388 &  \textcolor{blue}{26.442} &                    26.003 &                    29.489 &                    29.817 &                    28.740 &  \textcolor{blue}{25.077} &                    25.014 &                    24.227 \\
            \rowcolor{gray!15}[\dimexpr\tabcolsep + 1pt\relax]      SRMD &                    30.946 &  \textcolor{blue}{30.960} &                    29.906 &                    27.086 &                    27.072 &                    26.358 &   \textcolor{red}{26.393} &                    26.439 &                    25.998 &                    29.515 &                    29.806 &                    28.750 &                    25.072 &                    25.012 &                    24.204 \\
              \rowcolor{gray!15}[\dimexpr\tabcolsep + 1pt\relax]     SFT &                    30.960 &                    30.958 &   \textcolor{red}{29.972} &                    27.065 &                    27.066 &                    26.386 &                    26.388 &                    26.439 &  \textcolor{blue}{26.006} &  \textcolor{blue}{29.541} &  \textcolor{blue}{29.849} &  \textcolor{blue}{28.835} &   \textcolor{red}{25.088} &   \textcolor{red}{25.022} &   \textcolor{red}{24.241} \\
              \rowcolor{gray!15}[\dimexpr\tabcolsep + 1pt\relax]      DA &                    30.930 &                    30.934 &  \textcolor{blue}{29.929} &  \textcolor{blue}{27.088} &  \textcolor{blue}{27.093} &   \textcolor{red}{26.389} &  \textcolor{blue}{26.392} &   \textcolor{red}{26.449} &   \textcolor{red}{26.009} &                    29.515 &                    29.839 &                    28.801 &                    25.056 &  \textcolor{blue}{25.019} &  \textcolor{blue}{24.239} \\
           \rowcolor{gray!15}[\dimexpr\tabcolsep + 1pt\relax]   DA (all) &                    30.956 &                    30.958 &                    29.888 &                    27.033 &                    27.044 &                    26.358 &                    26.378 &                    26.437 &                    25.990 &                    29.451 &                    29.844 &                    28.781 &                    24.998 &                    24.999 &                    24.207 \\
            \rowcolor{gray!15}[\dimexpr\tabcolsep + 1pt\relax]     DGFMB &  \textcolor{blue}{30.969} &                    30.955 &                    29.891 &                    27.068 &                    27.060 &                    26.347 &                    26.386 &                    26.440 &                    26.001 &                    29.508 &                    29.811 &                    28.802 &                    25.067 &                    25.007 &                    24.228 \\
        \rowcolor{gray!15}[\dimexpr\tabcolsep + 1pt\relax] DGFMB (no FC) &   \textcolor{red}{30.985} &                    30.941 &                    29.909 &                    27.062 &                    27.077 &  \textcolor{blue}{26.388} &                    26.392 &                    26.436 &                    26.006 &                    29.508 &                    29.806 &                    28.781 &                    25.073 &                    25.014 &                    24.237 \\
\bottomrule
            \end{tabularx}
		\end{center}
    \end{adjustwidth}
\label{non_blind_psnr}
\end{table}

\begin{figure}[!htbp]
    \begin{adjustwidth}{-\extralength}{0cm}
        \centering
        \includegraphics[width=0.8\linewidth]{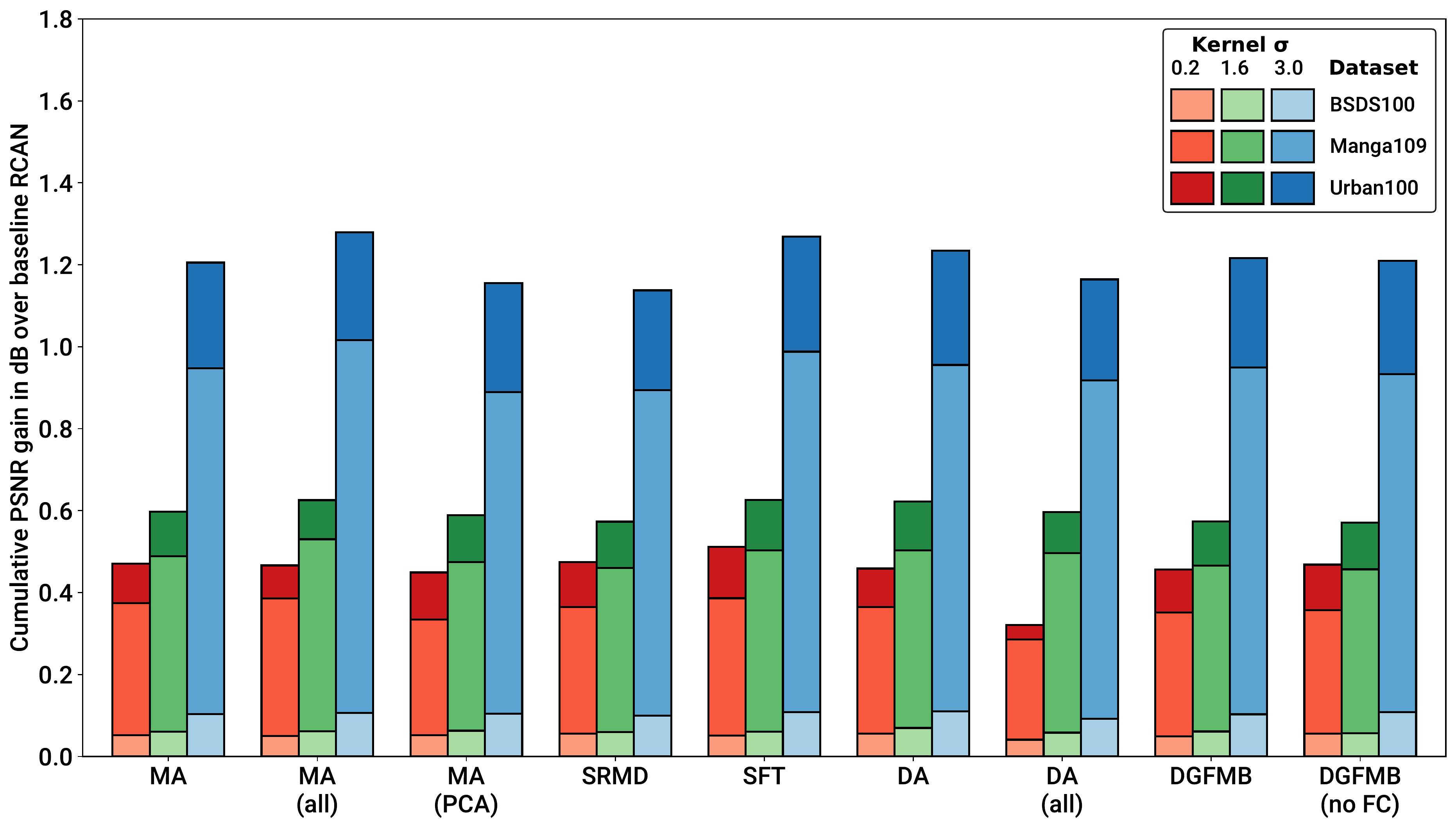}
    \end{adjustwidth}
    \caption{Bar graph showing improvement in PSNR over baseline RCAN for each metadata insertion block.  PSNR improvements are stacked on each other for each specific $\sigma$ to show cumulative PSNR gain across datasets.}
    \label{fig:non_blind}
\end{figure}

From the results, it is evident that metadata insertion provides a significant boost to performance across the board.  Somewhat surprisingly, the results also show that no single metadata insertion block has a clear advantage over the rest.  Every configuration tested, including those where multiple metadata insertion blocks are provided, produces roughly the same level of performance with only minor variations across dataset/degradation combinations.  This outcome suggests that each metadata block is producing the same amount of useful information from the input kernel.  Further complexity, such as the \gls{DA} block's kernel transformation or the \gls{SFT}/\gls{DGFMB} feature map concatenation, provides no further gain in performance.  Even adding further detail to the metadata, such as by converting the full blur kernel into a \gls{PCA}-reduced vector, provides no performance gains.  This again seems to suggest that the network is capable of extrapolating the kernel width to the full kernel description, without requiring any additional data engineering.  Furthermore, adding just a single block at the beginning of the network appears to be enough to inform the whole network, with additional layers providing no improvement (while a decrease in performance is actually observed in the case of \gls{DA}).  We hypothesise that this might be due to the fact that degradations are mostly resolved in the earlier low-frequency stages of the network.

Given that all metadata insertion blocks provide almost identical performance, we selected a single \gls{MA} block for our blind \gls{SR} testing, given its low overhead and simplicity with respect to the other approaches.  While it is clear the more complex metadata insertion blocks do not provide increased performance on this dataset, it is still possible that they might provide further benefit if other types of metadata are available. 

\subsection{Blur Kernel Degradation Prediction}\label{blur_kernel_prediction}
To test our degradation prediction mechanisms, we evaluated the performance of these methods on a range of conditions and datasets.

\subsubsection{Contrastive Learning}\label{contrastive_analysis}
For contrastive learning methods, the prediction vectors generated are not directly interpretable.  This makes it difficult to quantify the accuracy of the prediction without some form of clustering/regression analysis.  However, through the use of dimensionality reduction techniques such as \gls{TSNE} \cite{tsne}, the vectors can be easily reduced to 2-D, which provides an opportunity for qualitative screening of each model in the form of a graph.  

\begin{figure}[!ht]
    \begin{adjustwidth}{-\extralength}{0cm}
        \centering
        \includegraphics[width=\linewidth]{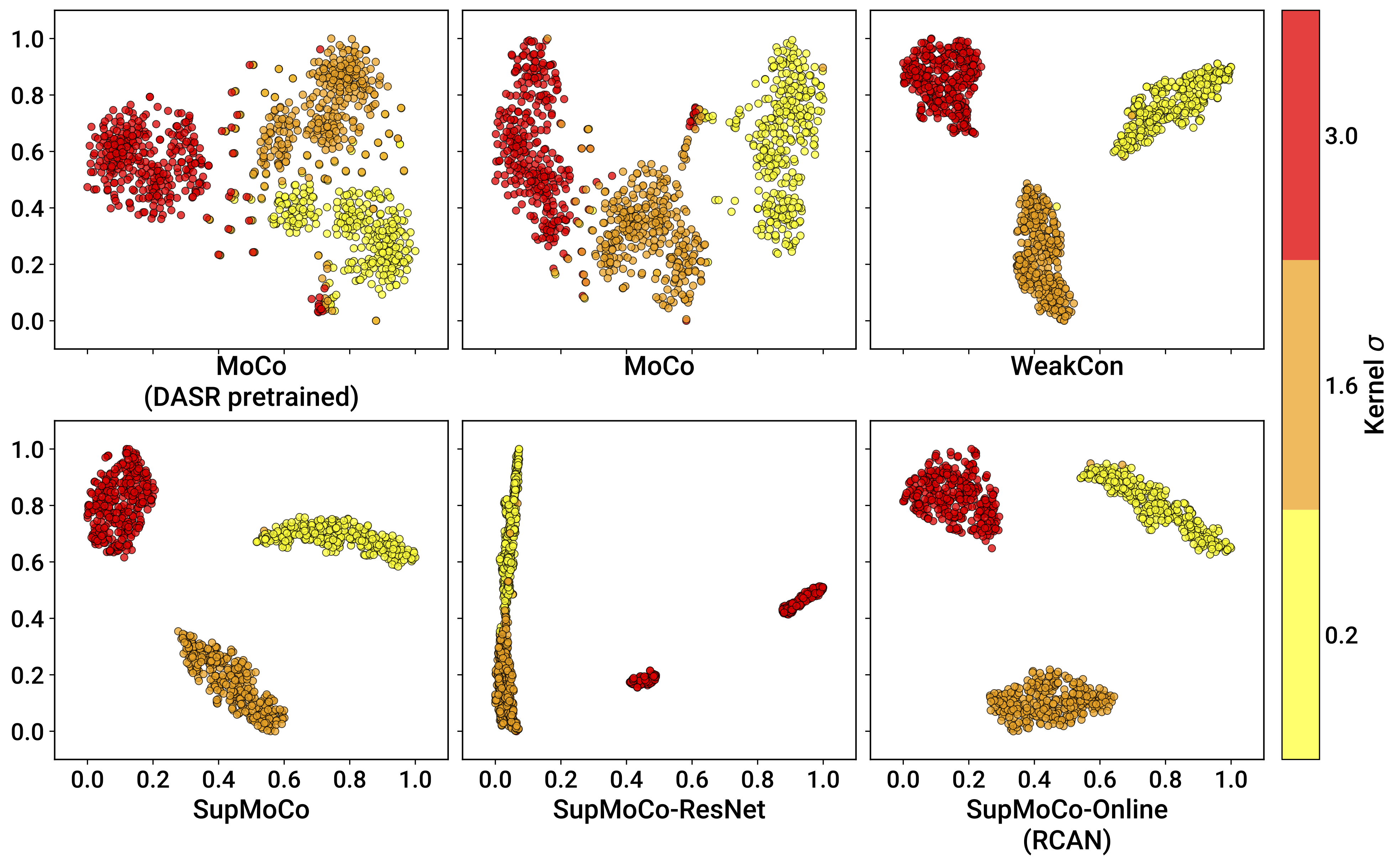}
    \end{adjustwidth}
    \caption{\gls{TSNE} plots (perplexity value of 40) showing the separation power of the different contrastive learning algorithms considered.  Each dimension was independently normalised in the range $[0,1]$ after computing the \gls{TSNE} results.  All models were evaluated on all of the BSDS100/Manga109/Urban100 test images, each degraded with one of the indicated kernel $\sigma$ (total of 927 images).  \gls{TSNE} fitting and reduction was performed separately on each set of outputs from the models considered.  The \gls{DASR} pretrained model was instantiated using the weights from \cite{DASR21}.  MoCo-based encoders (including the pretrained encoder from \gls{DASR}) can clearly separate the three different blur kernel populations, but the interface between each population is not well defined.  In contrast, the semi-supervised algorithms achieve almost perfect separation of the different blur kernels, forming well-defined clusters.  Variations to the formula such as using a larger encoder (ResNet) or continuing contrastive training during the \gls{SR} process appear to provide no benefit.}
    \label{fig:simple_tsne}
\end{figure}

We trained separate encoders using our three contrastive algorithms on the simple degradation pipeline.  The testing epoch and training details for each model are provided in \Cref{contrastive_training_epochs}.  We used the trained encoders to generate prediction vectors for the entire BSDS100/Manga109/Urban100 testing datasets (927 images) and applied \gls{TSNE} reduction for each set of outputs. The \gls{TSNE} results are presented in \Cref{fig:simple_tsne}.  It is immediately apparent that all of the models achieve some level of separation between the three different $\sigma$ values.  However, the semi-supervised methods produce very clear clustering (with just a few outliers) while the \gls{moco} methods generate clusters with less well-defined edges.  The influence of the labelling systems clearly produces a very large repulsion effect between the different $\sigma$ widths, which the unsupervised \gls{moco} system cannot match.  Interestingly, there is no discernable distinction between the WeakCon and \gls{supmoco} plots, despite their different modes of action.  Additionally, minor modifications to the training process such as swapping the encoder for a larger model (e.g. ResNet) or continuing to train the predictor in tandem with an \gls{SR} model (\gls{SR} results in Section \ref{simple_pipeline_results}) appear to provide no benefit or even degrade the output clusters.
\begin{table}[!htbp]
	\caption{The evaluation epoch selected for each contrastive encoder considered in our simple pipeline analysis.  For \gls{supmoco} schemes, each model was trained with triple precision (three labels) with respect to the $\sigma$ value.  For WeakCon, the distance $w_{ij}$ was calculated through the Euclidean distance between the query and negative sample normalised $\sigma$ (as proposed in \cite{IDMBSR}).  The positive patches per query patch refers to how many patches are produced per iteration to act as positive samples, as stipulated in Equation \ref{moco_multi_patches}.  Epochs were qualitatively selected based on when the contrastive loss starts to plateau and degradation clustering on the validation set is observed.  The results in Table \ref{simple_blind_psnr} show that further training of the encoders beyond this point provides little to no gain.}
	    \begin{center}
     \tablesize{\small} 
            \begin{tabularx}{\linewidth}{ccc}
\toprule
\textbf{Model} & \textbf{Positive patches per query patch} & \textbf{Epoch Selected}\\
\midrule
MoCo & 1 & 2104\\
SupMoCo & 3 & 542\\
WeakCon & 1 & 567 \\
SupMoCo (regression) & N/A & 376\\
SupMoCo (contrastive + regression) & 3 & 115 \\
SupMoCo (ResNet) & 3 & 326\\
\bottomrule
            \end{tabularx}
		\end{center}
\label{contrastive_training_epochs}
\end{table}

\subsubsection{Regression Analysis}
For our iterative and regression models, the output prediction is much simpler to interpret.  Direct $\sigma$ and \gls{PCA} kernel estimates can be immediately compared with the actual value.  We trained a variety of iterative \gls{DAN} models, using \gls{RCAN} as our base \gls{SR} model for consistency.  Several separate \gls{RCAN}-\gls{DAN} models were implemented; one specifically predicting the $\sigma$ and others predicting a 10-element \gls{PCA} representation of each kernel. We also trained two DANv1 models (predicting \gls{PCA} kernels) from scratch for comparison: one using a fixed learning rate of 2e-4 (matching the original implementation in \cite{DAN}) and one using our own cosine annealing scheduler with a restart value of 250,000 (matching our other models).  To compare with their results, along with the results of $\times$4 pretrained \gls{SR} models from the literature, we evaluated their prediction capability across the BSDS100/Manga109/Urban100 datasets and different $\sigma$.  The pretrained models for \gls{IKC}, \gls{DAN}v1 and \gls{DAN}v2 were extracted from their respective official code repositories.  Each of these models were also trained on DIV2K/Flickr2K images, but degradations were generated online (with $\sigma$ in the range $[0.2,4]$), which should result in superior performance.  

\begin{figure}[!ht]
        \centering
        \includegraphics[width=\linewidth]{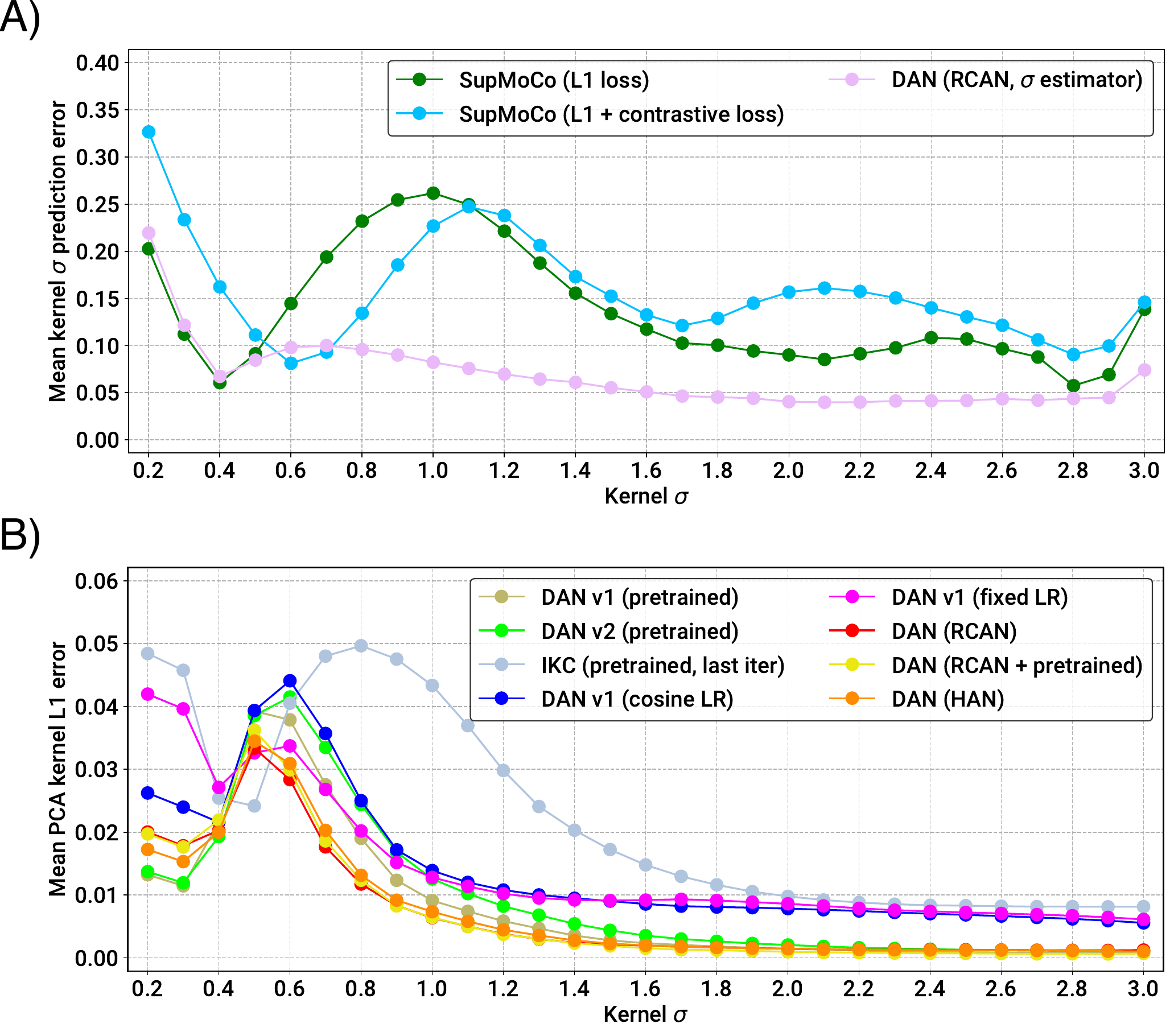}
    \caption{Prediction capabilities of regression and iterative-based encoders on the BSDS100, Manga109 and Urban100 testing sets (total of 309 images per $\sigma$).  A) Plot showing the relation between the average kernel prediction error and the actual kernel width for direct regression models. \gls{DAN} provides the best accuracy in general, but all models appear to suffer when the blur kernel is close to its lower limit.  B)  Plot showing the relation between average prediction error (in \gls{PCA}-space) and the actual $\sigma$ for iterative models. \gls{IKC} is the weakest model, while all \gls{DAN} models appear to follow the same distribution, with minor variations.  Again, all models appear to suffer at the lower end of the spectrum.}
    \label{fig:simple_regression}
\end{figure}

Figure \ref{fig:simple_regression}A shows the prediction error of the direct regression models that were trained (both contrastive and iterative models).  The results clearly show that the \gls{DAN} predictor is the strongest of those tested, with errors below 0.05 in some cases (representing an error of less than 2.5\%).  The contrastive/regression methods, while producing respectable results in select scenarios, seem to suffer across most of the distribution tested.  For both types of models, the error seems to increase when the width is at its lower range.  We hypothesise that, at this point, it is difficult to distinguish between $\sigma$ of 0.2-0.4, given that the corresponding kernels are quite small.  

Figure \ref{fig:simple_regression}B shows the results of the \gls{PCA} prediction models.  The plot shows that our \gls{RCAN}-\gls{DAN} models achieve very similar prediction performance to the pretrained DANs.  What makes this result remarkable is the fact that our models were trained for much less time than the pretrained \gls{DAN} models, both of which were trained for $\approx$7,000 epochs.  Training DANv1 from scratch for the same amount of time as our models (1,000 epochs) shows that the prediction performance at this point is markedly worse.  It is clear that the larger and more capable \gls{RCAN} model is helping boost the $\sigma$ prediction performance significantly.  On the other hand, the pretrained \gls{IKC} model is significantly outclassed by all \gls{DAN} models in almost all scenarios.  It is also worth noting that the prediction of kernels at the lower end of the spectrum suffers from increased error, across the board.    
\subsection{Blind SR on Simple Pipeline}\label{simple_pipeline_results}
\begin{figure}[!ht]
        \centering
        \includegraphics[width=\linewidth]{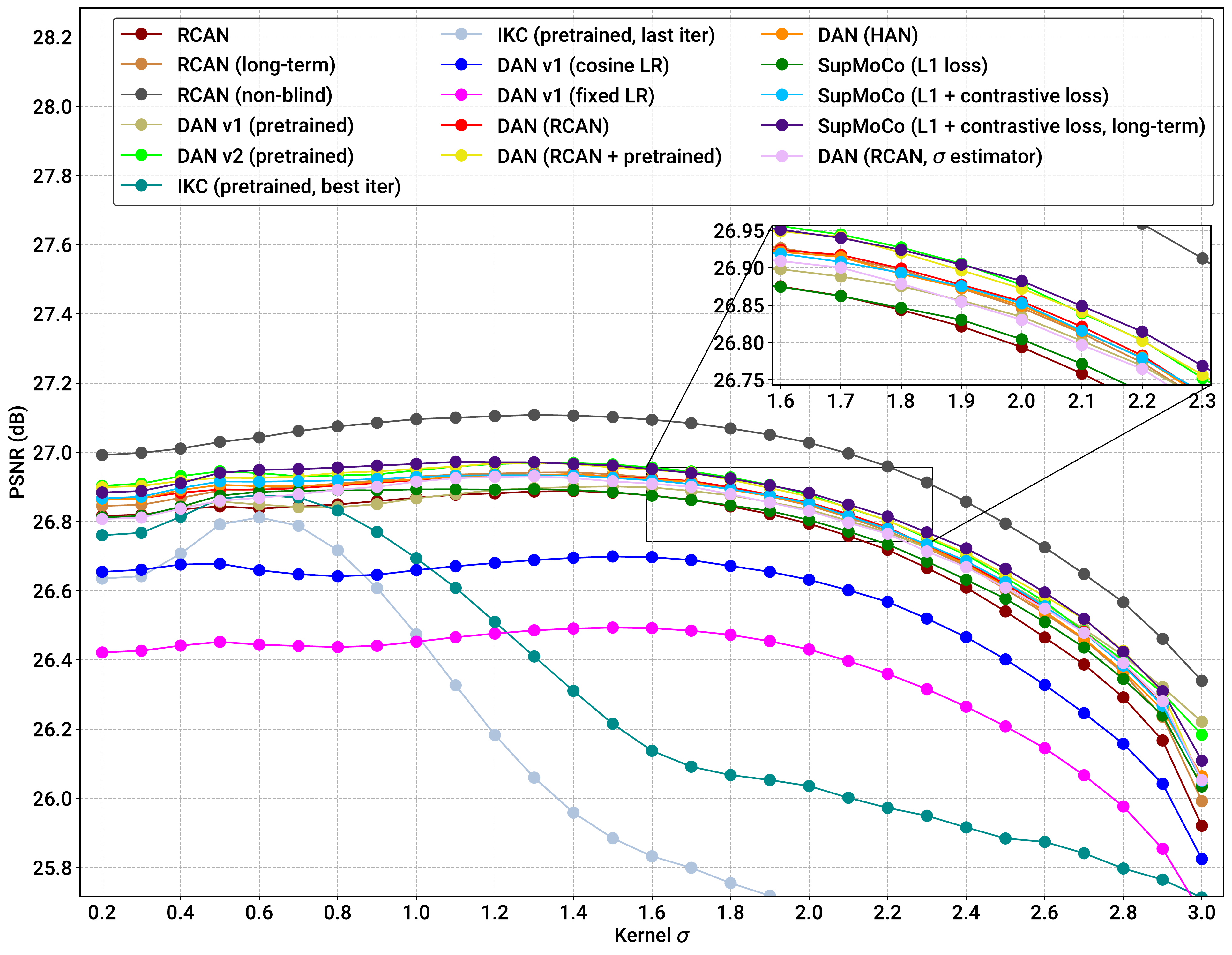}
    \caption{Plot showing the relationship of \gls{SR} performance (as measured by \gls{PSNR}) with the level of blurring within an image.  \gls{PSNR} was measured by running each model on the BSDS100/Manga109/Urban100 datasets (309 images) degraded with each specified $\sigma$.  `L1 loss' refers to the regression component added to our contrastive encoders (Figure \ref{fig:3}).  All models appear to show degraded performance with higher levels of blurring, indicating that none of the models analysed are capable of fully removing blurring effects from the final image.  Matching the results in Table \ref{simple_blind_psnr}, \gls{DAN} models and the pretrained \gls{IKC} (when analysing only the last output iteration) model are significantly worse than the other models analysed.  Our own \gls{RCAN}-\gls{DAN} models have a very similar distribution to the pretrained \gls{DAN} models, while our long-term RCAN-SupMoCo and pre-initialised \gls{RCAN}-\gls{DAN} match their performance.  The non-blind model outperforms all other models, indicating further improvements are possible with better degradation prediction.}
    \label{fig:psnr_vs_sigma}
\end{figure}

The real test for our combined \gls{SR} and predictor models is the blind \gls{SR} performance.  Table \ref{simple_blind_psnr}  presents the blind \gls{SR} \gls{PSNR} results of all the models considered on the test sets under various levels of blur $\sigma$.  \gls{SSIM} results are also provided in the supplementary information (Table S2).  Figures \ref{fig:bar_chart_simple} and \ref{fig:psnr_vs_sigma} further complement these results, with a bar chart comparison of key models and a closer look at the \gls{SR} performance across various levels of of $\sigma$, respectively.  

\begin{table}[!htbp]
	\caption{\gls{PSNR} (dB) \gls{SR} results on the simple pipeline comparing blind \gls{SR} methods.  `low' refers to a $\sigma$ of 0.2, `med' refers to a $\sigma$ of 1.6 and `high' refers to a $\sigma$ of 3.0.  For \gls{RCAN}/\gls{HAN} models with metadata, this is inserted using one \gls{MA} block at the front of the network.  \gls{IKC} was tested twice: once by evaluating the best iteration of the model output (as implemented in the official codebase) and once by evaluating the final iteration (iteration 7).  `noisy sigma' refers to the use of normalised $\sigma$ metadata that has been corrupted with Gaussian noise (mean 0, standard deviation of 0.1) for both training and testing.   `Long term' refers to models that have been screened after training for 2,200 epochs, rather than the standard 1,000.  The best result for each set is shown in \textcolor{red}{red}, while the second-best result is shown in \textcolor{blue}{blue}.}	\renewcommand{\arraystretch}{1.3}
 \setlength{\tabcolsep}{4pt}
 	  \begin{adjustwidth}{-\extralength}{0cm}
	    \begin{center}
     \tablesize{\scriptsize} 
            \begin{tabularx}{\linewidth}{P{3.9cm}ccccccccccccccc}

\toprule
                               \textbf{Model} & \multicolumn{3}{c}{\textbf{Set5}} & \multicolumn{3}{c}{\textbf{Set14}} & \multicolumn{3}{c}{\textbf{BSDS100}} & \multicolumn{3}{c}{\textbf{Manga109}} & \multicolumn{3}{c}{\textbf{Urban100}} \\
                                     &                       \textbf{low} &                       \textbf{med} &                      \textbf{high} &                       \textbf{low} &                       \textbf{med} &                      \textbf{high} &                       \textbf{low} &                       \textbf{med} &                      \textbf{high} &                       \textbf{low} &                       \textbf{med} &                      \textbf{high} &                       \textbf{low} &                       \textbf{med} &                      \textbf{high} \\
\hline
\rowcolor{gray!5}[\dimexpr\tabcolsep + 1pt\relax] \textbf{Classical} &&&&&&&&&&&&&&& \\
 \rowcolor{gray!5}[\dimexpr\tabcolsep + 1pt\relax]          Bicubic &                    27.084 &                    25.857 &                    23.867 &                    24.532 &                    23.695 &                    22.286 &                    24.647 &                    23.998 &                    22.910 &                    23.608 &                    22.564 &                    20.932 &                    21.805 &                    21.104 &                    19.944 \\
 \rowcolor{gray!5}[\dimexpr\tabcolsep + 1pt\relax]           Lanczos &                    27.462 &                    26.210 &                    24.039 &                    24.760 &                    23.925 &                    22.409 &                    24.811 &                    24.173 &                    23.007 &                    23.923 &                    22.850 &                    21.071 &                    21.989 &                    21.293 &                    20.046 \\
 \rowcolor{gray!15}[\dimexpr\tabcolsep + 1pt\relax]           \textbf{Pretrained} &&&&&&&&&&&&&&& \\
  \rowcolor{gray!15}[\dimexpr\tabcolsep + 1pt\relax] IKC (pretrained-best-iter) &  \textcolor{blue}{30.828} &                    30.494 &   \textcolor{red}{30.049} &                    26.981 &                    26.729 &   \textcolor{red}{26.603} &                    26.303 &                    26.195 &   \textcolor{red}{25.890} &                    29.210 &                    27.926 &                    27.416 &                    24.768 &                    24.291 &                    23.834 \\
  \rowcolor{gray!15}[\dimexpr\tabcolsep + 1pt\relax] IKC (pretrained-last-iter) &                    30.662 &                    30.057 &                    29.812 &                    26.904 &                    26.628 &                    26.181 &                    26.238 &                    25.998 &                    25.659 &                    29.034 &                    27.489 &                    26.617 &                    24.633 &                    24.010 &                    23.566 \\
\rowcolor{gray!15}[\dimexpr\tabcolsep + 1pt\relax]  DASR (pretrained) &                    30.545 &                    30.463 &                    29.701 &                    26.829 &                    26.701 &                    26.060 &                    26.200 &                    26.178 &                    25.699 &                    28.865 &                    28.844 &                    27.858 &                    24.487 &                    24.280 &                    23.624 \\
\rowcolor{gray!15}[\dimexpr\tabcolsep + 1pt\relax] DANv1 (pretrained) &                      30.807 &  \textcolor{blue}{30.739} &  \textcolor{blue}{30.049} &  \textcolor{blue}{26.983} &  \textcolor{blue}{26.925} &  \textcolor{blue}{26.360} &  \textcolor{blue}{26.309} &  \textcolor{blue}{26.329} &  \textcolor{blue}{25.873} &  \textcolor{blue}{29.230} &  \textcolor{blue}{29.549} &   \textcolor{red}{28.664} &  \textcolor{blue}{24.897} &  \textcolor{blue}{24.817} &  \textcolor{blue}{24.127}  \\
\rowcolor{gray!15}[\dimexpr\tabcolsep + 1pt\relax] DANv2 (pretrained) &   \textcolor{red}{30.850} &   \textcolor{red}{30.881} &                    30.042 &   \textcolor{red}{27.033} &   \textcolor{red}{26.999} &                    26.314 &   \textcolor{red}{26.349} &   \textcolor{red}{26.392} &                    25.832 &   \textcolor{red}{29.313} &   \textcolor{red}{29.550} &  \textcolor{blue}{28.592} &   \textcolor{red}{25.048} &   \textcolor{red}{24.926} &   \textcolor{red}{24.127} \\
                 \rowcolor{gray!5}[\dimexpr\tabcolsep + 1pt\relax]           \textbf{Non-Blind} &&&&&&&&&&&&&&& \\
 \rowcolor{gray!5}[\dimexpr\tabcolsep + 1pt\relax]                  RCAN-MA (true sigma) &   \textcolor{red}{30.956} &   \textcolor{red}{30.973} &   \textcolor{red}{29.880} &   \textcolor{red}{27.091} &   \textcolor{red}{27.094} &  \textcolor{blue}{26.383} &  \textcolor{blue}{26.389} &  \textcolor{blue}{26.440} &  \textcolor{blue}{26.002} &  \textcolor{blue}{29.529} &  \textcolor{blue}{29.834} &  \textcolor{blue}{28.799} &  \textcolor{blue}{25.059} &  \textcolor{blue}{25.008} &  \textcolor{blue}{24.218} \\
  \rowcolor{gray!5}[\dimexpr\tabcolsep + 1pt\relax]  HAN-MA (true sigma) &  \textcolor{blue}{30.940} &  \textcolor{blue}{30.905} &  \textcolor{blue}{29.834} &  \textcolor{blue}{27.029} &  \textcolor{blue}{27.067} &   \textcolor{red}{26.387} &   \textcolor{red}{26.391} &   \textcolor{red}{26.444} &   \textcolor{red}{26.003} &   \textcolor{red}{29.543} &   \textcolor{red}{29.844} &   \textcolor{red}{28.832} &   \textcolor{red}{25.089} &   \textcolor{red}{25.021} &   \textcolor{red}{24.234} \\
 \rowcolor{gray!5}[\dimexpr\tabcolsep + 1pt\relax] RCAN-MA (noisy sigma) &                   30.838 &                    30.700 &                    29.780 &                    27.023 &                    26.966 &                    26.168 &                    26.354 &                    26.386 &                    25.905 &                    29.276 &                    29.373 &                    28.142 &                    24.959 &                    24.839 &                    23.933 \\
\rowcolor{gray!15}[\dimexpr\tabcolsep + 1pt\relax]           \textbf{RCAN/DAN} &&&&&&&&&&&&&&& \\
 \rowcolor{gray!15}[\dimexpr\tabcolsep + 1pt\relax]               DANv1 &                    30.432 &                    30.387 &                    29.436 &                    26.773 &                    26.754 &                    26.022 &                    26.178 &                    26.212 &                    25.762 &                    28.709 &                    28.937 &                    27.656 &                    24.378 &                    24.326 &                    23.583 \\
 \rowcolor{gray!15}[\dimexpr\tabcolsep + 1pt\relax]      DANv1 (cosine) &                    30.627 &                    30.466 &                    29.537 &                    26.858 &                    26.855 &                    26.101 &                    26.270 &                    26.296 &                    25.831 &                    29.037 &                    29.227 &                    27.908 &                    24.657 &                    24.568 &                    23.735 \\
 \rowcolor{gray!15}[\dimexpr\tabcolsep + 1pt\relax] RCAN (batch size 4) &                    30.686 &                    30.538 &                    29.646 &                    26.957 &                    26.966 &                    26.198 &                    26.324 &                    26.376 &                    25.907 &                    29.172 &                    29.395 &                    28.160 &  \textcolor{blue}{24.893} &                    24.846 &                    23.943 \\
 \rowcolor{gray!15}[\dimexpr\tabcolsep + 1pt\relax]            RCAN-DAN &   \textcolor{red}{30.813} &  \textcolor{blue}{30.627} &  \textcolor{blue}{29.741} &  \textcolor{blue}{26.997} &  \textcolor{blue}{26.996} &  \textcolor{blue}{26.208} &   \textcolor{red}{26.348} &  \textcolor{blue}{26.387} &  \textcolor{blue}{25.912} &   \textcolor{red}{29.303} &   \textcolor{red}{29.509} &   \textcolor{red}{28.280} &   \textcolor{red}{24.940} &   \textcolor{red}{24.876} &   \textcolor{red}{23.979} \\
 \rowcolor{gray!15}[\dimexpr\tabcolsep + 1pt\relax]    RCAN-DAN (sigma) &  \textcolor{blue}{30.747} &   \textcolor{red}{30.666} &   \textcolor{red}{29.782} &   \textcolor{red}{27.013} &   \textcolor{red}{27.014} &   \textcolor{red}{26.235} &  \textcolor{blue}{26.340} &   \textcolor{red}{26.391} &   \textcolor{red}{25.924} &  \textcolor{blue}{29.284} &  \textcolor{blue}{29.462} &  \textcolor{blue}{28.275} &                    24.798 &  \textcolor{blue}{24.874} &  \textcolor{blue}{23.958} \\
    \rowcolor{gray!5}[\dimexpr\tabcolsep + 1pt\relax]           \textbf{HAN/DAN} &&&&&&&&&&&&&&& \\
     \rowcolor{gray!5}[\dimexpr\tabcolsep + 1pt\relax]               HAN (batch size 4) &  \textcolor{blue}{30.779} &   \textcolor{red}{30.620} &  \textcolor{blue}{29.572} &  \textcolor{blue}{26.998} &   \textcolor{red}{27.002} &  \textcolor{blue}{26.174} &  \textcolor{blue}{26.341} &  \textcolor{blue}{26.383} &  \textcolor{blue}{25.897} &  \textcolor{blue}{29.226} &  \textcolor{blue}{29.448} &  \textcolor{blue}{28.013} &  \textcolor{blue}{24.922} &  \textcolor{blue}{24.845} &  \textcolor{blue}{23.877} \\
      \rowcolor{gray!5}[\dimexpr\tabcolsep + 1pt\relax]       HAN-DAN &   \textcolor{red}{30.817} &  \textcolor{blue}{30.567} &   \textcolor{red}{29.707} &   \textcolor{red}{27.021} &  \textcolor{blue}{26.993} &   \textcolor{red}{26.221} &   \textcolor{red}{26.356} &   \textcolor{red}{26.388} &   \textcolor{red}{25.908} &   \textcolor{red}{29.289} &   \textcolor{red}{29.477} &   \textcolor{red}{28.287} &   \textcolor{red}{24.943} &   \textcolor{red}{24.899} &   \textcolor{red}{23.995} \\
                             \rowcolor{gray!15}[\dimexpr\tabcolsep + 1pt\relax]           \textbf{RCAN/Contrastive} &&&&&&&&&&&&&&& \\
    \rowcolor{gray!15}[\dimexpr\tabcolsep + 1pt\relax]  RCAN (batch size 8) &                    30.675 &                    30.484 &                    29.635 &                    27.007 &                    27.003 &                    26.149 &                    26.337 &                    26.379 &                    25.898 &                    29.206 &                    29.406 &                    27.956 &   \textcolor{red}{24.962} &   \textcolor{red}{24.899} &  \textcolor{blue}{23.960} \\
    \rowcolor{gray!15}[\dimexpr\tabcolsep + 1pt\relax]            RCAN-MoCo &   \textcolor{red}{30.870} &                    30.677 &                    29.714 &   \textcolor{red}{27.032} &                    26.980 &  \textcolor{blue}{26.208} &                    26.338 &                    26.377 &                    25.918 &                    29.210 &                    29.394 &                    28.243 &                    24.919 &                    24.850 &                    23.952 \\
   \rowcolor{gray!15}[\dimexpr\tabcolsep + 1pt\relax]          RCAN-SupMoCo &                    30.819 &  \textcolor{blue}{30.700} &  \textcolor{blue}{29.720} &                    27.019 &                    27.005 &                    26.185 &   \textcolor{red}{26.350} &   \textcolor{red}{26.397} &                    25.910 &  \textcolor{blue}{29.257} &  \textcolor{blue}{29.437} &                    28.216 &  \textcolor{blue}{24.958} &                    24.876 &   \textcolor{red}{23.961} \\
   \rowcolor{gray!15}[\dimexpr\tabcolsep + 1pt\relax]       RCAN-regression &                    30.730 &                    30.618 &                    29.642 &                    26.955 &                    26.986 &                    26.206 &                    26.336 &                    26.384 &                    25.913 &                    29.188 &                    29.421 &   \textcolor{red}{28.284} &                    24.911 &                    24.819 &                    23.906 \\
\rowcolor{gray!15}[\dimexpr\tabcolsep + 1pt\relax]  RCAN-SupMoCo-regression &  \textcolor{blue}{30.861} &   \textcolor{red}{30.785} &                    29.676 &                    27.015 &   \textcolor{red}{27.031} &                    26.177 &  \textcolor{blue}{26.348} &  \textcolor{blue}{26.389} &   \textcolor{red}{25.931} &   \textcolor{red}{29.302} &   \textcolor{red}{29.491} &  \textcolor{blue}{28.262} &                    24.950 &  \textcolor{blue}{24.877} &                    23.959 \\
 \rowcolor{gray!15}[\dimexpr\tabcolsep + 1pt\relax]            RCAN-WeakCon &                    30.741 &                    30.600 &                    29.720 &  \textcolor{blue}{27.021} &  \textcolor{blue}{27.011} &   \textcolor{red}{26.224} &                    26.342 &                    26.387 &  \textcolor{blue}{25.924} &                    29.204 &                    29.433 &                    28.189 &                    24.913 &                    24.862 &                    23.955 \\
 \rowcolor{gray!15}[\dimexpr\tabcolsep + 1pt\relax]   RCAN-SupMoCo (online) &                    30.777 &                    30.695 &   \textcolor{red}{29.771} &                    27.013 &                    26.992 &                    25.931 &                    26.336 &                    26.389 &                    25.915 &                    29.220 &                    29.401 &                    28.189 &                    24.915 &                    24.850 &                    23.890 \\
\rowcolor{gray!15}[\dimexpr\tabcolsep + 1pt\relax]    RCAN-SupMoCo (ResNet) &                    30.712 &                    30.604 &                    29.658 &                    26.974 &                    26.984 &                    26.103 &                    26.317 &                    26.363 &                    25.869 &                    29.221 &                    29.422 &                    27.266 &                    24.832 &                    24.785 &                    23.704 \\
  \rowcolor{gray!5}[\dimexpr\tabcolsep + 1pt\relax]           \textbf{HAN/Contrastive} &&&&&&&&&&&&&&& \\
   \rowcolor{gray!5}[\dimexpr\tabcolsep + 1pt\relax]         HAN (batch size 8) &  \textcolor{blue}{30.705} &  \textcolor{blue}{30.592} &  \textcolor{blue}{29.593} &   \textcolor{red}{27.006} &   \textcolor{red}{26.993} &  \textcolor{blue}{26.142} &  \textcolor{blue}{26.342} &   \textcolor{red}{26.394} &  \textcolor{blue}{25.910} &   \textcolor{red}{29.259} &   \textcolor{red}{29.498} &  \textcolor{blue}{28.053} &   \textcolor{red}{24.927} &   \textcolor{red}{24.895} &  \textcolor{blue}{23.902} \\
 \rowcolor{gray!5}[\dimexpr\tabcolsep + 1pt\relax] HAN-SupMoCo-regression &   \textcolor{red}{30.734} &   \textcolor{red}{30.659} &   \textcolor{red}{29.741} &  \textcolor{blue}{27.005} &  \textcolor{blue}{26.983} &   \textcolor{red}{26.192} &   \textcolor{red}{26.343} &  \textcolor{blue}{26.376} &   \textcolor{red}{25.913} &  \textcolor{blue}{29.195} &  \textcolor{blue}{29.381} &   \textcolor{red}{28.278} &  \textcolor{blue}{24.926} &  \textcolor{blue}{24.839} &   \textcolor{red}{23.955} \\
\rowcolor{gray!15}[\dimexpr\tabcolsep + 1pt\relax] \textbf{Extensions} &&&&&&&&&&&&&&& \\
 \rowcolor{gray!15}[\dimexpr\tabcolsep + 1pt\relax]     RCAN-DAN (pretrained estimator) &  \textcolor{blue}{30.763} &                    30.612 &  \textcolor{blue}{29.711} &   \textcolor{red}{27.036} &                    26.988 &  \textcolor{blue}{26.202} &   \textcolor{red}{26.355} &  \textcolor{blue}{26.394} &  \textcolor{blue}{25.919} &   \textcolor{red}{29.371} &   \textcolor{red}{29.564} &  \textcolor{blue}{28.239} &  \textcolor{blue}{24.971} &                    24.888 &  \textcolor{blue}{23.980} \\
\rowcolor{gray!15}[\dimexpr\tabcolsep + 1pt\relax]       RCAN (batch size 8, long-term) &                    30.736 &   \textcolor{red}{30.699} &   \textcolor{red}{29.723} &                    27.011 &  \textcolor{blue}{27.018} &                    26.171 &                    26.343 &                    26.390 &                    25.914 &                    29.230 &                    29.491 &                    28.101 &                    24.962 &  \textcolor{blue}{24.899} &                    23.960 \\
\rowcolor{gray!15}[\dimexpr\tabcolsep + 1pt\relax]  RCAN-SupMoCo-regression (long-term) &   \textcolor{red}{30.832} &  \textcolor{blue}{30.640} &                    29.690 &  \textcolor{blue}{27.023} &   \textcolor{red}{27.019} &   \textcolor{red}{26.217} &  \textcolor{blue}{26.355} &   \textcolor{red}{26.411} &   \textcolor{red}{25.944} &  \textcolor{blue}{29.297} &  \textcolor{blue}{29.514} &   \textcolor{red}{28.375} &   \textcolor{red}{24.999} &   \textcolor{red}{24.929} &   \textcolor{red}{24.007} \\
\bottomrule
            \end{tabularx}
		\end{center}
    \end{adjustwidth}
\label{simple_blind_psnr}
\end{table}
With reference to the model categories highlighted in Table \ref{simple_blind_psnr}, we make the following observations: 
\begin{itemize}
    \item \textbf{Pretrained models}:  We provide the results for the pretrained models evaluated in Figure \ref{fig:simple_regression}B, along with the results for the pretrained \gls{DASR} \cite{DASR21}, a blind \gls{SR} model with a MoCo-based contrastive encoder.  The \gls{DAN} models have the best results in most cases (with DANv2 having the best performance overall).  For \gls{IKC}, another iterative model, we present two sets of metrics:  \gls{IKC} (pretrained-best-iter) shows the results obtained when selecting the best image from all \gls{SR} output iterations (7 in total), as is the implementation in the official \gls{IKC} codebase.  \gls{IKC} (pretrained-last-iter) shows the results obtained when selecting the image from the last iteration (as is done for the \gls{DAN} models).  The former method produces the best results (even surpassing \gls{DAN} in some cases), but cannot be applied in true blind scenarios where a reference \gls{HR} image is not available.  
    \item \textbf{Non-blind models}:  The non-blind models fed with the true $\sigma$ achieve the best performance of all models studied.  This is true both for \gls{RCAN} and \gls{HAN}, with \gls{HAN} having a slight edge overall.  The wide margin over all other blind models clearly shows that significantly improved performance is possible if the degradation prediction system can be improved.  We also trained and tested a model (\gls{RCAN}-\gls{MA} (noisy sigma)) which was provided with the normalised $\sigma$ values corrupted by noise (mean 0, standard deviation 0.1).  This error level is slightly higher than that of the \gls{DAN} models tested (Figure \ref{fig:simple_regression}A), allowing this model to act as a performance reference for our estimation methods. 
    \item \textbf{RCAN-DAN models}:  As observed in Figure \ref{fig:simple_regression}, the DAN models that were trained from scratch are significantly worse than the \gls{RCAN} models, including the fully non-blind \gls{RCAN}, across all datasets (Figure \ref{fig:bar_chart_simple}) and $\sigma$ values (Figure \ref{fig:psnr_vs_sigma}).  The \gls{RCAN}-\gls{DAN} models show a consistent performance boost over \gls{RCAN} across the board.  As noted earlier in Section \ref{metadata_insertion_results}, predicting \gls{PCA}-reduced kernels appears to provide no advantage over directly predicting $\sigma$. 
    \item \textbf{RCAN-Contrastive models}:  For the contrastive models, the results are much less clear-cut.  The different contrastive blind models exhibit superior performance to \gls{RCAN} under most conditions (except for Urban100), but none of the algorithms tested (\gls{moco}, \gls{supmoco}, WeakCon and direct regression) seem to provide any particular advantage over each other.  The encoder trained with combined regression and \gls{supmoco} appears to provide a slight boost over the other techniques (Figure \ref{fig:bar_chart_simple}), but this is not consistent across the datasets and $\sigma$ values analysed.  This is a surprising result given that the clear clusters formed by \gls{supmoco}/WeakCon (as shown in Figure \ref{fig:simple_tsne}) would have been expected to improve the encoders' predictive power.  We hypothesise that the encoded representation is difficult for even deep learning models to interpret, and a clear-cut route from the encoded vector to the actual blur $\sigma$ is difficult to produce.  We also observe that both the \gls{RCAN}-\gls{DAN} and \gls{RCAN}-\gls{supmoco} models clearly surpass the noisy sigma non-blind \gls{RCAN} model on datasets with medium and high $\sigma$, while they perform slightly worse on datasets with low $\sigma$.  This matches the results in Figure \ref{fig:simple_regression}, where it is clear that the performance of all predictors suffer when the $\sigma$ is low.
    \item \textbf{HAN models}:  Upgraded \gls{HAN} models appear to follow similar trends as \gls{RCAN} models.  The inclusion of \gls{DAN} provides a clear boost in performance, but this time the inclusion of the \gls{supmoco}-regression predictor seems to only boost performance when the $\sigma$ is high.
    \item \textbf{Extensions}:  We also trained a \gls{RCAN}-\gls{DAN} model where we pre-initialised the predictor with that from the pretrained DANv1 model.  The minor improvements indicate that, for the most part, the predictor is achieving similar prediction accuracy to that of the pretrained models (as is also indicated in Figure \ref{fig:simple_regression}).  We also extended the training of the baseline \gls{RCAN} and the \gls{RCAN}-\gls{supmoco}-regression model to 2,200 epochs.  The expanded training continues to improve performance and, perhaps crucially, the contrastive model continues to show a margin of improvement over the baseline \gls{RCAN} model.  In fact, this extended model starts to achieve similar or better performance than the pretrained DAN models.  This is achieved with a significantly shorter training time ($2,200$ vs $\approx7,000$ epochs) and a fixed set of degradations, indicating that our models would surpass the performance of the pretrained \gls{DAN} models if trained with the same conditions.  
\end{itemize}

\begin{figure}[!ht]
    \begin{adjustwidth}{-\extralength}{0cm}
        \centering
        \includegraphics[width=0.8\linewidth]{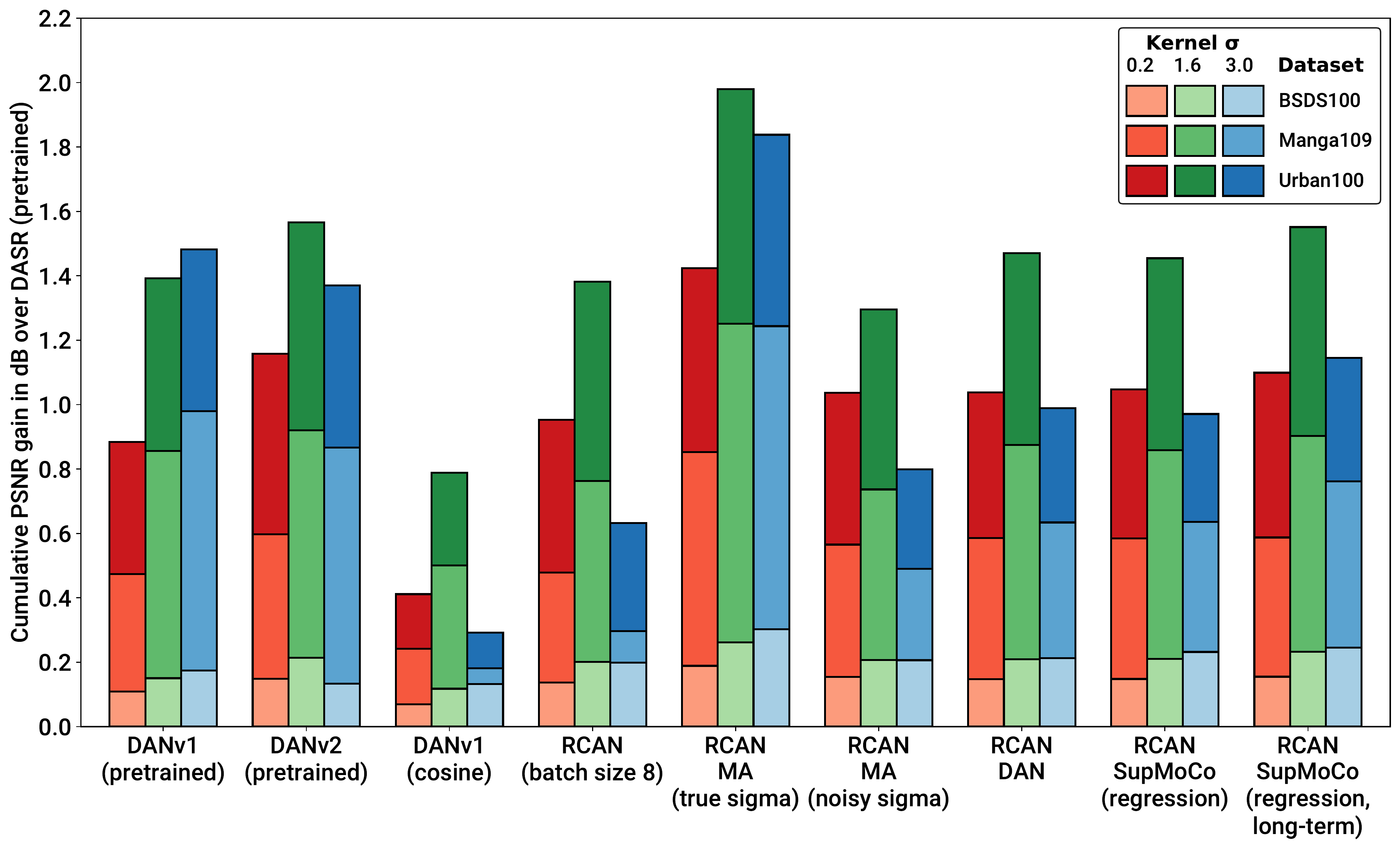}
    \end{adjustwidth}
    \caption{Bar graph showing improvement in PSNR over pretrained DASR for key models from Table \ref{simple_blind_psnr}.  PSNR improvements are stacked on each other for each specific $\sigma$ to show cumulative PSNR gain across datasets.  Our hybrid models can match and even surpass the pretrained DAN models, despite the large disparity in training time.}
    \label{fig:bar_chart_simple}
\end{figure}

Additionally, we implemented, trained and tested the \gls{Real-ESRGAN} and \gls{ELAN} models with the addition of the \gls{MA} metadata insertion block (with the same hyperparameters as presented in Section \ref{implementation_details}).  The testing results are available in the supplementary information (Table S3 containing \gls{Real-ESRGAN} \gls{LPIPS} results, and Tables S4 and S5 containing the \gls{PSNR} and \gls{SSIM} results for \gls{ELAN}, respectively).  For \gls{Real-ESRGAN}, the addition of the true metadata (non-blind) makes a clear improvement over the base model.  We also observed a consistent improvement in performance across datasets and $\sigma$ values for the \gls{DAN} upgraded model. However, attaching the best performing \gls{supmoco} encoder provided no clear advantage.  We hypothesise that the \gls{Real-ESRGAN} model is more sensitive to the accuracy of the kernel prediction, and thus sees limited benefit from the less accurate contrastive encoder (as we have shown for the \gls{DAN} vs contrastive methods (Figure \ref{fig:simple_regression})).  

For \gls{ELAN}, the baseline model is very weak, and is actually surpassed by Lanczos upsampling in one case (both in terms of \gls{PSNR} and \gls{SSIM}).  The addition of the true metadata only appeared to help when \gls{MA} was distributed through the whole network, upon which it increased the performance of the network massively (> 3dB in some cases).  It is clear that \gls{ELAN} is not performing well on these blurred datasets (\gls{ELAN} was originally tested only on bicubically downsampled datasets).  However, \gls{MA} still appears to be able to significantly improve the model's performance under the right conditions.  Further investigation would be required to first adapt \gls{ELAN} for such degraded datasets before attempting to use this model as part of our blind framework.    
\subsection{Complex Degradation Prediction}\label{complex_deg_prediction}
For our extended analysis on more realistic degradations, we trained three contrastive encoders (\gls{moco}, \gls{supmoco} and WeakCon) and one \gls{RCAN}-\gls{DAN} model on the complex pipeline dataset (Section \ref{implementation_details}). Given the large quantity of degradations, we devised a number of testing scenarios, each applied on the combined images of BSDS100, Manga109 and Urban100 (309 images total).  The scenarios we selected are detailed in Table \ref{complex_degradations_scenarios}.  We will refer to these testing sets for the rest of this analysis.   We evaluated the prediction capabilities of the contrastive and iterative models separately.  We purposefully limited the testing blur kernel shapes to isotropic/anisotropic Gaussians to simplify analysis.
\begin{table}[!htb]
	\caption{The different testing scenarios considered for the complex analysis.  Each scenario was applied on all images of the BSDS100, Manga109 and Urban100 datasets (309 images total).  Cases which include noise have double the amount of images (618) as the pipeline was applied twice; once with colour noise and once with grey noise.  The final scenario consists of every possible combination of the degradations considered (16 total combinations with 4,944 images total, including colour and grey noise).  For all cases, isotropic blurring was applied with a $\sigma$ of 2.0, anisotropic blurring was applied with a horizontal $\sigma$ of 2.0, a vertical $\sigma$ of 1.0 and a random rotation, Gaussian/Poisson noise were applied with a sigma/scale of 20.0/2.0 respectively, and JPEG/JM H.264 compression were applied with a quality factor/QPI of 60/30 respectively.  All scenarios also included $\times$4 bicubic downsampling inserted at the appropriate point (following the sequence in \Cref{eq:deg2}).}
 \renewcommand{\arraystretch}{1.3}
	    \begin{center}
     \tablesize{\small} 
            \begin{tabularx}{\linewidth}{P{3.4cm}P{2cm}P{2cm}P{2cm}c}
\toprule
\textbf{Test Scenario} & \textbf{Blurring} & \textbf{Noise} & \textbf{Compression} & \textbf{Total Images} \\
\midrule
JPEG & N/A & N/A & JPEG & 309\\
JM & N/A & N/A & JM H.264 & 309\\
Poisson & N/A & Poisson & N/A & 618\\
Gaussian & N/A  & Gaussian & N/A & 618\\
Iso & Isotropic & N/A & N/A & 309\\
Aniso & Anisotropic & N/A & N/A & 309   \\
Iso + Gaussian & Isotropic & Gaussian & N/A & 618\\
Gaussian + JPEG & N/A & Gaussian & JPEG & 618\\
Iso + Gaussian + JPEG & Isotropic & Gaussian & JPEG & 618\\
Aniso + Poisson + JM & Anisotropic & Poisson & JM H.264& 618\\
Iso/Aniso + Gaussian/Poisson + JPEG/JM & Iso \& Anisotropic & Gaussian \& Poisson & JPEG \& JM H.264 & 4944\\
\bottomrule
            \end{tabularx}
		\end{center}
\label{complex_degradations_scenarios}
\end{table}
\subsubsection{Contrastive Learning}
\begin{figure}[!htbp]
        
        \begin{adjustwidth}{-\extralength}{0cm}
\centering
        \includegraphics[width=0.8\linewidth]{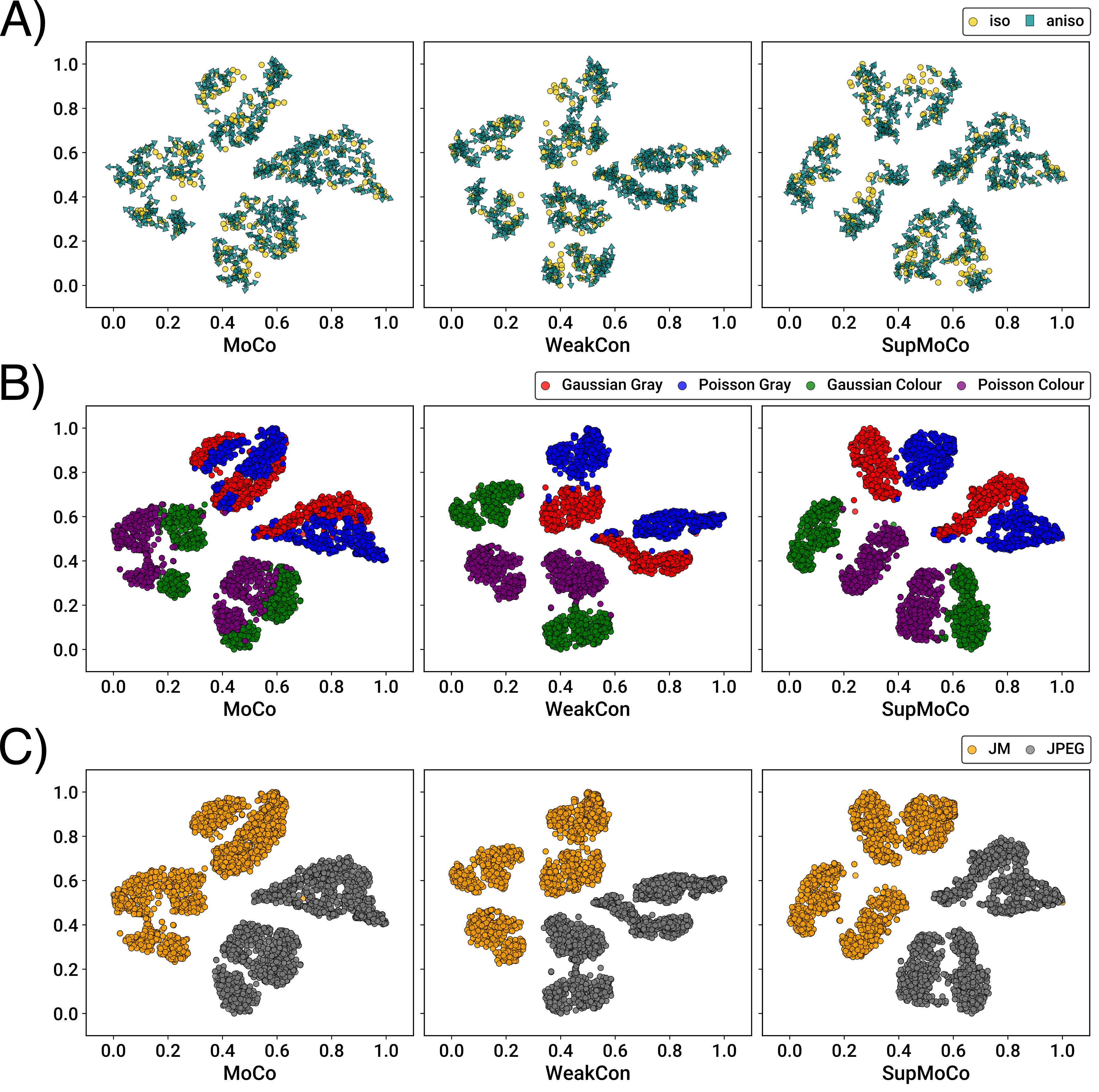}
\end{adjustwidth}
    \caption{t-SNE plots (perplexity value of 40) showing the separation power of the different contrastive learning algorithms considered on the complex pipeline.  All models were evaluated on the \textit{Iso/Aniso +
Gaussian/Poisson + JPEG/JM} testing scenario (4,944 images).  Each dimension was independently normalised in
the range $[0, 1]$ after computing the t-SNE results.  Each column shows the same data in each row, but each panel has a different shading for the indicated degradation types. A)  t-SNE plots with each point labelled according to the blur kernel applied. Only 560 images (randomly selected) are plotted for each panel, to reduce cluttering.  Arrows indicate the rotation of the anistropic kernels.  While a few clusters of isotropic and anisotropic kernels are present (especially for \gls{supmoco}), the blur kernels do not appear to be well classified for either of the encoders considered.  B) t-SNE plots with each point coloured according to the noise injected.  All three encoders are capable of separating the colour noise, but \gls{moco} struggles with the grey noise separation.  Both \gls{supmoco} and WeakCon also have a number of notable outliers for grey noise samples. C) t-SNE plots with each point coloured according to the compression applied.  All three encoders are capable of separating the compression types, with very clear clustering.}
    \label{fig:tsne_plot_complex}
\end{figure}

For each of the contrastive algorithms, we trained an encoder (all with the same architecture as used for the simple pipeline) with the following protocol:
\begin{itemize}
    \item We first pre-trained the encoder with an online pipeline of noise (same parameters as the full complex pipeline, but with an equal probability to select grey or colour noise) and bicubic downsampling.  We found that this pre-training helps reduce loss stagnation for the \gls{supmoco} encoder, so we applied this to all encoders.  The \gls{supmoco} encoder was trained with double precision at this stage.  We used 3 positive patches for \gls{supmoco} and 1 positive patch for both \gls{moco} and WeakCon.
    \item After 1,099 epochs, we started training the encoder on the full online complex pipeline (Section \ref{implementation_details}).  The \gls{supmoco} encoder was switched to triple precision from here onwards.
    \item We stopped all encoders after 2,001 total epochs, and evaluated them at this checkpoint.
    \item For \gls{supmoco}, the decision tree in Section \ref{supmoco_description} was used to assign class labels.  For WeakCon, $w_{ij}$ was computed as the Euclidean distance between query/negative sample vectors containing: the vertical and horizontal blur $\sigma$, the Gaussian/Poisson sigma/scale respectively and the JPEG/JM H.264 quality factor/QPI, respectively (6 elements total).  All values were normalised to $[0,1]$ prior to computation.
\end{itemize}
As with the simple pipeline, contrastive encodings are not directly interpretable and so we analysed the clustering capabilities of each encoder through \gls{TSNE} visualizations.  We evaluated each encoder on the full testing scenario (\textit{Iso/Aniso +
Gaussian/Poisson + JPEG/JM} in Table \ref{complex_degradations_scenarios}), and applied \gls{TSNE} independently for each model.  The results are shown in Figure \ref{fig:tsne_plot_complex}.

It is clear from the \gls{TSNE} plots that the clustering of the dataset is now significantly more complex than that observed in Figure \ref{fig:simple_tsne}.  However, all three encoders appear to have successfully learnt how to distinguish the two compression types and are also mostly successful when clustering the four types of noise (\gls{moco} is slightly weaker for grey noise).  In the supplementary information (Figure S1), we also show that the encoders are capable of separating different intensities of both compression and noise, albeit with less separation of the two noise types.  

For blurring, the separation between isotropic and anisotropic kernels is much less logical.  It appears that each encoder was attempting to form sub-clusters for each type of kernel in some cases (in particular for \gls{supmoco}) but the separation is significantly less clear cut than that obtained in Figure \ref{fig:simple_tsne}.  Further analysis would be required to decipher whether clustering is weak simply due to the difficulty of the exercise, or whether clustering is being mostly influenced by the other degradations considered in the pipeline.

As observed with the simple pipeline, it is again apparent that the different methods of semi-supervision seem to be converging to similar results.  This is also in spite of the fact that WeakCon was supplied with only 6 degradation elements while \gls{supmoco} was supplied with the full degradation metadata through its class system.  Further investigation into their learning process could reveal further insight into the effects of each algorithm.  

\subsubsection{Iterative Parameter Regression}

The \gls{RCAN}-\gls{DAN} model was trained on the complex pipeline dataset with identical hyperparameters to that of the simple pipeline.  For degradation prediction, we set the \gls{DAN} model to predict a vector with the following elements (15 total):
\begin{itemize}
    \item Individual elements for the following blur parameters: vertical and horizontal $\sigma$, rotation, individual $\beta$ for generalised Gaussian and plateau kernels and the sinc cutoff frequency. Whenever one of these elements was unused (e.g. cutoff frequency for Gaussian kernels), this was set to 0.  All elements were normalised to $[0, 1]$ according to their respective ranges (Section \ref{implementation_details}).
    \item Four boolean (0 or 1) elements categorising whether the kernel shape was:
    \begin{itemize}
        \item Isotropic or anisotropic
        \item Generalised 
        \item Plateau-type
        \item Sinc
    \end{itemize} 
    \item Individual elements for the Gaussian sigma and Poisson scale (both normalised to $[0,1]$).
    \item A boolean indicating whether the noise was colour or grey type.
    \item Individual elements for the JM H.264 QPI and JPEG quality factor (both normalised to $[0,1]$).
\end{itemize}

We tested the prediction accuracy by evaluating the model on a number of our testing scenarios, and then quantifying degradation prediction error.  The results are shown in Table \ref{complex_blind_dan_error}.  As observed with the contrastive models, blur kernel parameter prediction accuracy is extremely low, even when no other degradations are present.  On the other hand, both noise and compression prediction is significantly better, with sub 0.1 error in all cases, even when all degradations are present.  We hypothesise that since blur kernels are introduced as the first degradation in the pipeline, most of the blurring information could be masked by the time noise addition and compression have been applied.  

To the best of our knowledge, we are the first to present fully explicit blind degradation prediction on this complex pipeline.  We hope that the prediction results achieved in this analysis can act as a baseline from which further advances and improvements can be made. 

\begin{table}[!htbp]
\caption{Mean L1 error for the \gls{DAN} predictor within the \gls{RCAN}-\gls{DAN} model trained on the complex pipeline. Prediction accuracy is high for the compression and noise degradations, but weak for blur.  Blur parameters incorporate all the blur-specific numerical parameters and the four booleans.  Noise parameters incorporate the Gaussian/Poisson sigma/scale and colour/grey boolean.  Compression parameters incorporate the JPEG/JM H.264 quality factor/QPI.  The degradation scenarios tested here are described in Table~\ref{complex_degradations_scenarios}.}
  \setlength{\tabcolsep}{0.5pt}
	\renewcommand{\arraystretch}{1.3}
    \begin{center}
        \begin{tabularx}{\linewidth}{P{4.5cm}P{3cm}P{3.6cm}P{3cm}}
        \toprule
            \textbf{Test Scenario} & \textbf{Blurring} & \textbf{Noise} & \textbf{Compression}\\
            & ($\sigma$, kernel type) & (scale, gray/colour type) & (QPI/quality)\\
        \midrule
            \textbf{Iso/Aniso} & 0.318 & N/A & N/A\\
            \textbf{Gaussian/Poisson} & N/A & 0.042 & N/A\\
            \textbf{JPEG/JM} & N/A & N/A & 0.078\\
            \textbf{Iso/Aniso + Gaussian/Poisson + JPEG/JM} & 0.317 & 0.036 & 0.070 \\
        \bottomrule
        \end{tabularx}
    \end{center}
\label{complex_blind_dan_error}
\end{table}
\subsection{Blind SR on Complex Pipeline}\label{complex_sr}

For blind \gls{SR} on the complex pipeline, we focus on just \gls{RCAN} and \gls{RCAN} upgraded models to simplify analysis.  We use a single \gls{MA} block to insert metadata into the \gls{SR} core in all cases apart from one, where we distribute \gls{MA} throughout \gls{RCAN}.  We also trained a number of non-blind models (fed with different quantities of the correct metadata) as comparison points.  \gls{PSNR} \gls{SR} results comparing the baseline \gls{RCAN} to the blind models are provided in Table \ref{complex_blind_psnr} (Table S6 in the supplementary information provides the \gls{SSIM} results).

\begin{table}[!htbp]
	\caption{\gls{PSNR} (dB) \gls{SR} results on the complex pipeline comparing blind \gls{SR} methods.  For \gls{RCAN} models with metadata or degradation prediction, this is inserted using one \gls{MA} block at the front of the network (apart from \gls{RCAN} (\gls{supmoco}, all), where 200 independent \gls{MA} blocks are inserted throughout the network).  The degradation scenarios tested here are described in Table \ref{complex_degradations_scenarios}.  The non-blind models were fed with a vector containing the vertical/horizontal blur $\sigma$, the kernel type (7 possible shapes), the Gaussian/Poisson sigma/scale, a boolean indicating the addition of grey or colour noise, and the JPEG/JM H.264 quality factor/QPI.  Non-blind models marked with a degradation (e.g. no blur) have the marked degradation parameters removed from their input vector.  All values were normalised to $[0,1]$ apart from the kernel type.  The best result for each set is shown in \textcolor{red}{red}, while the second-best result is shown in \textcolor{blue}{blue}.}
  \setlength{\tabcolsep}{0.5pt}
	\renewcommand{\arraystretch}{1.3}
  	  \begin{adjustwidth}{-\extralength}{0cm}
	    \begin{center}
      \tablesize{\scriptsize} 
            \begin{tabularx}{\linewidth}{cP{3.2cm}P{1.2cm}P{1.2cm}P{1.2cm}P{1.2cm}P{1.2cm}P{1.2cm}P{1.2cm}P{1.2cm}P{1.2cm}P{1.2cm}P{1.4cm}}
\toprule
              \textbf{Dataset}     &         \textbf{Model} &                      \textbf{JPEG} &                        \textbf{JM} &                   \textbf{Poisson} &                  \textbf{Gaussian} &                       \textbf{Iso} &                     \textbf{Aniso} &            \textbf{Iso + Gaussian }&           \textbf{Gaussian + JPEG} &     \textbf{Iso + Gaussian + JPEG} &      \textbf{Aniso + Poisson + JM} & \textbf{Iso/Aniso + Gaussian/Poisson + JPEG/JM} \\
\midrule
          \rowcolor{gray!5}[\dimexpr\tabcolsep + 1pt\relax]                &     Bicubic &                    23.790 &                    23.843 &                    21.714 &                    21.830 &                    23.689 &                    24.014 &                    21.269 &                    21.588 &                    21.185 &                    21.376 &                                 21.266 \\
         \rowcolor{gray!5}[\dimexpr\tabcolsep + 1pt\relax]               &   Lanczos &                    23.831 &                    23.944 &                    21.435 &                    21.558 &                    23.845 &                    24.185 &                    21.034 &                    21.318 &                    20.969 &                    21.172 &                                 21.054 \\
      \rowcolor{gray!5}[\dimexpr\tabcolsep + 1pt\relax]      &  RCAN (batch size 4) &  \textcolor{blue}{24.443} &                    24.566 &                    23.847 &  \textcolor{blue}{23.910} &  \textcolor{blue}{25.416} &                    25.492 &                    23.331 &                    23.516 &                    23.023 &                    23.081 &                                 23.049 \\
    \rowcolor{gray!5}[\dimexpr\tabcolsep + 1pt\relax]       &    RCAN (batch size 8) &                    24.428 &                    24.541 &  \textcolor{blue}{23.868} &                    23.895 &                    25.382 &                    25.443 &                    23.315 &                    23.510 &                    23.026 &                    23.079 &                                 23.047 \\
     \rowcolor{gray!5}[\dimexpr\tabcolsep + 1pt\relax]       &      RCAN (non-blind) &                       N/A &                       N/A &                       N/A &                       N/A &                       N/A &                       N/A &                       N/A &                       N/A &   \textcolor{red}{23.052} &                    23.007 &                                 23.048 \\
   \rowcolor{gray!5}[\dimexpr\tabcolsep + 1pt\relax]    &  RCAN (non-blind, no blur) &                       N/A &                       N/A &                       N/A &                       N/A &                       N/A &                       N/A &                       N/A &  \textcolor{blue}{23.527} &                    23.039 &                    23.019 &                                 23.045 \\
  \rowcolor{gray!5}[\dimexpr\tabcolsep + 1pt\relax]    &  RCAN (non-blind, no noise) &                       N/A &                       N/A &                       N/A &                       N/A &                       N/A &                       N/A &                       N/A &                       N/A &                    22.997 &                    23.081 &                                 23.038 \\
\rowcolor{gray!5}[\dimexpr\tabcolsep + 1pt\relax] & RCAN (non-blind, no compression) &                       N/A &                       N/A &                       N/A &                       N/A &                       N/A &                       N/A &   \textcolor{red}{23.346} &                       N/A &  \textcolor{blue}{23.049} &                    23.041 &                                 23.033 \\   \rowcolor{gray!5}[\dimexpr\tabcolsep + 1pt\relax]    &    RCAN (MoCo) &                    24.388 &                    24.548 &                    23.823 &                    23.848 &                    25.246 &                    25.399 &                    23.311 &                    23.505 &                    23.022 &                    23.087 &                                 23.048 \\
          \rowcolor{gray!5}[\dimexpr\tabcolsep + 1pt\relax]       &   RCAN (WeakCon) &                    24.422 &                    24.532 &                    23.770 &                    23.634 &                    25.297 &                    25.418 &                    23.197 &                    23.510 &                    23.025 &  \textcolor{blue}{23.090} &                                 23.051 \\
        \rowcolor{gray!5}[\dimexpr\tabcolsep + 1pt\relax]        &    RCAN (SupMoCo) &                    24.390 &                    24.507 &                    23.710 &                    23.751 &                    25.254 &                    25.405 &                    23.237 &                    23.516 &                    23.023 &                    23.087 &                                 23.049 \\
      \rowcolor{gray!5}[\dimexpr\tabcolsep + 1pt\relax]      &   RCAN (SupMoCo, all) &                    24.412 &  \textcolor{blue}{24.572} &                    23.805 &                    23.873 &                    25.329 &                    25.431 &                    23.298 &   \textcolor{red}{23.530} &                    23.028 &                    23.088 &               \textcolor{blue}{23.054} \\
         \rowcolor{gray!5}[\dimexpr\tabcolsep + 1pt\relax]          &       RCAN-DAN &   \textcolor{red}{24.447} &   \textcolor{red}{24.589} &   \textcolor{red}{23.893} &   \textcolor{red}{23.920} &  25.412 &   \textcolor{blue}{25.509} &  \textcolor{blue}{23.343} &                    23.527 &                    23.033 &   \textcolor{red}{23.092} &                \textcolor{red}{23.058} \\
        \rowcolor{gray!5}[\dimexpr\tabcolsep + 1pt\relax] \multirow{-14}{*}{\rotatebox[origin=c]{0}{\textbf{BSDS100}}} & RCAN (trained on simple pipeline) &                    23.691 &                    24.059 &                    18.838 &                    18.946 &   \textcolor{red}{26.335} &   \textcolor{red}{25.733} &                    18.778 &                    19.053 &                    19.172 &                    18.803 &                                 18.955 \\
         \hdashline
         \rowcolor{gray!15}[\dimexpr\tabcolsep + 1pt\relax]      &  Bicubic &                    22.814 &                    22.983 &                    20.757 &                    21.418 &                    22.084 &                    22.587 &                    20.402 &                    21.170 &                    20.300 &                    20.120 &                                 20.201 \\
              \rowcolor{gray!15}[\dimexpr\tabcolsep + 1pt\relax]           &  Lanczos &                    22.980 &                    23.225 &                    20.605 &                    21.330 &                    22.329 &                    22.869 &                    20.334 &                    21.079 &                    20.245 &                    20.029 &                                 20.126 \\
        \rowcolor{gray!15}[\dimexpr\tabcolsep + 1pt\relax]    &   RCAN (batch size 4) &                    25.164 &                    25.419 &   \textcolor{red}{24.471} &  \textcolor{blue}{24.753} &                    26.387 &  \textcolor{blue}{26.433} &                    23.618 &                    23.987 &                    23.031 &                    23.104 &                                 23.060 \\
        \rowcolor{gray!15}[\dimexpr\tabcolsep + 1pt\relax]     &  RCAN (batch size 8) &                    25.198 &                    25.561 &  \textcolor{blue}{24.453} &                    24.733 &                    26.264 &                    26.270 &                    23.598 &                    23.982 &                    23.039 &                    23.104 &                                 23.062 \\
         \rowcolor{gray!15}[\dimexpr\tabcolsep + 1pt\relax]       &  RCAN (non-blind) &                       N/A &                       N/A &                       N/A &                       N/A &                       N/A &                       N/A &                       N/A &                       N/A &  \textcolor{blue}{23.140} &                    23.106 &                                 23.119 \\
    \rowcolor{gray!15}[\dimexpr\tabcolsep + 1pt\relax]    & RCAN (non-blind, no blur) &                       N/A &                       N/A &                       N/A &                       N/A &                       N/A &                       N/A &                       N/A &   \textcolor{red}{24.206} &                    23.087 &  \textcolor{blue}{23.136} &               \textcolor{blue}{23.121} \\
   \rowcolor{gray!15}[\dimexpr\tabcolsep + 1pt\relax]    & RCAN (non-blind, no noise) &                       N/A &                       N/A &                       N/A &                       N/A &                       N/A &                       N/A &                       N/A &                       N/A &                    23.099 &                    23.084 &                                 23.084 \\
\rowcolor{gray!15}[\dimexpr\tabcolsep + 1pt\relax]  & RCAN (non-blind, no compression) &                       N/A &                       N/A &                       N/A &                       N/A &                       N/A &                       N/A &   \textcolor{red}{23.722} &                       N/A &   \textcolor{red}{23.146} &                    23.122 &                \textcolor{red}{23.123} \\
              \rowcolor{gray!15}[\dimexpr\tabcolsep + 1pt\relax]      &   RCAN (MoCo) &                    24.961 &                    25.430 &                    24.255 &                    24.509 &                    25.878 &                    26.118 &                    23.599 &                    23.797 &                    23.023 &                    23.096 &                                 23.051 \\
              \rowcolor{gray!15}[\dimexpr\tabcolsep + 1pt\relax]    &  RCAN (WeakCon) &                    25.115 &                    25.498 &                    23.643 &                    23.838 &                    25.967 &                    26.102 &                    23.130 &                    23.748 &                    23.010 &                    23.083 &                                 23.037 \\
             \rowcolor{gray!15}[\dimexpr\tabcolsep + 1pt\relax]     &  RCAN (SupMoCo) &                    25.195 &                    25.612 &                    23.921 &                    24.570 &                    25.869 &                    26.114 &                    23.531 &                    24.025 &                    23.023 &                    23.107 &                                 23.056 \\
          \rowcolor{gray!15}[\dimexpr\tabcolsep + 1pt\relax]    & RCAN (SupMoCo, all) &   \textcolor{red}{25.335} &   \textcolor{red}{25.823} &                    24.329 &   \textcolor{red}{24.786} &                    25.963 &                    26.161 &                    23.479 &  \textcolor{blue}{24.122} &                    22.992 &                    23.084 &                                 23.035 \\
               \rowcolor{gray!15}[\dimexpr\tabcolsep + 1pt\relax]       &    RCAN-DAN &  \textcolor{blue}{25.315} &  \textcolor{blue}{25.770} &                    24.369 &                    24.715 &  \textcolor{blue}{26.447} &                    26.431 &  \textcolor{blue}{23.652} &                    24.051 &                    23.082 &   \textcolor{red}{23.140} &                                 23.098 \\
          \rowcolor{gray!15}[\dimexpr\tabcolsep + 1pt\relax]     \multirow{-14}{*}{\rotatebox[origin=c]{0}{\textbf{Manga109}}} &  RCAN (trained on simple pipeline) &                    23.148 &                    23.954 &                    18.443 &                    19.498 &   \textcolor{red}{29.329} &   \textcolor{red}{26.539} &                    18.746 &                    19.340 &                    19.056 &                    17.851 &                                 18.406 \\
                \hdashline
         \rowcolor{gray!5}[\dimexpr\tabcolsep + 1pt\relax]       & Bicubic &                    21.244 &                    21.353 &                    19.934 &                    20.073 &                    20.772 &                    21.140 &                    19.339 &                    19.864 &                    19.245 &                    19.468 &                                 19.345 \\
             \rowcolor{gray!5}[\dimexpr\tabcolsep + 1pt\relax]          &   Lanczos &                    21.332 &                    21.496 &                    19.787 &                    19.944 &                    20.939 &                    21.326 &                    19.239 &                    19.725 &                    19.150 &                    19.375 &                                 19.248 \\
        \rowcolor{gray!5}[\dimexpr\tabcolsep + 1pt\relax]     & RCAN (batch size 4) &  \textcolor{blue}{22.564} &  \textcolor{blue}{22.854} &                    22.201 &  \textcolor{blue}{22.263} &  23.130 &  23.236 &                    21.426 &                    21.814 &                    21.088 &                    21.356 &                                 21.214 \\
        \rowcolor{gray!5}[\dimexpr\tabcolsep + 1pt\relax]     & RCAN (batch size 8) &                    22.552 &                    22.840 &  \textcolor{blue}{22.214} &                    22.238 &                    23.115 &                    23.214 &                    21.418 &                    21.816 &                    21.099 &                    21.364 &                                 21.221 \\
        \rowcolor{gray!5}[\dimexpr\tabcolsep + 1pt\relax]     &    RCAN (non-blind) &                       N/A &                       N/A &                       N/A &                       N/A &                       N/A &                       N/A &                       N/A &                       N/A &   \textcolor{red}{21.165} &                    21.369 &                                 21.239 \\
    \rowcolor{gray!5}[\dimexpr\tabcolsep + 1pt\relax]  &  RCAN (non-blind, no blur) &                       N/A &                       N/A &                       N/A &                       N/A &                       N/A &                       N/A &                       N/A &   \textcolor{red}{21.855} &                    21.105 &                    21.380 &                                 21.210 \\
  \rowcolor{gray!5}[\dimexpr\tabcolsep + 1pt\relax]  &   RCAN (non-blind, no noise) &                       N/A &                       N/A &                       N/A &                       N/A &                       N/A &                       N/A &                       N/A &                       N/A &                    21.128 &                    21.359 &                                 21.203 \\
\rowcolor{gray!5}[\dimexpr\tabcolsep + 1pt\relax] & RCAN (non-blind, no compression) &                       N/A &                       N/A &                       N/A &                       N/A &                       N/A &                       N/A &   \textcolor{red}{21.494} &                       N/A &  \textcolor{blue}{21.159} &  \textcolor{blue}{21.389} &                \textcolor{red}{21.267} \\
               \rowcolor{gray!5}[\dimexpr\tabcolsep + 1pt\relax]     &  RCAN (MoCo) &                    22.559 &                    22.851 &                    22.159 &                    22.182 &                    22.934 &                    23.139 &                    21.398 &                    21.793 &                    21.099 &                    21.378 &                                 21.227 \\
             \rowcolor{gray!5}[\dimexpr\tabcolsep + 1pt\relax]    &  RCAN (WeakCon) &                    22.554 &                    22.847 &                    22.080 &                    21.805 &                    23.027 &                    23.175 &                    21.108 &                    21.809 &                    21.083 &                    21.372 &                                 21.215 \\
             \rowcolor{gray!5}[\dimexpr\tabcolsep + 1pt\relax]    &  RCAN (SupMoCo) &                    22.520 &                    22.768 &                    21.983 &                    22.024 &                    22.906 &                    23.116 &                    21.293 &                    21.813 &                    21.089 &                    21.374 &                                 21.221 \\
           \rowcolor{gray!5}[\dimexpr\tabcolsep + 1pt\relax]  & RCAN (SupMoCo, all) &                    22.541 &                    22.842 &                    22.117 &                    22.190 &                    23.029 &                    23.165 &                    21.327 &                    21.813 &                    21.063 &                    21.344 &                                 21.195 \\
             \rowcolor{gray!5}[\dimexpr\tabcolsep + 1pt\relax]        &  RCAN-DAN &   \textcolor{red}{22.596} &   \textcolor{red}{22.901} &   \textcolor{red}{22.273} &   \textcolor{red}{22.297} &   \textcolor{blue}{23.177} &   \textcolor{blue}{23.294} &  \textcolor{blue}{21.459} &  \textcolor{blue}{21.840} &                    21.125 &   \textcolor{red}{21.393} &               \textcolor{blue}{21.250} \\
      \rowcolor{gray!5}[\dimexpr\tabcolsep + 1pt\relax]  \multirow{-14}{*}{\rotatebox[origin=c]{0}{\textbf{Urban100}}}  &     RCAN (trained on simple pipeline) &                    21.291 &                    22.217 &                    17.835 &                    18.091 &   \textcolor{red}{24.717} &   \textcolor{red}{23.753} &                    17.676 &                    17.938 &                    17.894 &                    17.627 &                                 17.727  \\
\bottomrule
            \end{tabularx}
		\end{center}
  \end{adjustwidth}
\label{complex_blind_psnr}
\end{table}

We make the following observations on these results:
\begin{itemize}
    \item \textbf{Compression- and noise-only scenarios:}  In these scenarios, the \gls{RCAN}-\gls{DAN} model shows clear improvement over all other baseline and contrastive encoders (apart from some cases on Manga109).  Improvement is most significant in the compression scenarios. 
    \item \textbf{Blur-only scenarios:} Since the blurring scenarios are very similar or identical to the simple pipeline, the models from Table \ref{simple_blind_psnr} (\gls{RCAN} is also shown in Table \ref{complex_blind_psnr}) are significantly stronger.  The \gls{DAN} model overtakes the baseline in some cases, but is very inconsistent.
    \item \textbf{Multiple combinations:}  In the multiple degradation scenarios, the \gls{DAN} model consistently overtakes the baselines, but \gls{PSNR}/\gls{SSIM} increases are minimal. 
\end{itemize}

For all scenarios, there are a number of other surprising results.  The contrastive methods appear to be providing no benefit to \gls{SR} performance, in almost every case.  Furthermore, the non-blind models are often overtaken by the \gls{DAN} model in certain scenarios, and the amount of metadata available to the non-blind models does not appear to correlate with the final \gls{SR} performance.  It is clear that the metadata we have available for these degradations are having a much lesser impact on \gls{SR} performance than on the simple pipeline.  Since the contrastive encoders have shown to be slightly weaker than \gls{DAN} in the simple pipeline case (Figure \ref{fig:simple_regression}), it is clear that their limited prediction accuracy is also limiting potential gains in \gls{SR} performance on this pipeline.  This dataset is significantly more difficult than the simple case, not just due to the increased amount of degradations, but also as the models appear less receptive to the insertion of metadata.  We again hope that these results will act as a baseline for further exploration into complex blind \gls{SR}.

\subsection{Blind SR on Real LR Images}\label{real_data_tests}
As a final test to compare models from both pipelines, we ran a select number of models on real-world images from RealSRSet \cite{Zhang21}.  These results are shown in Figure \ref{fig:blind_sr}, with an additional image provided in the supplementary information (Figure S2).  This qualitative inspection clearly show that models trained on the complex pipeline are significantly better at dealing with real-world degradations than simple pipeline models.  Figure \ref{fig:blind_sr} shows that the complex pipeline models can remove compression artifacts, sharpen images and smoothen noise.  In particular, the dog image shows that \gls{RCAN}-\gls{DAN} can deal with noise more effectively than the baseline \gls{RCAN}.  The simple pipeline model results are all very similar to each other, as none of them are capable of dealing with degradations other than isotropic blurring.
\begin{figure}[!htbp]
        
            \begin{adjustwidth}{-\extralength}{0cm}
\centering
        \includegraphics[width=0.75\linewidth]{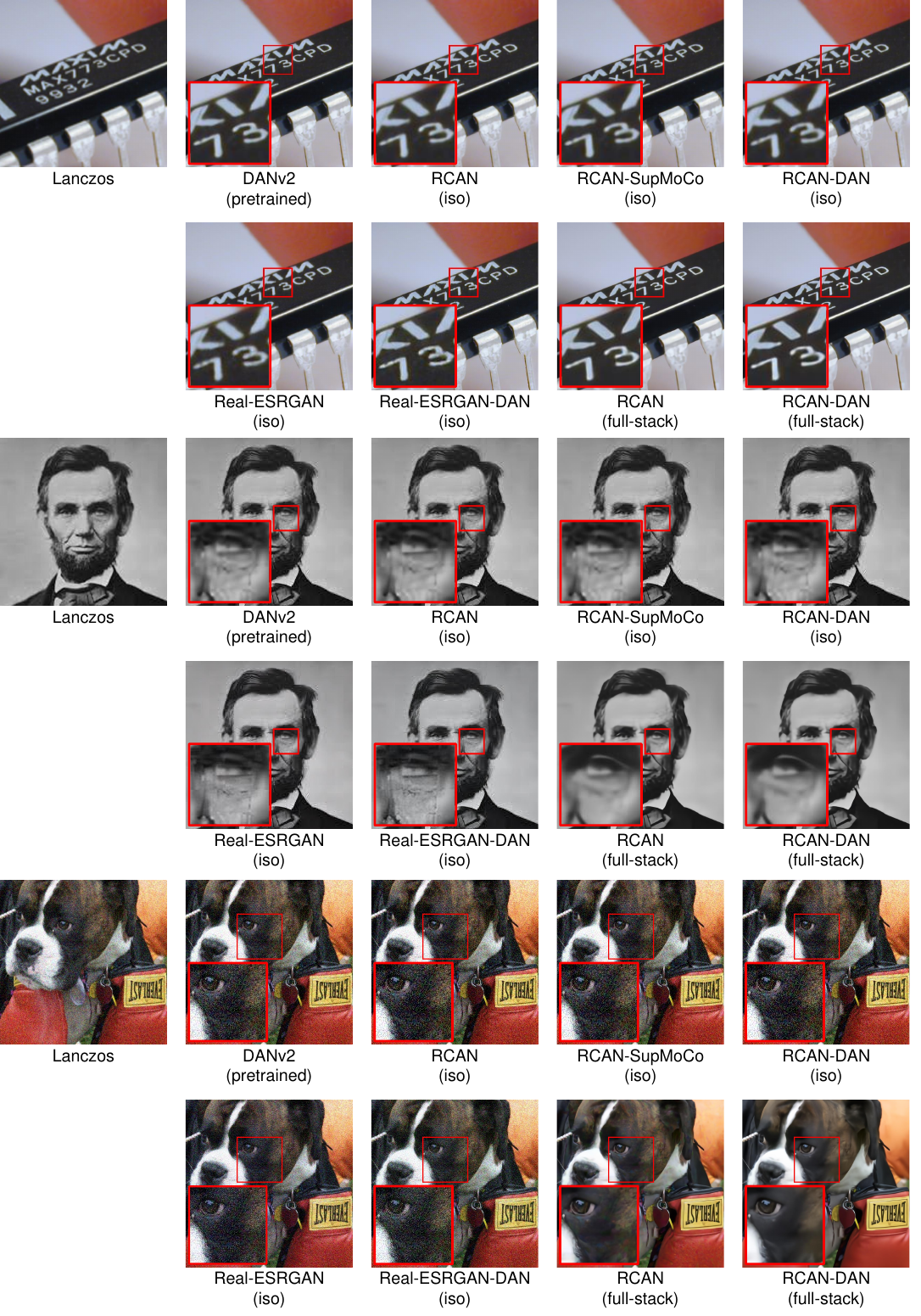}
\end{adjustwidth}
    \caption{Comparison of the SR results of various models on images from RealSRSet \cite{Zhang21}.  All simple pipeline models (marked as iso) and the pretrained \gls{DAN} model are incapable of dealing with degradations such as noise or compression. The complex pipeline models (marked as full-stack) produce significantly improved results.  These models can sharpen details (first image), remove compression artefacts (second image) and smooth over noise (third image).}
    \label{fig:blind_sr}
\end{figure}
\section{Conclusions}\label{sec:conc}

In this work, a framework for combining degradation prediction systems with any \gls{SR} network was proposed.  By using a single metadata insertion block to influence the feature maps of a convolutional layer, any degradation vector from a prediction model can, in many cases, be used to improve \gls{SR} network performance.  This premise was tested by implementing various contrastive and iterative degradation prediction mechanisms and coupling them with high-performing \gls{SR} architectures.  When tested on a dataset having images degraded with Gaussian blurring and downsampling, we show that our blind mechanisms achieve as much (or more) blur $\sigma$ prediction accuracy as the original methods, with significantly less training time.  Furthermore, both blind degradation performance (in combined training cases, such as with \gls{DAN}) and \gls{SR} performance is substantially elevated through the use of larger and stronger networks such as \gls{RCAN} \cite{rcan} or \gls{HAN} \cite{han}. Our results show that our hybrid models surpass the performance of the baseline non-blind and blind models under the same conditions.  
Other \gls{SR} architecture categories such as the \gls{SOTA} perceptual-loss based \gls{Real-ESRGAN} \cite{Wang21} and the transformer-based \gls{ELAN} architecture \cite{elan} could also work within our framework, but the performance of these methods is more sensitive to the accuracy of the degradation prediction and the dataset used for training. We show that this premise also holds true for blind \gls{SR} of a more complex pipeline involving various blurring, noise addition and compression operations.

Our framework should enable blind \gls{SR} research to be significantly expedited, as researchers can now focus on their degradation prediction mechanisms, rather than having to derive a custom \gls{SR} architecture for each new method.  There are various future avenues that could be explored to further assess the applications of our framework.  Apart from testing out new combinations of blind prediction, metadata insertion and \gls{SR} architectures, our framework could also be applied to new types of metadata.  For example, blind prediction systems could be swapped out for image classification systems, informing the \gls{SR} architecture with details on the image content (e.g. facial features for face \gls{SR} \cite{Xin18b}).  Furthermore, the framework could be extended to video \gls{SR} \cite{Liu22} where additional sources of metadata are available, such as the number of frames to be used in the super-resolution of a given frame and other details on the compression scheme, such as P- and B-frames (in addition to I-frames as considered in this work).
\vspace{6pt} 

\supplementary{The following supporting information can be downloaded at:  \linksupplementary{s1},  Table S1: SSIM SR results for metadata insertion block comparison; Table S2: SSIM SR results for simple pipeline comparison; Table S3: LPIPS SR results for Real-ESRGAN models on the simple pipeline; Table S4: PSNR results for ELAN models on the simple pipeline; Table S5: SSIM results for ELAN models on the simple pipeline; Table S6: SSIM SR results for complex pipeline comparison; Figure S1: additional t-SNE plots for complex pipeline contrastive encoders; Figure S2: Additional blind SR results on RealSRSet.}

\authorcontributions{Conceptualization, all authors; methodology, M.A. and K.G.C; software, M.A., K.G.C. and C.G.; validation, M.A. and K.G.C.; formal analysis, M.A. and K.G.C.; investigation, M.A. and K.G.C.; resources, K.P.C., R.A.F. and J.A.; data curation, M.A. and K.G.C.; writing---original draft preparation, M.A., K.G.C. and C.G.; writing---review and editing, all authors; visualization, M.A. and K.G.C.; supervision, K.P.C., R.A.F., J.A.; project administration, M.A., K.P.C. and J.A.; funding acquisition, R.A.F. All authors have read and agreed to the published version of the manuscript.}

\funding{This research work forms part of the Deep-FIR project, which is financed by the Malta Council for Science \& Technology (MCST), for and on behalf of the Foundation for Science \& Technology, through the FUSION: R\&I Technology Development Programme, grant number R\&I-2017-002-T.}
 
\institutionalreview{Not applicable.}

\informedconsent{Not applicable.}

\dataavailability{All code, data and model weights for the analysis presented in this paper are available here: \url{https://github.com/um-dsrg/RUMpy}} 

\conflictsofinterest{The authors declare no conflict of interest.} 

\begin{adjustwidth}{-\extralength}{0cm}

\reftitle{References}
\bibliography{bibliography}

%
\end{adjustwidth}
\end{document}


\title{\huge Supplementary Material for `The Best of Both Worlds: a Framework for Combining Degradation Prediction with High Performance Super-Resolution Networks'}
\author{Matthew Aquilina, Keith G. Ciantar, Christian Galea, Kenneth P. Camilleri, Reuben A. Farrugia, John Abela}
\maketitle
\vspace{-0.7cm}
\noindent This document contains a number of supplementary results and figures which support the main text.  Please refer to the main manuscript for further details.
\section*{\normalfont \Large \textbf{Simple Pipeline Results}}
\begin{table}[H]
	\caption{SSIM SR results corresponding to Table 1 from the main paper. 'low' refers to a $\sigma$ of 0.2, `med' refers to a $\sigma$ of 1.6 and `high' refers to a $\sigma$ of 3.0. The best result for each set is shown in \textcolor{red}{red}, while the second-best result is shown in \textcolor{blue}{blue}.}
  \setlength{\tabcolsep}{4pt}

	\renewcommand{\arraystretch}{1.3}
	    \begin{center}
            \begin{tabularx}{\linewidth}{cccccccccccccccc}
\toprule
         \textbf{Model} & \multicolumn{3}{c}{\textbf{Set5}} & \multicolumn{3}{c}{\textbf{Set14}} & \multicolumn{3}{c}{\textbf{BSDS100}} & \multicolumn{3}{c}{\textbf{Manga109}} & \multicolumn{3}{c}{\textbf{Urban100}} \\
               &                       \textbf{low} &                       \textbf{med} &                      \textbf{high} &                       \textbf{low} &                       \textbf{med} &                      \textbf{high} &                       \textbf{low} &                       \textbf{med} &                      \textbf{high} &                       \textbf{low} &                       \textbf{med} &                      \textbf{high} &                       \textbf{low} &                       \textbf{med} &                      \textbf{high} \\
\midrule
\rowcolor{gray!5}[\dimexpr\tabcolsep + 1pt\relax] \textbf{Baselines} &&&&&&&&&&&&&&& \\
    \rowcolor{gray!5}[\dimexpr\tabcolsep + 1pt\relax]        Bicubic &                    0.7909 &                    0.7513 &                    0.6736 &                    0.6803 &                    0.6384 &                    0.5700 &                    0.6452 &                    0.6058 &                    0.5458 &                    0.7716 &                    0.7323 &                    0.6659 &                    0.6376 &                    0.5941 &                    0.5240 \\
\rowcolor{gray!5}[\dimexpr\tabcolsep + 1pt\relax]        Lanczos &                    0.7986 &                    0.7625 &                    0.6814 &                    0.6888 &                    0.6487 &                    0.5761 &                    0.6534 &                    0.6149 &                    0.5510 &                    0.7775 &                    0.7418 &                    0.6720 &                    0.6452 &                    0.6044 &                    0.5302 \\
\rowcolor{gray!15}[\dimexpr\tabcolsep + 1pt\relax] \textbf{Non-Blind} &&&&&&&&&&&&&&& \\
   \rowcolor{gray!15}[\dimexpr\tabcolsep + 1pt\relax]        RCAN &                    0.8830 &                    0.8783 &                    0.8564 &                    0.7664 &                    0.7632 &                    0.7314 &                    0.7217 &                    0.7170 &                    0.6897 &                    0.9015 &                    0.9000 &                    0.8728 &                    0.7834 &                    0.7760 &                    0.7357 \\
  \rowcolor{gray!15}[\dimexpr\tabcolsep + 1pt\relax]           MA &                    0.8845 &  \textcolor{blue}{0.8837} &                    0.8615 &                    0.7671 &                    0.7669 &  \textcolor{blue}{0.7408} &                    0.7221 &                    0.7212 &                    0.6953 &                    0.9041 &                    0.9047 &                    0.8857 &                    0.7849 &                    0.7812 &                    0.7479 \\
 \rowcolor{gray!15}[\dimexpr\tabcolsep + 1pt\relax]     MA (all) &                    0.8836 &                    0.8828 &                    0.8620 &                    0.7667 &                    0.7670 &                    0.7407 &                    0.7216 &                    0.7213 &   \textcolor{red}{0.6958} &                    0.9036 &  \textcolor{blue}{0.9051} &  \textcolor{blue}{0.8863} &                    0.7837 &                    0.7808 &                    0.7483 \\
 \rowcolor{gray!15}[\dimexpr\tabcolsep + 1pt\relax]       MA (PCA) &                    0.8846 &                    0.8834 &                    0.8617 &  \textcolor{blue}{0.7677} &                    0.7667 &                    0.7401 &  \textcolor{blue}{0.7224} &                    0.7214 &                    0.6951 &                    0.9040 &                    0.9048 &                    0.8852 &  \textcolor{blue}{0.7852} &                    0.7812 &                    0.7475 \\
 \rowcolor{gray!15}[\dimexpr\tabcolsep + 1pt\relax]         SRMD &                    0.8844 &                    0.8836 &                    0.8622 &                    0.7673 &                    0.7670 &                    0.7406 &                    0.7222 &                    0.7212 &                    0.6956 &                    0.9039 &                    0.9048 &                    0.8855 &                    0.7844 &                    0.7809 &                    0.7477 \\
  \rowcolor{gray!15}[\dimexpr\tabcolsep + 1pt\relax]         SFT &                    0.8847 &                    0.8832 &  \textcolor{blue}{0.8627} &   \textcolor{red}{0.7677} &  \textcolor{blue}{0.7671} &                    0.7406 &   \textcolor{red}{0.7228} &                    0.7215 &                    0.6957 &   \textcolor{red}{0.9043} &   \textcolor{red}{0.9052} &   \textcolor{red}{0.8863} &   \textcolor{red}{0.7859} &   \textcolor{red}{0.7817} &                    0.7482 \\
   \rowcolor{gray!15}[\dimexpr\tabcolsep + 1pt\relax]         DA &                    0.8845 &                    0.8836 &   \textcolor{red}{0.8628} &                    0.7676 &   \textcolor{red}{0.7673} &   \textcolor{red}{0.7410} &                    0.7221 &  \textcolor{blue}{0.7215} &                    0.6957 &                    0.9040 &                    0.9049 &                    0.8861 &                    0.7846 &  \textcolor{blue}{0.7816} &  \textcolor{blue}{0.7484} \\
   \rowcolor{gray!15}[\dimexpr\tabcolsep + 1pt\relax]   DA (all) &  \textcolor{blue}{0.8847} &   \textcolor{red}{0.8837} &                    0.8622 &                    0.7666 &                    0.7667 &                    0.7401 &                    0.7217 &                    0.7211 &  \textcolor{blue}{0.6957} &                    0.9033 &                    0.9050 &                    0.8856 &                    0.7822 &                    0.7807 &                    0.7480 \\
    \rowcolor{gray!15}[\dimexpr\tabcolsep + 1pt\relax]     DGFMB &                    0.8845 &                    0.8836 &                    0.8623 &                    0.7672 &                    0.7669 &                    0.7406 &                    0.7224 &   \textcolor{red}{0.7216} &                    0.6956 &  \textcolor{blue}{0.9041} &                    0.9048 &                    0.8860 &                    0.7850 &                    0.7812 &   \textcolor{red}{0.7485} \\
 \rowcolor{gray!15}[\dimexpr\tabcolsep + 1pt\relax] DGFMB (no FC) &   \textcolor{red}{0.8849} &                    0.8834 &                    0.8625 &                    0.7673 &                    0.7669 &                    0.7406 &                    0.7224 &                    0.7211 &                    0.6954 &                    0.9040 &                    0.9046 &                    0.8856 &                    0.7851 &                    0.7810 &                    0.7479 \\
\bottomrule
            \end{tabularx}
		\end{center}
\label{non_blind_ssim}
\end{table}

\begin{table}[H]
	\caption{SSIM SR results corresponding to Table 3 from the main paper. 'low' refers to a $\sigma$ of 0.2, `med' refers to a $\sigma$ of 1.6 and `high' refers to a $\sigma$ of 3.0. The best result for each set is shown in \textcolor{red}{red}, while the second-best result is shown in \textcolor{blue}{blue}.}
  \setlength{\tabcolsep}{2.8pt}

	\renewcommand{\arraystretch}{1.3}
	    \begin{center}
            \begin{tabularx}{\linewidth}{P{3.6cm}ccccccccccccccc}
\toprule
                  \textbf{Model} & \multicolumn{3}{c}{\textbf{Set5}} & \multicolumn{3}{c}{\textbf{Set14}} & \multicolumn{3}{c}{\textbf{BSDS100}} & \multicolumn{3}{c}{\textbf{Manga109}} & \multicolumn{3}{c}{\textbf{Urban100}} \\
                       &                       \textbf{low} &                       \textbf{med} &                      \textbf{high} &                       \textbf{low} &                       \textbf{med} &                      \textbf{high} &                       \textbf{low} &                       \textbf{med} &                      \textbf{high} &                       \textbf{low} &                       \textbf{med} &                      \textbf{high} &                       \textbf{low} &                       \textbf{med} &                      \textbf{high} \\
\midrule
               \rowcolor{gray!5}[\dimexpr\tabcolsep + 1pt\relax] \textbf{Classical} &&&&&&&&&&&&&&& \\
 \rowcolor{gray!5}[\dimexpr\tabcolsep + 1pt\relax]           Bicubic &                    0.7909 &                    0.7513 &                    0.6736 &                    0.6803 &                    0.6384 &                    0.5700 &                    0.6452 &                    0.6058 &                    0.5458 &                    0.7716 &                    0.7323 &                    0.6659 &                    0.6376 &                    0.5941 &                    0.5240 \\
         \rowcolor{gray!5}[\dimexpr\tabcolsep + 1pt\relax]              Lanczos &                    0.7986 &                    0.7625 &                    0.6814 &                    0.6888 &                    0.6487 &                    0.5761 &                    0.6534 &                    0.6149 &                    0.5510 &                    0.7775 &                    0.7418 &                    0.6720 &                    0.6452 &                    0.6044 &                    0.5302 \\
    \rowcolor{gray!15}[\dimexpr\tabcolsep + 1pt\relax] \textbf{Pretrained} &&&&&&&&&&&&&&& \\     
 \rowcolor{gray!15}[\dimexpr\tabcolsep + 1pt\relax] IKC (pretrained-best-iter) &  \textcolor{blue}{0.8832} &                    0.8751 &  \textcolor{blue}{0.8604} &                    0.7658 &                    0.7486 &   \textcolor{red}{0.7382} &                    0.7214 &                    0.7083 &   \textcolor{red}{0.6912} &  \textcolor{blue}{0.9019} &                    0.8763 &                    0.8612 &                    0.7769 &                    0.7488 &                    0.7262 \\
\rowcolor{gray!15}[\dimexpr\tabcolsep + 1pt\relax]  IKC (pretrained-last-iter) &                    0.8827 &                    0.8663 &                    0.8585 &  \textcolor{blue}{0.7675} &                    0.7436 &                    0.7269 &   \textcolor{red}{0.7233} &                    0.6967 &                    0.6872 &                    0.9015 &                    0.8694 &                    0.8454 &                    0.7754 &                    0.7389 &                    0.7149 \\
  \rowcolor{gray!15}[\dimexpr\tabcolsep + 1pt\relax]         DASR (pretrained) &                    0.8797 &                    0.8747 &                    0.8539 &                    0.7617 &                    0.7512 &                    0.7205 &                    0.7177 &                    0.7087 &                    0.6796 &                    0.8960 &                    0.8909 &                    0.8690 &                    0.7665 &                    0.7521 &                    0.7182 \\
  \rowcolor{gray!15}[\dimexpr\tabcolsep + 1pt\relax]        DANv1 (pretrained) &                    0.8828 &  \textcolor{blue}{0.8787} &   \textcolor{red}{0.8610} &                    0.7657 &  \textcolor{blue}{0.7594} &  \textcolor{blue}{0.7331} &                    0.7211 &  \textcolor{blue}{0.7144} &  \textcolor{blue}{0.6877} &                    0.9011 &  \textcolor{blue}{0.9007} &   \textcolor{red}{0.8839} &  \textcolor{blue}{0.7793} &  \textcolor{blue}{0.7706} &   \textcolor{red}{0.7401} \\
  \rowcolor{gray!15}[\dimexpr\tabcolsep + 1pt\relax]        DANv2 (pretrained) &   \textcolor{red}{0.8849} &   \textcolor{red}{0.8812} &                    0.8589 &   \textcolor{red}{0.7678} &   \textcolor{red}{0.7617} &                    0.7329 &  \textcolor{blue}{0.7233} &   \textcolor{red}{0.7171} &                    0.6868 &   \textcolor{red}{0.9028} &   \textcolor{red}{0.9009} &  \textcolor{blue}{0.8827} &   \textcolor{red}{0.7850} &   \textcolor{red}{0.7736} &  \textcolor{blue}{0.7395} \\

                 \rowcolor{gray!5}[\dimexpr\tabcolsep + 1pt\relax]           \textbf{Non-Blind} &&&&&&&&&&&&&&& \\
 \rowcolor{gray!5}[\dimexpr\tabcolsep + 1pt\relax]                  RCAN-MA (true sigma) &   \textcolor{red}{0.8845} &   \textcolor{red}{0.8837} &   \textcolor{red}{0.8615} &   \textcolor{red}{0.7671} &  \textcolor{blue}{0.7669} &   \textcolor{red}{0.7408} &                    0.7221 &  \textcolor{blue}{0.7212} &  \textcolor{blue}{0.6953} &  \textcolor{blue}{0.9041} &  \textcolor{blue}{0.9047} &  \textcolor{blue}{0.8857} &  \textcolor{blue}{0.7849} &  \textcolor{blue}{0.7812} &  \textcolor{blue}{0.7479} \\
   \rowcolor{gray!5}[\dimexpr\tabcolsep + 1pt\relax]   HAN-MA (true sigma) &  \textcolor{blue}{0.8842} &  \textcolor{blue}{0.8831} &  \textcolor{blue}{0.8611} &  \textcolor{blue}{0.7671} &   \textcolor{red}{0.7670} &  \textcolor{blue}{0.7407} &  \textcolor{blue}{0.7222} &   \textcolor{red}{0.7213} &   \textcolor{red}{0.6956} &   \textcolor{red}{0.9043} &   \textcolor{red}{0.9050} &   \textcolor{red}{0.8862} &   \textcolor{red}{0.7850} &   \textcolor{red}{0.7814} &   \textcolor{red}{0.7483} \\
  \rowcolor{gray!5}[\dimexpr\tabcolsep + 1pt\relax]  RCAN-MA (noisy sigma) &                    0.8839 &                    0.8808 &                    0.8589 &                    0.7669 &                    0.7639 &                    0.7328 &   \textcolor{red}{0.7223} &                    0.7177 &                    0.6902 &                    0.9025 &                    0.9010 &                    0.8754 &                    0.7825 &                    0.7741 &                    0.7343 \\
\rowcolor{gray!15}[\dimexpr\tabcolsep + 1pt\relax]           \textbf{RCAN/DAN} &&&&&&&&&&&&&&& \\
 \rowcolor{gray!15}[\dimexpr\tabcolsep + 1pt\relax]               DANv1 &                    0.8791 &                    0.8749 &                    0.8511 &                    0.7608 &                    0.7569 &                    0.7261 &                    0.7170 &                    0.7120 &                    0.6840 &                    0.8921 &                    0.8911 &                    0.8652 &                    0.7620 &                    0.7550 &                    0.7187 \\
  \rowcolor{gray!15}[\dimexpr\tabcolsep + 1pt\relax]     DANv1 (cosine) &                    0.8820 &                    0.8774 &                    0.8540 &                    0.7638 &                    0.7604 &                    0.7305 &                    0.7195 &                    0.7149 &                    0.6870 &                    0.8980 &                    0.8961 &                    0.8720 &                    0.7721 &                    0.7642 &                    0.7269 \\
 \rowcolor{gray!15}[\dimexpr\tabcolsep + 1pt\relax] RCAN (batch size 4) &                    0.8830 &                    0.8792 &                    0.8564 &                    0.7662 &                    0.7636 &   \textcolor{red}{0.7329} &                    0.7221 &                    0.7174 &                    0.6900 &                    0.9019 &                    0.9007 &                    0.8760 &  \textcolor{blue}{0.7816} &                    0.7743 &                    0.7351 \\
 \rowcolor{gray!15}[\dimexpr\tabcolsep + 1pt\relax]            RCAN-DAN &   \textcolor{red}{0.8840} &  \textcolor{blue}{0.8796} &  \textcolor{blue}{0.8583} &  \textcolor{blue}{0.7669} &  \textcolor{blue}{0.7644} &                    0.7323 &  \textcolor{blue}{0.7223} &  \textcolor{blue}{0.7183} &  \textcolor{blue}{0.6902} &  \textcolor{blue}{0.9027} &   \textcolor{red}{0.9019} &   \textcolor{red}{0.8776} &   \textcolor{red}{0.7828} &  \textcolor{blue}{0.7754} &   \textcolor{red}{0.7360} \\
 \rowcolor{gray!15}[\dimexpr\tabcolsep + 1pt\relax]    RCAN-DAN (sigma) &  \textcolor{blue}{0.8838} &   \textcolor{red}{0.8807} &   \textcolor{red}{0.8591} &   \textcolor{red}{0.7676} &   \textcolor{red}{0.7654} &  \textcolor{blue}{0.7328} &   \textcolor{red}{0.7229} &   \textcolor{red}{0.7193} &   \textcolor{red}{0.6908} &   \textcolor{red}{0.9029} &  \textcolor{blue}{0.9019} &  \textcolor{blue}{0.8775} &                    0.7811 &   \textcolor{red}{0.7757} &  \textcolor{blue}{0.7356} \\
    \rowcolor{gray!5}[\dimexpr\tabcolsep + 1pt\relax]           \textbf{HAN/DAN} &&&&&&&&&&&&&&& \\
     \rowcolor{gray!5}[\dimexpr\tabcolsep + 1pt\relax]               HAN (batch size 4) &  \textcolor{blue}{0.8837} &  \textcolor{blue}{0.8795} &  \textcolor{blue}{0.8550} &  \textcolor{blue}{0.7668} &  \textcolor{blue}{0.7636} &  \textcolor{blue}{0.7316} &  \textcolor{blue}{0.7221} &   \textcolor{red}{0.7175} &   \textcolor{red}{0.6895} &  \textcolor{blue}{0.9017} &  \textcolor{blue}{0.9003} &  \textcolor{blue}{0.8739} &  \textcolor{blue}{0.7822} &  \textcolor{blue}{0.7743} &  \textcolor{blue}{0.7321} \\
     \rowcolor{gray!5}[\dimexpr\tabcolsep + 1pt\relax]        HAN-DAN &   \textcolor{red}{0.8842} &   \textcolor{red}{0.8797} &   \textcolor{red}{0.8579} &   \textcolor{red}{0.7671} &   \textcolor{red}{0.7638} &   \textcolor{red}{0.7331} &   \textcolor{red}{0.7225} &  \textcolor{blue}{0.7172} &  \textcolor{blue}{0.6894} &   \textcolor{red}{0.9027} &   \textcolor{red}{0.9010} &   \textcolor{red}{0.8773} &   \textcolor{red}{0.7829} &   \textcolor{red}{0.7751} &   \textcolor{red}{0.7354} \\
                             \rowcolor{gray!15}[\dimexpr\tabcolsep + 1pt\relax]           \textbf{RCAN/Contrastive} &&&&&&&&&&&&&&& \\
    \rowcolor{gray!15}[\dimexpr\tabcolsep + 1pt\relax]  RCAN (batch size 8) &                    0.8830 &                    0.8783 &                    0.8564 &                    0.7664 &                    0.7632 &                    0.7314 &                    0.7217 &                    0.7170 &                    0.6897 &                    0.9015 &                    0.9000 &                    0.8728 &   \textcolor{red}{0.7834} &   \textcolor{red}{0.7760} &   \textcolor{red}{0.7357} \\
   \rowcolor{gray!15}[\dimexpr\tabcolsep + 1pt\relax]               RCAN-MoCo &  \textcolor{blue}{0.8844} &                    0.8801 &                    0.8577 &                    0.7671 &                    0.7629 &  \textcolor{blue}{0.7332} &                    0.7221 &                    0.7174 &                    0.6904 &                    0.9022 &                    0.9008 &                    0.8770 &                    0.7819 &                    0.7742 &                    0.7343 \\
    \rowcolor{gray!15}[\dimexpr\tabcolsep + 1pt\relax]           RCAN-SupMoCo &                    0.8839 &  \textcolor{blue}{0.8804} &                    0.8578 &                    0.7671 &                    0.7630 &                    0.7329 &                    0.7227 &                    0.7177 &                    0.6901 &                    0.9030 &  \textcolor{blue}{0.9010} &  \textcolor{blue}{0.8771} &                    0.7831 &                    0.7748 &  \textcolor{blue}{0.7356} \\
  \rowcolor{gray!15}[\dimexpr\tabcolsep + 1pt\relax]          RCAN-regression &                    0.8835 &                    0.8795 &                    0.8559 &                    0.7668 &                    0.7640 &                    0.7331 &  \textcolor{blue}{0.7229} &                    0.7180 &                    0.6899 &                    0.9024 &                    0.9005 &   \textcolor{red}{0.8772} &                    0.7825 &                    0.7736 &                    0.7337 \\
 \rowcolor{gray!15}[\dimexpr\tabcolsep + 1pt\relax]   RCAN-SupMoCo-regression &   \textcolor{red}{0.8845} &   \textcolor{red}{0.8805} &                    0.8567 &                    0.7667 &   \textcolor{red}{0.7645} &                    0.7321 &                    0.7223 &                    0.7177 &  \textcolor{blue}{0.6905} &   \textcolor{red}{0.9031} &   \textcolor{red}{0.9016} &                    0.8770 &  \textcolor{blue}{0.7831} &  \textcolor{blue}{0.7749} &                    0.7348 \\
    \rowcolor{gray!15}[\dimexpr\tabcolsep + 1pt\relax]           RCAN-WeakCon &                    0.8837 &                    0.8794 &  \textcolor{blue}{0.8582} &  \textcolor{blue}{0.7673} &  \textcolor{blue}{0.7641} &   \textcolor{red}{0.7334} &   \textcolor{red}{0.7229} &   \textcolor{red}{0.7189} &   \textcolor{red}{0.6905} &                    0.9027 &                    0.9009 &                    0.8760 &                    0.7826 &                    0.7747 &                    0.7350 \\
  \rowcolor{gray!15}[\dimexpr\tabcolsep + 1pt\relax]    RCAN-SupMoCo (online) &                    0.8836 &                    0.8799 &   \textcolor{red}{0.8591} &   \textcolor{red}{0.7673} &                    0.7634 &                    0.7245 &                    0.7227 &  \textcolor{blue}{0.7186} &                    0.6903 &  \textcolor{blue}{0.9030} &                    0.9005 &                    0.8763 &                    0.7826 &                    0.7743 &                    0.7317 \\
 \rowcolor{gray!15}[\dimexpr\tabcolsep + 1pt\relax]     RCAN-SupMoCo (ResNet) &                    0.8829 &                    0.8789 &                    0.8570 &                    0.7666 &                    0.7630 &                    0.7292 &                    0.7221 &                    0.7170 &                    0.6882 &                    0.9017 &                    0.8998 &                    0.8629 &                    0.7798 &                    0.7714 &                    0.7250 \\
  \rowcolor{gray!5}[\dimexpr\tabcolsep + 1pt\relax]           \textbf{HAN/Contrastive} &&&&&&&&&&&&&&& \\
   \rowcolor{gray!5}[\dimexpr\tabcolsep + 1pt\relax]         HAN (batch size 8) &   \textcolor{red}{0.8836} &   \textcolor{red}{0.8802} &  \textcolor{blue}{0.8554} &   \textcolor{red}{0.7669} &   \textcolor{red}{0.7638} &  \textcolor{blue}{0.7311} &   \textcolor{red}{0.7223} &   \textcolor{red}{0.7180} &  \textcolor{blue}{0.6896} &   \textcolor{red}{0.9025} &   \textcolor{red}{0.9013} &  \textcolor{blue}{0.8747} &   \textcolor{red}{0.7826} &   \textcolor{red}{0.7754} &  \textcolor{blue}{0.7337} \\
 \rowcolor{gray!5}[\dimexpr\tabcolsep + 1pt\relax] HAN-SupMoCo-regression &  \textcolor{blue}{0.8835} &  \textcolor{blue}{0.8801} &   \textcolor{red}{0.8581} &  \textcolor{blue}{0.7665} &  \textcolor{blue}{0.7634} &   \textcolor{red}{0.7332} &  \textcolor{blue}{0.7220} &  \textcolor{blue}{0.7167} &   \textcolor{red}{0.6899} &  \textcolor{blue}{0.9019} &  \textcolor{blue}{0.8999} &   \textcolor{red}{0.8774} &  \textcolor{blue}{0.7815} &  \textcolor{blue}{0.7735} &   \textcolor{red}{0.7348} \\
\rowcolor{gray!15}[\dimexpr\tabcolsep + 1pt\relax] \textbf{Extensions} &&&&&&&&&&&&&&& \\
 \rowcolor{gray!15}[\dimexpr\tabcolsep + 1pt\relax]     RCAN-DAN (pretrained estimator) &  \textcolor{blue}{0.8839} &                    0.8794 &  \textcolor{blue}{0.8576} &   \textcolor{red}{0.7678} &  \textcolor{blue}{0.7647} &  \textcolor{blue}{0.7332} &  \textcolor{blue}{0.7227} &  \textcolor{blue}{0.7185} &  \textcolor{blue}{0.6908} &  \textcolor{blue}{0.9034} &   \textcolor{red}{0.9024} &  \textcolor{blue}{0.8769} &  \textcolor{blue}{0.7837} &  \textcolor{blue}{0.7762} &  \textcolor{blue}{0.7363} \\
 \rowcolor{gray!15}[\dimexpr\tabcolsep + 1pt\relax]      RCAN (batch size 8, long-term) &                    0.8836 &   \textcolor{red}{0.8806} &   \textcolor{red}{0.8580} &                    0.7670 &                    0.7642 &                    0.7325 &                    0.7222 &                    0.7178 &                    0.6902 &                    0.9027 &                    0.9014 &                    0.8757 &                    0.7834 &                    0.7760 &                    0.7357 \\
\rowcolor{gray!15}[\dimexpr\tabcolsep + 1pt\relax]  RCAN-SupMoCo-regression (long-term) &   \textcolor{red}{0.8842} &  \textcolor{blue}{0.8797} &                    0.8570 &  \textcolor{blue}{0.7675} &   \textcolor{red}{0.7649} &   \textcolor{red}{0.7333} &   \textcolor{red}{0.7231} &   \textcolor{red}{0.7188} &   \textcolor{red}{0.6913} &   \textcolor{red}{0.9037} &  \textcolor{blue}{0.9021} &   \textcolor{red}{0.8786} &   \textcolor{red}{0.7852} &   \textcolor{red}{0.7771} &   \textcolor{red}{0.7377} \\
\bottomrule
            \end{tabularx}
		\end{center}
\label{simple_blind_ssim}
\end{table}

\begin{table}[H]
	\caption{LPIPS results (smaller is better) for Real-ESRGAN models tested.  The datasets are identical to those used for the simple pipeline analysis. The best result for each set is shown in \textcolor{red}{red}, while the second-best result is shown in \textcolor{blue}{blue}.}
  \setlength{\tabcolsep}{2.8pt}

	\renewcommand{\arraystretch}{1.3}
	    \begin{center}
            \begin{tabularx}{\linewidth}{P{3.6cm}ccccccccccccccc}
\toprule
                            \textbf{Model} & \multicolumn{3}{c}{\textbf{Set5}} & \multicolumn{3}{c}{\textbf{Set14}} & \multicolumn{3}{c}{\textbf{BSDS100}} & \multicolumn{3}{c}{\textbf{Manga109}} & \multicolumn{3}{c}{\textbf{Urban100}} \\
                       &                       \textbf{low} &                       \textbf{med} &                      \textbf{high} &                       \textbf{low} &                       \textbf{med} &                      \textbf{high} &                       \textbf{low} &                       \textbf{med} &                      \textbf{high} &                       \textbf{low} &                       \textbf{med} &                      \textbf{high} &                       \textbf{low} &                       \textbf{med} &                      \textbf{high} \\
\midrule
                           Bicubic &                    0.1131 &                    0.1582 &                    0.2543 &                    0.1672 &                    0.2134 &                    0.3052 &                    0.2040 &                    0.2465 &                    0.3300 &                    0.1398 &                    0.1905 &                    0.2900 &                    0.2117 &                    0.2594 &                    0.3501 \\
                          Lanczos &                    0.1051 &                    0.1458 &                    0.2442 &                    0.1607 &                    0.2024 &                    0.2956 &                    0.1992 &                    0.2370 &                    0.3216 &                    0.1307 &                    0.1781 &                    0.2798 &                    0.2057 &                    0.2495 &                    0.3409 \\
       Real-ESRGAN (batch size 4) &                    0.0211 &  \textcolor{blue}{0.0266} &                    0.0468 &                    0.0433 &                    0.0490 &                    0.0738 &                    0.0555 &                    0.0611 &                    0.0878 &                    0.0210 &                    0.0225 &                    0.0340 &                    0.0492 &                    0.0535 &                    0.0789 \\
       Real-ESRGAN (batch size 8) &                    0.0250 &                    0.0299 &                    0.0440 &                    0.0447 &                    0.0502 &                    0.0792 &                    0.0553 &                    0.0624 &                    0.0872 &                    0.0208 &                    0.0231 &                    0.0368 &                    0.0525 &                    0.0585 &                    0.0831 \\
       Real-ESRGAN (MA-non-blind) &  \textcolor{blue}{0.0200} &   \textcolor{red}{0.0203} &   \textcolor{red}{0.0271} &   \textcolor{red}{0.0418} &   \textcolor{red}{0.0421} &   \textcolor{red}{0.0535} &  \textcolor{blue}{0.0518} &   \textcolor{red}{0.0540} &   \textcolor{red}{0.0650} &   \textcolor{red}{0.0181} &   \textcolor{red}{0.0177} &   \textcolor{red}{0.0241} &  \textcolor{blue}{0.0463} &   \textcolor{red}{0.0480} &   \textcolor{red}{0.0593} \\
 Real-ESRGAN (SupMoCo-regression) &                    0.0227 &                    0.0302 &                    0.0416 &                    0.0478 &                    0.0532 &                    0.0899 &                    0.0607 &                    0.0661 &  \textcolor{blue}{0.0815} &                    0.0213 &                    0.0241 &                    0.0327 &                    0.0529 &                    0.0543 &                    0.0767 \\
                  Real-ESRGAN-DAN &   \textcolor{red}{0.0185} &                    0.0275 &  \textcolor{blue}{0.0392} &  \textcolor{blue}{0.0421} &  \textcolor{blue}{0.0465} &  \textcolor{blue}{0.0663} &   \textcolor{red}{0.0491} &  \textcolor{blue}{0.0578} &                    0.0833 &  \textcolor{blue}{0.0196} &  \textcolor{blue}{0.0217} &  \textcolor{blue}{0.0289} &   \textcolor{red}{0.0458} &  \textcolor{blue}{0.0509} &  \textcolor{blue}{0.0729} \\
\bottomrule
            \end{tabularx}
		\end{center}
\label{esrgan_lpips}
\end{table}

\begin{table}[H]
	\caption{PSNR (dB) results for ELAN models tested.  The datasets are identical to those used for the simple pipeline analysis.  ELAN (MA-all-non-blind) refers to an ELAN model upgraded with a single meta-attention block, with its attention vector multiplied with the network feature map at 36 points through the network (one for each attention block). The best result for each set is shown in \textcolor{red}{red}, while the second-best result is shown in \textcolor{blue}{blue}.}
  \setlength{\tabcolsep}{3pt}

	\renewcommand{\arraystretch}{1.3}
	    \begin{center}
            \begin{tabularx}{\linewidth}{cccccccccccccccc}
\toprule
                   \textbf{Model} & \multicolumn{3}{c}{\textbf{Set5}} & \multicolumn{3}{c}{\textbf{Set14}} & \multicolumn{3}{c}{\textbf{BSDS100}} & \multicolumn{3}{c}{\textbf{Manga109}} & \multicolumn{3}{c}{\textbf{Urban100}} \\
                       &                       \textbf{low} &                       \textbf{med} &                      \textbf{high} &                       \textbf{low} &                       \textbf{med} &                      \textbf{high} &                       \textbf{low} &                       \textbf{med} &                      \textbf{high} &                       \textbf{low} &                       \textbf{med} &                      \textbf{high} &                       \textbf{low} &                       \textbf{med} &                      \textbf{high} \\
\midrule
                 Bicubic &                    27.084 &                    25.857 &                    23.867 &                    24.532 &                    23.695 &                    22.286 &                    24.647 &                    23.998 &                    22.910 &                    23.608 &                    22.564 &                    20.932 &                    21.805 &                    21.104 &                    19.944 \\
                 Lanczos &  \textcolor{blue}{27.462} &                    26.210 &                    24.039 &                    24.760 &                    23.925 &                    22.409 &                    24.811 &                    24.173 &                    23.007 &                    23.923 &                    22.850 &                    21.071 &                    21.989 &                    21.293 &                    20.046 \\
                    ELAN &                    27.341 &  \textcolor{blue}{27.400} &  \textcolor{blue}{24.759} &                    24.813 &  \textcolor{blue}{24.715} &  \textcolor{blue}{22.914} &                    24.815 &  \textcolor{blue}{24.789} &  \textcolor{blue}{23.403} &                    24.233 &  \textcolor{blue}{23.894} &  \textcolor{blue}{21.648} &                    22.051 &  \textcolor{blue}{21.982} &  \textcolor{blue}{20.477} \\
     ELAN (MA-non-blind) &                    27.401 &                    27.391 &                    24.731 &  \textcolor{blue}{24.841} &                    24.714 &                    22.899 &  \textcolor{blue}{24.841} &                    24.788 &                    23.396 &  \textcolor{blue}{24.255} &                    23.883 &                    21.629 &  \textcolor{blue}{22.077} &                    21.977 &                    20.465 \\
 ELAN (MA-all-non-blind) &   \textcolor{red}{30.431} &   \textcolor{red}{30.405} &   \textcolor{red}{29.007} &   \textcolor{red}{26.733} &   \textcolor{red}{26.728} &   \textcolor{red}{25.922} &   \textcolor{red}{26.156} &   \textcolor{red}{26.177} &   \textcolor{red}{25.541} &   \textcolor{red}{28.539} &   \textcolor{red}{28.764} &   \textcolor{red}{27.545} &   \textcolor{red}{24.590} &   \textcolor{red}{24.546} &   \textcolor{red}{23.849} \\
\bottomrule
            \end{tabularx}
		\end{center}
\label{elan_psnr}
\end{table}

\begin{table}[H]
	\caption{SSIM results for ELAN models tested.  The datasets are identical to those used for the simple pipeline analysis. ELAN (MA-all-non-blind) refers to an ELAN model upgraded with a single meta-attention block, with its attention vector multiplied with the network feature map at 36 points through the network (one for each attention block). The best result for each set is shown in \textcolor{red}{red}, while the second-best result is shown in \textcolor{blue}{blue}.}
  \setlength{\tabcolsep}{3pt}

	\renewcommand{\arraystretch}{1.3}
	    \begin{center}
            \begin{tabularx}{\linewidth}{cccccccccccccccc}
\toprule
                   \textbf{Model} & \multicolumn{3}{c}{\textbf{Set5}} & \multicolumn{3}{c}{\textbf{Set14}} & \multicolumn{3}{c}{\textbf{BSDS100}} & \multicolumn{3}{c}{\textbf{Manga109}} & \multicolumn{3}{c}{\textbf{Urban100}} \\
                       &                       \textbf{low} &                       \textbf{med} &                      \textbf{high} &                       \textbf{low} &                       \textbf{med} &                      \textbf{high} &                       \textbf{low} &                       \textbf{med} &                      \textbf{high} &                       \textbf{low} &                       \textbf{med} &                      \textbf{high} &                       \textbf{low} &                       \textbf{med} &                      \textbf{high} \\
\midrule
                 Bicubic &                    0.7909 &                    0.7513 &                    0.6736 &                    0.6803 &                    0.6384 &                    0.5700 &                    0.6452 &                    0.6058 &                    0.5458 &                    0.7716 &                    0.7323 &                    0.6659 &                    0.6376 &                    0.5941 &                    0.5240 \\
                 Lanczos &  \textcolor{blue}{0.7986} &                    0.7625 &                    0.6814 &                    0.6888 &                    0.6487 &                    0.5761 &                    0.6534 &                    0.6149 &                    0.5510 &                    0.7775 &                    0.7418 &                    0.6720 &                    0.6452 &                    0.6044 &                    0.5302 \\
                    ELAN &                    0.7962 &  \textcolor{blue}{0.7933} &  \textcolor{blue}{0.7078} &                    0.7074 &                    0.6856 &  \textcolor{blue}{0.5986} &                    0.6772 &  \textcolor{blue}{0.6512} &  \textcolor{blue}{0.5709} &                    0.7805 &                    0.7697 &  \textcolor{blue}{0.6914} &                    0.6597 &                    0.6405 &                    0.5528 \\
     ELAN (MA-non-blind) &                    0.7970 &                    0.7932 &                    0.7071 &  \textcolor{blue}{0.7079} &  \textcolor{blue}{0.6857} &                    0.5981 &  \textcolor{blue}{0.6775} &                    0.6508 &                    0.5702 &  \textcolor{blue}{0.7814} &  \textcolor{blue}{0.7703} &                    0.6913 &  \textcolor{blue}{0.6610} &  \textcolor{blue}{0.6411} &  \textcolor{blue}{0.5530} \\
 ELAN (MA-all-non-blind) &   \textcolor{red}{0.8778} &   \textcolor{red}{0.8773} &   \textcolor{red}{0.8529} &   \textcolor{red}{0.7591} &   \textcolor{red}{0.7585} &   \textcolor{red}{0.7311} &   \textcolor{red}{0.7142} &   \textcolor{red}{0.7133} &   \textcolor{red}{0.6879} &   \textcolor{red}{0.8887} &   \textcolor{red}{0.8895} &   \textcolor{red}{0.8664} &   \textcolor{red}{0.7669} &   \textcolor{red}{0.7637} &   \textcolor{red}{0.7336} \\
\bottomrule
            \end{tabularx}
		\end{center}
\label{elan_ssim}
\end{table}
\newpage
\section*{\normalfont \Large \textbf{Complex Pipeline Results}}

\begin{figure}[H]
        \centering
        \includegraphics[width=\linewidth]{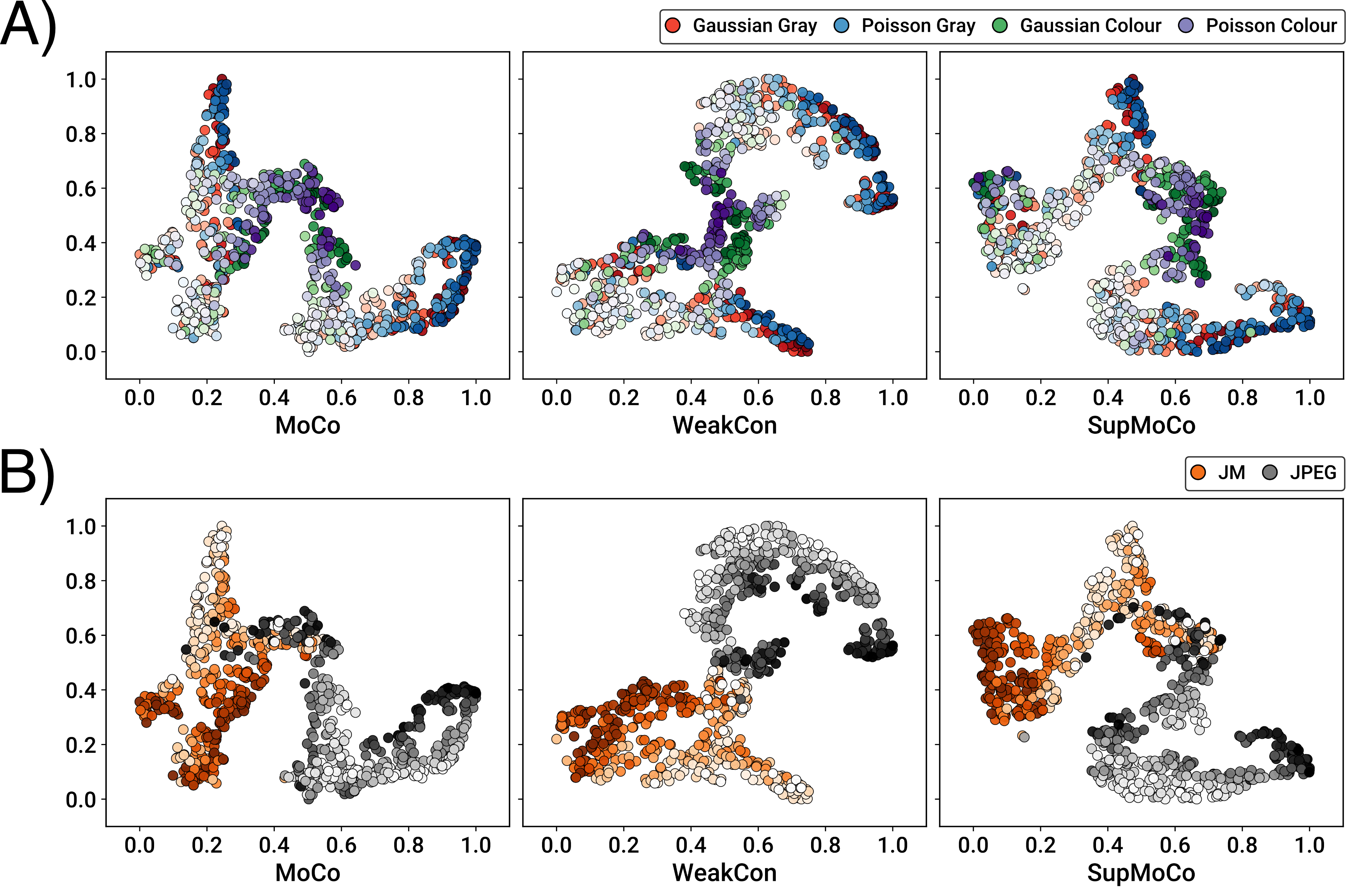}

    \caption{t-SNE plots (perplexity value of 40) showing the separation power of the different contrastive learning algorithms considered on the complex pipeline.  All models were evaluated at epoch 2000.  The test dataset consisted of 2472 images (8 sets of 309 images of BSDS100/Manga109/Urban100) degraded using the full complex pipeline (random degradation parameters).  Only 800 images (randomly selected) are plotted per panel, to reduce cluttering.  Each dimension was independently normalised in
the range $[0, 1]$ after computing the t-SNE results. A)  t-SNE plots with each image coloured according to the noise applied.  The colour intensity of each point corresponds to the noise magnitude (in the range $[1,30]$ for Gaussian noise and $[0.05,3]$ for Poisson noise).  All three encoders are capable of separating noise magnitudes, but lose the ability to distinctly separate the two noise classes.  B) t-SNE plots with each image coloured according to the compression applied.  The colour intensity of each point corresponds to the compression level, which is in the range $[30,95]$ (higher is better) for JPEG and $[20,40]$ (lower is better) for JM H.264. All three encoders are capable of separating compression magnitudes, but struggle to separate the two compression types when the compression is very low.}
    \label{fig:complex_tsne_supp}
\end{figure}

\begin{table}[H]
	\caption{SSIM SR results corresponding to Table 6 from the main paper.  Please refer to Table 4 in the main paper for full details on each testing scenario.  The best result for each set is shown in \textcolor{red}{red}, while the second-best result is shown in \textcolor{blue}{blue}.}
  \setlength{\tabcolsep}{0.5pt}
	\renewcommand{\arraystretch}{1.2}
	    \begin{center}
            \begin{tabularx}{\linewidth}{cP{3.2cm}P{1.2cm}P{1.2cm}P{1.2cm}P{1.2cm}P{1.2cm}P{1.2cm}P{1.2cm}P{1.2cm}P{1.2cm}P{1.2cm}P{1.4cm}}
\toprule
              \textbf{Dataset}     &         \textbf{Model} &                      \textbf{JPEG} &                        \textbf{JM} &                   \textbf{Poisson} &                  \textbf{Gaussian} &                       \textbf{Iso} &                     \textbf{Aniso} &            \textbf{Iso + Gaussian }&           \textbf{Gaussian + JPEG} &     \textbf{Iso + Gaussian + JPEG} &      \textbf{Aniso + Poisson + JM} & \textbf{Iso/Aniso + Gaussian/Poisson + JPEG/JM} \\
\midrule
      \rowcolor{gray!5}[\dimexpr\tabcolsep + 1pt\relax] &    Bicubic &                    0.5777 &                    0.5870 &                    0.4443 &                    0.4341 &                    0.5880 &                    0.6075 &                    0.3847 &                    0.4284 &                    0.3932 &                    0.4229 &                                 0.4067 \\
              \rowcolor{gray!5}[\dimexpr\tabcolsep + 1pt\relax] &             Lanczos &                    0.5792 &                    0.5911 &                    0.4283 &                    0.4173 &                    0.5964 &                    0.6165 &                    0.3674 &                    0.4133 &                    0.3787 &                    0.4090 &                                 0.3922 \\
          \rowcolor{gray!5}[\dimexpr\tabcolsep + 1pt\relax] &     RCAN (batch size 4) &                    0.6117 &                    0.6213 &                    0.6004 &                    0.5995 &  \textcolor{blue}{0.6660} &                    0.6744 &                    0.5625 &                    0.5749 &                    0.5465 &                    0.5543 &                                 0.5503 \\
          \rowcolor{gray!5}[\dimexpr\tabcolsep + 1pt\relax] &     RCAN (batch size 8) &                    0.6117 &                    0.6211 &                    0.6014 &                    0.5991 &                    0.6649 &                    0.6723 &                    0.5617 &                    0.5751 &                    0.5467 &                    0.5548 &                                 0.5505 \\
            \rowcolor{gray!5}[\dimexpr\tabcolsep + 1pt\relax] &      RCAN (non-blind) &                       N/A &                       N/A &                       N/A &                       N/A &                       N/A &                       N/A &                       N/A &                       N/A &                    0.5482 &                    0.5494 &                                 0.5496 \\
    \rowcolor{gray!5}[\dimexpr\tabcolsep + 1pt\relax] &     RCAN (non-blind, no blur) &                       N/A &                       N/A &                       N/A &                       N/A &                       N/A &                       N/A &                       N/A &   \textcolor{red}{0.5788} &                    0.5480 &                    0.5512 &                                 0.5503 \\
\rowcolor{gray!5}[\dimexpr\tabcolsep + 1pt\relax] & RCAN (non-blind, no noise) &                       N/A &                       N/A &                       N/A &                       N/A &                       N/A &                       N/A &                       N/A &                       N/A &  \textcolor{blue}{0.5483} &                    0.5542 &                \textcolor{red}{0.5519} \\
 \rowcolor{gray!5}[\dimexpr\tabcolsep + 1pt\relax] & RCAN (non-blind, no compression) &                       N/A &                       N/A &                       N/A &                       N/A &                       N/A &                       N/A &  \textcolor{blue}{0.5629} &                       N/A &   \textcolor{red}{0.5491} &                    0.5526 &                                 0.5499 \\
                 \rowcolor{gray!5}[\dimexpr\tabcolsep + 1pt\relax] &      RCAN (MoCo) &                    0.6108 &                    0.6217 &  \textcolor{blue}{0.6017} &  \textcolor{blue}{0.6002} &                    0.6590 &                    0.6699 &                    0.5619 &                    0.5750 &                    0.5469 &   \textcolor{red}{0.5561} &               \textcolor{blue}{0.5511} \\
              \rowcolor{gray!5}[\dimexpr\tabcolsep + 1pt\relax] &      RCAN (WeakCon) &                    0.6115 &                    0.6207 &                    0.5995 &                    0.5942 &                    0.6614 &                    0.6709 &                    0.5568 &                    0.5754 &                    0.5468 &  \textcolor{blue}{0.5557} &                                 0.5508 \\
             \rowcolor{gray!5}[\dimexpr\tabcolsep + 1pt\relax] &       RCAN (SupMoCo) &                    0.6106 &                    0.6220 &                    0.5968 &                    0.5949 &                    0.6612 &                    0.6733 &                    0.5569 &                    0.5754 &                    0.5464 &                    0.5550 &                                 0.5504 \\
          \rowcolor{gray!5}[\dimexpr\tabcolsep + 1pt\relax] &     RCAN (SupMoCo, all) &  \textcolor{blue}{0.6119} &  \textcolor{blue}{0.6225} &                    0.6009 &                    0.6001 &                    0.6655 &                    0.6740 &                    0.5617 &  \textcolor{blue}{0.5755} &                    0.5465 &                    0.5554 &                                 0.5507 \\
                 \rowcolor{gray!5}[\dimexpr\tabcolsep + 1pt\relax] &          RCAN-DAN &   \textcolor{red}{0.6127} &   \textcolor{red}{0.6229} &   \textcolor{red}{0.6030} &   \textcolor{red}{0.6004} &                    0.6657 &  \textcolor{blue}{0.6759} &   \textcolor{red}{0.5632} &                    0.5754 &                    0.5470 &                    0.5553 &                                 0.5508 \\
 \rowcolor{gray!5}[\dimexpr\tabcolsep + 1pt\relax] \multirow{-14}{*}{\rotatebox[origin=c]{0}{\textbf{BSDS100}}} & RCAN (trained on simple pipeline) &                    0.5742 &                    0.6000 &                    0.3166 &                    0.2976 &   \textcolor{red}{0.7145} &   \textcolor{red}{0.6953} &                    0.2492 &                    0.3197 &                    0.2929 &                    0.2979 &                                 0.2921 \\
         \hdashline
            \rowcolor{gray!15}[\dimexpr\tabcolsep + 1pt\relax] & Bicubic &                    0.7041 &                    0.7315 &                    0.4972 &                    0.5311 &                    0.7131 &                    0.7337 &                    0.4783 &                    0.5390 &                    0.5001 &                    0.4821 &                                 0.4898 \\
                       \rowcolor{gray!15}[\dimexpr\tabcolsep + 1pt\relax] &     Lanczos &                    0.7034 &                    0.7347 &                    0.4793 &                    0.5135 &                    0.7223 &                    0.7428 &                    0.4607 &                    0.5236 &                    0.4858 &                    0.4642 &                                 0.4735 \\
            \rowcolor{gray!15}[\dimexpr\tabcolsep + 1pt\relax] &    RCAN (batch size 4) &                    0.8093 &                    0.8173 &   \textcolor{red}{0.7959} &   \textcolor{red}{0.8028} &                    0.8389 &  \textcolor{blue}{0.8446} &                    0.7623 &                    0.7777 &                    0.7418 &                    0.7468 &                                 0.7438 \\
           \rowcolor{gray!15}[\dimexpr\tabcolsep + 1pt\relax] &     RCAN (batch size 8) &                    0.8093 &                    0.8182 &                    0.7958 &                    0.8017 &                    0.8356 &                    0.8397 &                    0.7613 &                    0.7784 &                    0.7427 &                    0.7467 &                                 0.7441 \\
             \rowcolor{gray!15}[\dimexpr\tabcolsep + 1pt\relax] &      RCAN (non-blind) &                       N/A &                       N/A &                       N/A &                       N/A &                       N/A &                       N/A &                       N/A &                       N/A &  \textcolor{blue}{0.7461} &                    0.7469 &                                 0.7466 \\
      \rowcolor{gray!15}[\dimexpr\tabcolsep + 1pt\relax] &    RCAN (non-blind, no blur) &                       N/A &                       N/A &                       N/A &                       N/A &                       N/A &                       N/A &                       N/A &   \textcolor{red}{0.7850} &                    0.7443 &  \textcolor{blue}{0.7483} &               \textcolor{blue}{0.7469} \\
      \rowcolor{gray!15}[\dimexpr\tabcolsep + 1pt\relax] &   RCAN (non-blind, no noise) &                       N/A &                       N/A &                       N/A &                       N/A &                       N/A &                       N/A &                       N/A &                       N/A &                    0.7438 &                    0.7453 &                                 0.7444 \\
\rowcolor{gray!15}[\dimexpr\tabcolsep + 1pt\relax] &   RCAN (non-blind, no compression) &                       N/A &                       N/A &                       N/A &                       N/A &                       N/A &                       N/A &   \textcolor{red}{0.7657} &                       N/A &   \textcolor{red}{0.7477} &                    0.7476 &                \textcolor{red}{0.7470} \\
             \rowcolor{gray!15}[\dimexpr\tabcolsep + 1pt\relax] &           RCAN (MoCo) &                    0.8072 &                    0.8174 &                    0.7947 &                    0.8012 &                    0.8268 &                    0.8358 &                    0.7617 &                    0.7767 &                    0.7419 &                    0.7465 &                                 0.7436 \\
            \rowcolor{gray!15}[\dimexpr\tabcolsep + 1pt\relax] &         RCAN (WeakCon) &                    0.8078 &                    0.8177 &                    0.7879 &                    0.7901 &                    0.8295 &                    0.8370 &                    0.7464 &                    0.7758 &                    0.7415 &                    0.7465 &                                 0.7433 \\
            \rowcolor{gray!15}[\dimexpr\tabcolsep + 1pt\relax] &         RCAN (SupMoCo) &  \textcolor{blue}{0.8096} &  \textcolor{blue}{0.8193} &                    0.7889 &                    0.7999 &                    0.8294 &                    0.8380 &                    0.7594 &                    0.7793 &                    0.7418 &                    0.7471 &                                 0.7438 \\
           \rowcolor{gray!15}[\dimexpr\tabcolsep + 1pt\relax] &     RCAN (SupMoCo, all) &                    0.8091 &                    0.8192 &                    0.7922 &                    0.8015 &                    0.8309 &                    0.8377 &                    0.7575 &                    0.7793 &                    0.7409 &                    0.7464 &                                 0.7432 \\
              \rowcolor{gray!15}[\dimexpr\tabcolsep + 1pt\relax] &            RCAN-DAN &   \textcolor{red}{0.8111} &   \textcolor{red}{0.8212} &  \textcolor{blue}{0.7959} &  \textcolor{blue}{0.8024} &  \textcolor{blue}{0.8398} &                    0.8438 &  \textcolor{blue}{0.7626} &  \textcolor{blue}{0.7795} &                    0.7442 &   \textcolor{red}{0.7485} &                                 0.7456 \\
\rowcolor{gray!15}[\dimexpr\tabcolsep + 1pt\relax] \multirow{-14}{*}{\rotatebox[origin=c]{0}{\textbf{Manga109}}} & RCAN (trained on simple pipeline) &                    0.7012 &                    0.7575 &                    0.3702 &                    0.3997 &   \textcolor{red}{0.8975} &   \textcolor{red}{0.8600} &                    0.3368 &                    0.4332 &                    0.3998 &                    0.3266 &                                 0.3581 \\
                \hdashline
         \rowcolor{gray!5}[\dimexpr\tabcolsep + 1pt\relax]       & Bicubic &                    0.5791 &                    0.5964 &                    0.4485 &                    0.4372 &                    0.5737 &                    0.5969 &                    0.3804 &                    0.4321 &                    0.3868 &                    0.4189 &                                 0.4014 \\
                   \rowcolor{gray!5}[\dimexpr\tabcolsep + 1pt\relax] &        Lanczos &                    0.5809 &                    0.6016 &                    0.4367 &                    0.4241 &                    0.5834 &                    0.6069 &                    0.3671 &                    0.4206 &                    0.3759 &                    0.4083 &                                 0.3904 \\
           \rowcolor{gray!5}[\dimexpr\tabcolsep + 1pt\relax] &    RCAN (batch size 4) &  \textcolor{blue}{0.6696} &  \textcolor{blue}{0.6842} &                    0.6625 &  \textcolor{blue}{0.6614} &                    0.7006 &                    0.7086 &                    0.6077 &                    0.6329 &                    0.5861 &                    0.6074 &                                 0.5957 \\
           \rowcolor{gray!5}[\dimexpr\tabcolsep + 1pt\relax] &    RCAN (batch size 8) &                    0.6692 &                    0.6834 &  \textcolor{blue}{0.6629} &                    0.6601 &                    0.6985 &                    0.7063 &                    0.6071 &                    0.6340 &                    0.5872 &                    0.6071 &                                 0.5963 \\
          \rowcolor{gray!5}[\dimexpr\tabcolsep + 1pt\relax] &        RCAN (non-blind) &                       N/A &                       N/A &                       N/A &                       N/A &                       N/A &                       N/A &                       N/A &                       N/A &  \textcolor{blue}{0.5920} &                    0.6074 &                                 0.5983 \\
       \rowcolor{gray!5}[\dimexpr\tabcolsep + 1pt\relax] &  RCAN (non-blind, no blur) &                       N/A &                       N/A &                       N/A &                       N/A &                       N/A &                       N/A &                       N/A &   \textcolor{red}{0.6377} &                    0.5885 &                    0.6084 &                                 0.5971 \\
      \rowcolor{gray!5}[\dimexpr\tabcolsep + 1pt\relax] &  RCAN (non-blind, no noise) &                       N/A &                       N/A &                       N/A &                       N/A &                       N/A &                       N/A &                       N/A &                       N/A &                    0.5883 &                    0.6059 &                                 0.5953 \\
 \rowcolor{gray!5}[\dimexpr\tabcolsep + 1pt\relax] & RCAN (non-blind, no compression) &                       N/A &                       N/A &                       N/A &                       N/A &                       N/A &                       N/A &   \textcolor{red}{0.6122} &                       N/A &   \textcolor{red}{0.5929} &  \textcolor{blue}{0.6088} &                \textcolor{red}{0.6000} \\
                \rowcolor{gray!5}[\dimexpr\tabcolsep + 1pt\relax] &       RCAN (MoCo) &                    0.6686 &                    0.6831 &                    0.6626 &                    0.6594 &                    0.6886 &                    0.7017 &                    0.6052 &                    0.6340 &                    0.5874 &                    0.6084 &                                 0.5968 \\
               \rowcolor{gray!5}[\dimexpr\tabcolsep + 1pt\relax] &     RCAN (WeakCon) &                    0.6693 &                    0.6836 &                    0.6617 &                    0.6511 &                    0.6941 &                    0.7037 &                    0.5909 &                    0.6348 &                    0.5872 &                    0.6080 &                                 0.5966 \\
              \rowcolor{gray!5}[\dimexpr\tabcolsep + 1pt\relax] &      RCAN (SupMoCo) &                    0.6684 &                    0.6834 &                    0.6571 &                    0.6538 &                    0.6912 &                    0.7042 &                    0.5987 &                    0.6343 &                    0.5863 &                    0.6079 &                                 0.5960 \\
            \rowcolor{gray!5}[\dimexpr\tabcolsep + 1pt\relax] &   RCAN (SupMoCo, all) &                    0.6676 &                    0.6824 &                    0.6592 &                    0.6563 &                    0.6968 &                    0.7053 &                    0.6005 &                    0.6324 &                    0.5852 &                    0.6070 &                                 0.5952 \\
               \rowcolor{gray!5}[\dimexpr\tabcolsep + 1pt\relax] &            RCAN-DAN &   \textcolor{red}{0.6719} &   \textcolor{red}{0.6870} &   \textcolor{red}{0.6657} &   \textcolor{red}{0.6632} &  \textcolor{blue}{0.7027} &  \textcolor{blue}{0.7109} &  \textcolor{blue}{0.6094} &  \textcolor{blue}{0.6357} &                    0.5891 &   \textcolor{red}{0.6099} &               \textcolor{blue}{0.5985} \\
 \rowcolor{gray!5}[\dimexpr\tabcolsep + 1pt\relax] \multirow{-14}{*}{\rotatebox[origin=c]{0}{\textbf{Urban100}}} & RCAN (trained on simple pipeline) &                    0.5897 &                    0.6481 &                    0.3616 &                    0.3396 &   \textcolor{red}{0.7688} &   \textcolor{red}{0.7395} &                    0.2733 &                    0.3490 &                    0.3067 &                    0.3216 &                                 0.3105 \\
\bottomrule
            \end{tabularx}
		\end{center}
\label{complex_blind_psnr}
\end{table}

\begin{figure}[H]
        \centering
        \includegraphics[width=\linewidth]{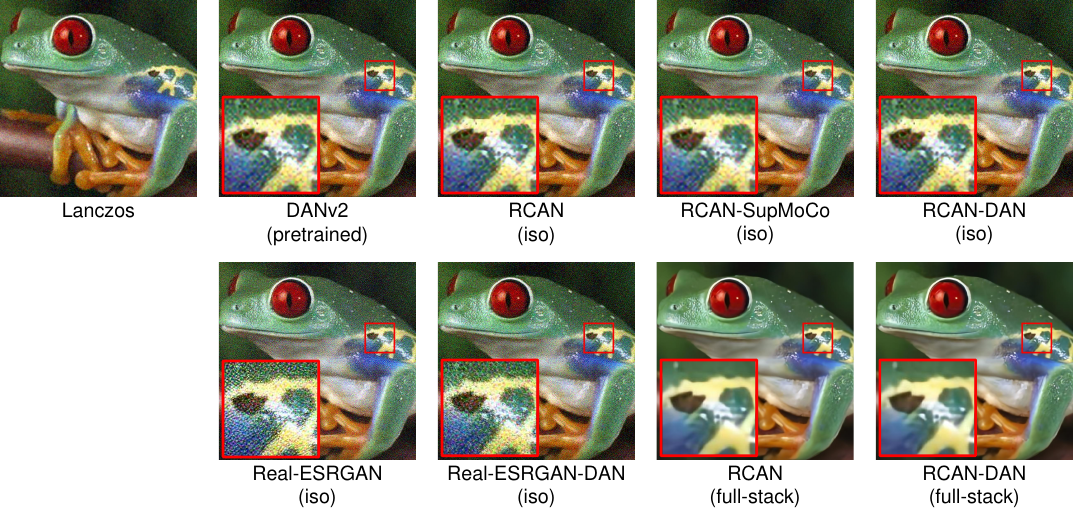}
    \caption{Comparison of the SR results of various models on the frog image from RealSRSet.  All simple pipeline models (marked as iso) are incapable of removing the noise from the image.  The complex pipeline models (marked as full-stack) produce significantly improved results.}
    \label{fig:frog_compare}
\end{figure}